%% file: icml_main.tex
\documentclass{article}

\usepackage{microtype}
\usepackage{booktabs} %

\usepackage{hyperref}

\usepackage[accepted]{icml2024}

\input{preamble}
\usepackage{amsmath}
\usepackage{amssymb}
\usepackage{mathtools}
\usepackage{amsthm}

\usepackage[capitalize,noabbrev]{cleveref}

\theoremstyle{plain}

\theoremstyle{definition}

\theoremstyle{remark}

\usepackage[textsize=tiny]{todonotes}

\icmltitlerunning{Leveraging VLM-Based Pipelines to Annotate 3D Objects}
\begin{document}

\twocolumn[
\icmltitle{Leveraging VLM-Based Pipelines to Annotate 3D Objects}

\icmlsetsymbol{equal}{*}

\begin{icmlauthorlist}
\icmlauthor{Rishabh Kabra}{dm,ucl}
\icmlauthor{Loic Matthey}{dm}
\icmlauthor{Alexander Lerchner}{dm}
\icmlauthor{Niloy J. Mitra}{ucl}
\end{icmlauthorlist}

\icmlaffiliation{dm}{Google DeepMind}
\icmlaffiliation{ucl}{University College London}

\icmlcorrespondingauthor{Rishabh Kabra}{rkabra@google.com}

\icmlkeywords{Machine Learning, ICML}

\urlstyle{same}
\begin{center}
    \url{https://leveraging-vlms-to-annotate-3d-objects.github.io}
\end{center}

\vskip 0.3in
    ]

\printAffiliationsAndNotice{}  %

\input{sec/0_abstract}
\input{sec/1_intro}

\bibliography{egbib}
\bibliographystyle{icml2024}

\input{X_suppl}

\end{document}

%% file: preamble.tex
\newcommand{\showorhide}[1]{}

\usepackage{multirow}
\usepackage[utf8]{inputenc} %
\usepackage[T1]{fontenc}    %
\usepackage{url}            %
\usepackage{booktabs}       %
\usepackage{amsfonts}       %
\usepackage{nicefrac}       %
\usepackage{microtype}      %
\usepackage{xcolor, soul}         %
\usepackage{graphicx, caption, subcaption}
\usepackage{longtable}
\usepackage{bbm}
\usepackage{mathtools}
\usepackage{enumitem}
\usepackage{verbatim}
\usepackage{listings}
\usepackage{tabularx}
\usepackage[percent]{overpic}

\definecolor{OliveGreen}{rgb}{0.58,0.77,0.49}
\sethlcolor{OliveGreen}
\newcommand{\rulesep}{\unskip\ \vrule\ }
\newcommand{\defeq}{\vcentcolon=}

\newcommand*{\CAMERAREADY}{}  %

%% file: sec/0_abstract.tex
\begin{figure*}[h!]
    \centering
    \includegraphics[width=0.9\textwidth]{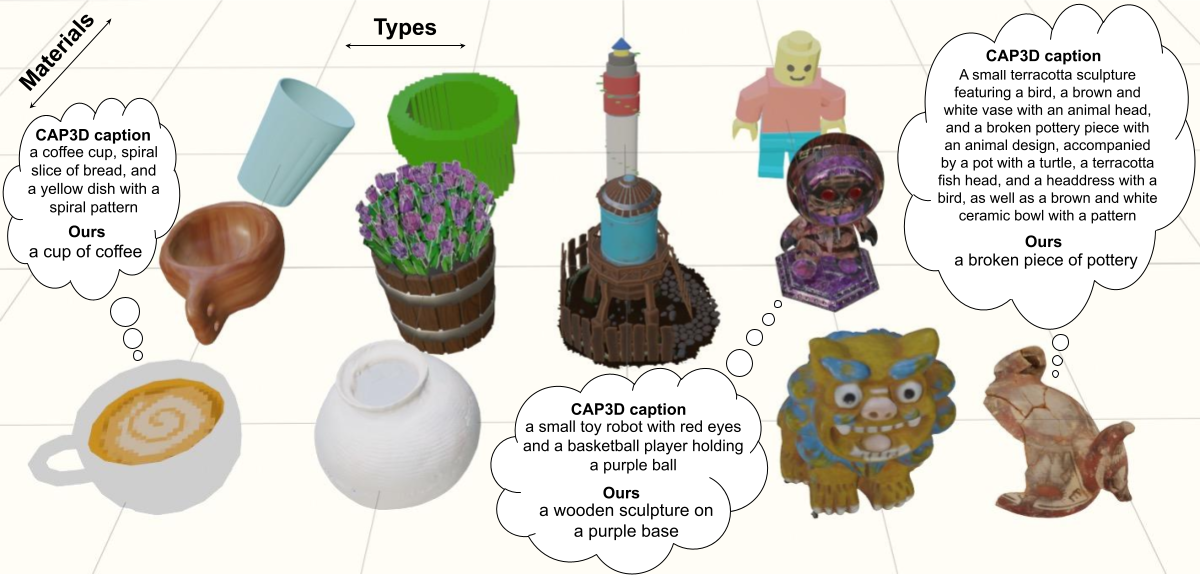}
    \caption{\textbf{An inferred conceptual subgrid for 3D objects.} We probe pretrained VLMs on how well they infer object properties such as type (e.g., ``cup'', ``pot'', ``tower'') and material (e.g., ``ceramic'', ``wood'', ``plastic''). We apply a visually grounded aggregation across 2D views to avoid hallucinated outputs. Our method, ScoreAgg, produces more reliable captions than the current SoTA, CAP3D \cite{luo2023scalable}, as shown in the callouts. 
    }
    \label{fig:objaverse_conceptual_grid}
\end{figure*}

\begin{abstract}
Pretrained vision language models (VLMs) present an opportunity to caption unlabeled 3D objects at scale. The leading approach to summarize VLM descriptions from different views of an object \cite{luo2023scalable} relies on a language model (GPT4) to produce the final output. 
This text-based aggregation is susceptible to hallucinations as it merges potentially contradictory descriptions.
We propose an alternative algorithm to marginalize over factors such as the viewpoint that affect the VLM's response. Instead of merging text-only responses, we utilize the VLM's joint image-text likelihoods.
We show our probabilistic aggregation is not only more reliable and efficient, but sets the SoTA on inferring object types with respect to human-verified labels.
The aggregated annotations are also useful for conditional inference; they improve downstream predictions (e.g., of object material) when the object’s type is specified as an auxiliary text-based input.
Such auxiliary inputs allow ablating the contribution of visual reasoning over visionless reasoning in an unsupervised setting.
With these supervised and unsupervised evaluations, we show how a VLM-based pipeline can be leveraged to produce reliable annotations for 764K objects from the Objaverse dataset.

\end{abstract}

%% file: sec/1_intro.tex
\section{Introduction}
\label{sec:intro}

Numerous applications could benefit from a zero-shot VLM pipeline to identify the type and nature of a 3D object from views of it. We assess the design choices for such a pipeline: (i) what images to use, (ii) what VLM, (iii) how to prompt the VLM, (iv) how to process multi-view or multi-prompt responses to produce an aggregate, and (v) what auxiliary information can be provided to improve inference. These stages can be evaluated and optimized using sparse labeled data. But to scale VLM pipeline assessment beyond validation accuracy, we also develop unsupervised metrics to track hallucination and ablate visual reasoning.

Datasets of 3D objects (e.g., Objaverse from \citet{deitke2023objaverse}) provide a rich testing ground for such a pipeline. We target the generation of useful text pairings for 764K object assets. We look not only at captioning but also inferring specific properties such as type, material, physical behavior, affordances, or relations like containment between objects. These property-specific, open-vocabulary annotations help index the dataset on a set of conceptual axes (Fig \ref{fig:objaverse_conceptual_grid}).

To address the challenge of inspecting a 3D object using 2D views, we generate VLM responses with accompanying image-text likelihoods. This facilitates a visually grounded score-based aggregation (ScoreAgg) to determine the most reliable responses across different object views. Our algorithm is novel, general, and can be applied to arbitrary factors (besides viewpoint) varied across VLM queries. It compares favorably with text-only summarization; the latter requires additional computation, some instruction tuning, yet tends to propagate contradictions (i.e., hallucinations). %

We also explore conditional inference via prompt-chaining (together with ScoreAgg) to boost the accuracy of VLM annotations. These methods help leverage off-the-shelf VLMs, without retraining or task-specific in-context learning. As ScoreAgg produces a probabilistic output, it also helps quantify the uncertainty across possible responses. The aggregation exhibits increasing accuracy as we sample more VLM responses per probe or run more VLM probes, thus permitting flexible use of computation to improve reliability. These capabilities could help optimize VLM pipelines more broadly than our particular focus on annotating 3D objects.

We organize the paper around the design choices we introduced for a zero-shot VLM annotation pipeline. Section~\ref{sec:background} lays out some background and prior work. In Section~\ref{sec:semantic_description}, we aim to summarize type annotations reliably under changes in view or prompt. In Section~\ref{sec:physical_properties}, we assess the inference of object material using a variety of VLMs and conditioning inputs. Both sections rely on human-verified labels as well as unsupervised metrics to evaluate the annotations and pipeline choices. In Appendix \ref{sec:appearance}, we also vary the rendering of images or object appearances to study the impact on VLM performance.

Our salient contributions are the following--we:
\begin{enumerate}[nolistsep]
    \item Introduce a visually grounded aggregation (\emph{ScoreAgg}) of VLM responses across multiple queries.
    \item Compare our annotations with a leading approach based on GPT4 (CAP3D \cite{luo2023scalable}). We use \emph{caption blow-up ratios} as a measure of hallucination to show our method is reliable where CAP3D is not.
    \item Establish a SoTA on type and material inference w.r.t. given and collected human labels respectively.
    \item Propose an unsupervised \emph{visual sensitivity metric} that is predictive of VLM accuracy.
    \item Are releasing 5M aggregated captions and annotations for Objaverse. These are available via our project page.%
\end{enumerate}

\section{Background and Setting}
\label{sec:background}

\begin{figure*}[!ht]
    \centering
    
    \begin{subfigure}{0.95\textwidth}
        \centering
        \textbf{A.} Multi-view differences can produce varying object descriptions
    \end{subfigure}%
    
    \begin{subfigure}[t]{.24\textwidth}
        \centering
        \includegraphics[width=\textwidth]{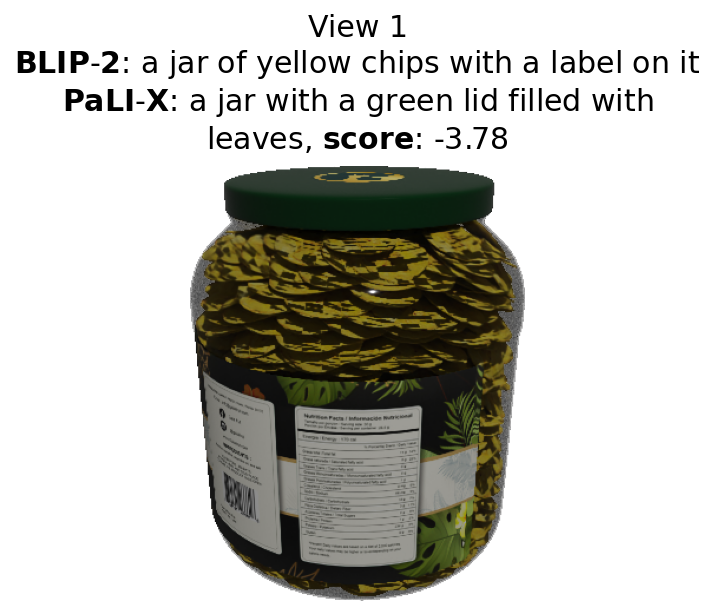}
    \end{subfigure}\hfill%
    \begin{subfigure}[t]{.30\textwidth}
        \centering
        \includegraphics[width=\textwidth]{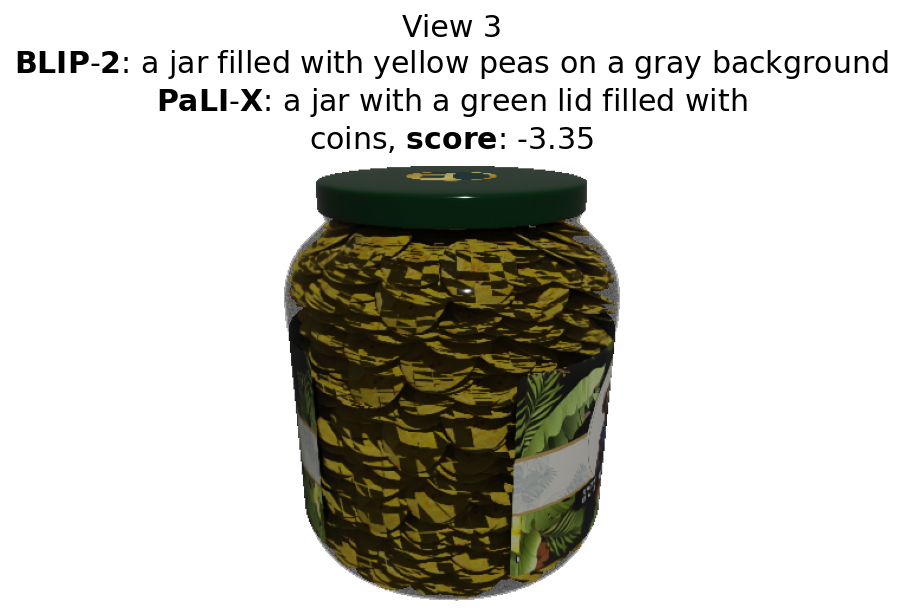}
    \end{subfigure}\hfill%
    \begin{subfigure}[t]{.25\textwidth}
        \centering
        \includegraphics[width=\textwidth]{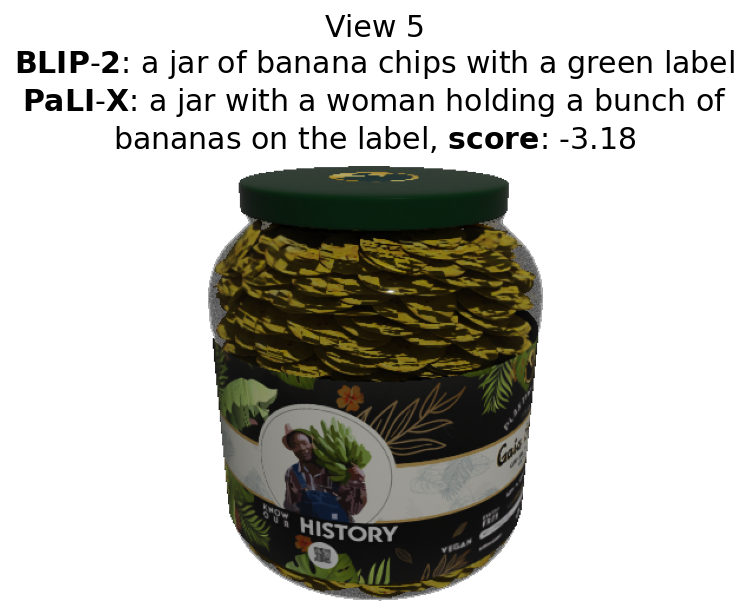}
    \end{subfigure}%
    \begin{subfigure}{.204\textwidth}
        \centering
        \includegraphics[width=\textwidth]{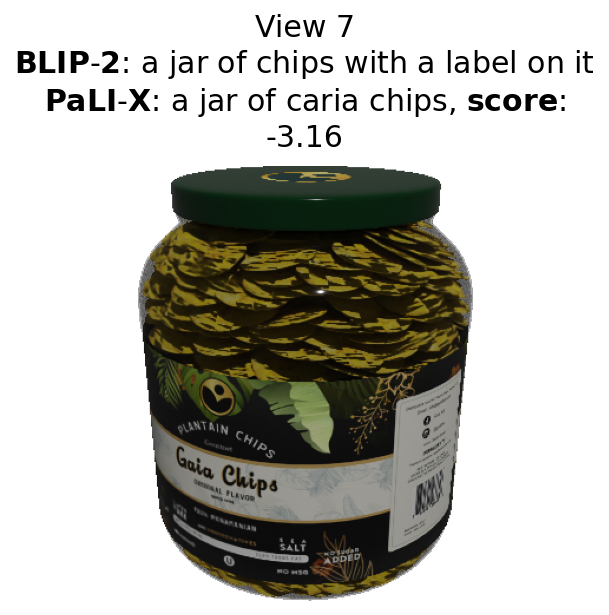}
    \end{subfigure}
    
    \vspace{\fill}
    \vspace{0.2cm}
    
    \begin{subfigure}{.40\textwidth}
        \centering
        \small{\textbf{B1.} Aggregation in text space using an LLM and engineered prompt (CAP3D)}
        \includegraphics[trim={7cm 9cm 7.8cm 10.4cm},clip,width=\textwidth]{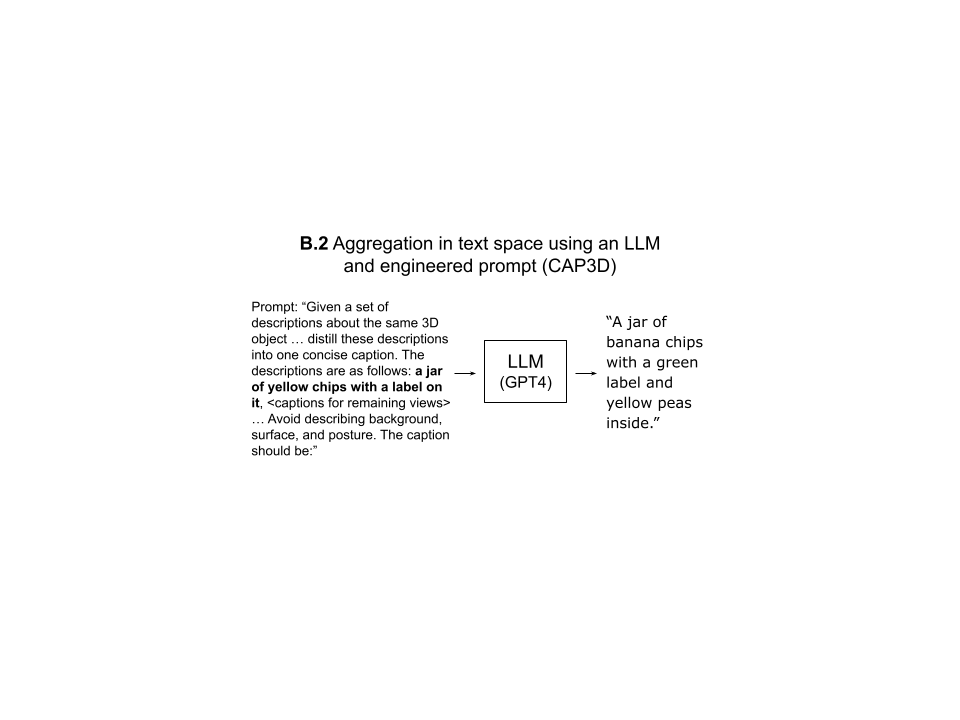}
    \end{subfigure}%
    \rulesep
    \begin{subfigure}{.55\textwidth}
        \centering
        \small{\textbf{B2.} Aggregation using available VLM scores of each description (\textbf{ours})}
        \includegraphics[trim={0.25cm 0.3cm 0cm 0.4cm},clip,width=\textwidth]{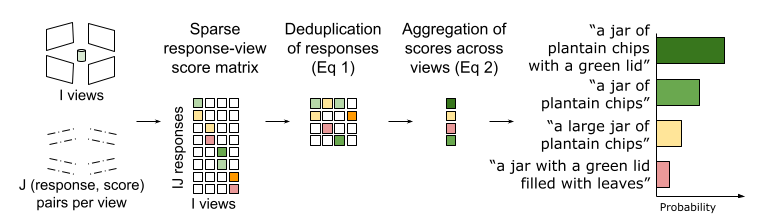}
    \end{subfigure}%
    
    \caption{\textbf{A.}~Three of eight regularly spaced views of a 3D object. Each view is accompanied by the top caption produced by two different models: BLIP-2 \& PaLI-X. Captions from BLIP-2 were obtained from the competitive CAP3D baseline, whereas captions from PaLI-X were generated with accompanying scores for this work. Both models show an expected variation in responses across views. \textbf{B1.}~To aggregate multi-view captions, CAP3D feeds them to GPT4 and prompts it for an object-level summary. The LLM is unable to reconcile captions from different views, and simply adds up the contents. \textbf{B2.}~Our algorithm downweights unreliable responses by combining scores across views. We show its top-4 outputs. %
    }
    \label{fig:headline}
\end{figure*}

\textbf{Dataset.} Our main target is Objaverse 1.0 \cite{deitke2023objaverse}, an internet-scale collection of 800K diverse but poorly annotated 3D models. They were uploaded by 100K artists to the Sketchfab platform. While the uploaded tags and descriptions are inconsistent and unreliable, a subset of 44K objects called Objaverse-LVIS is accompanied by human-verified categories. We rely on it to validate our semantic annotations.
We also introduce a subset with material labels to test material inference. %
Other datasets we considered include OmniObject3D \cite{wu2023omniobject3d}, ABO \cite{collins2022abo}, and ScanNeRF \cite{deluigi2023scannerf}. But they lack the scale and potential of Objaverse---for instance, the number of object classes in other datasets is a few hundred, compared to 1156 in Objaverse-LVIS alone. 

\textbf{Baseline.} A three-module pipeline was recently proposed to generate captions for Objaverse. Though we aim to go beyond captioning and supervised evaluation, we rely on CAP3D \cite{luo2023scalable} as the primary baseline for our work. Their pipeline is as follows: a VLM (BLIP-2 \cite{li2023blip}) first produces 5 candidate captions for 8 object views; CLIP \cite{radford2021learning} filters all but one caption per view, and GPT4 \cite{openai2023gpt4} generates an aggregated caption. We found this last step to be prone to hallucinations (discussed in detail in Section \ref{sec:semantic_description}). Our procedure is similar up to CAP3D’s first stage, but we don't use any further modules for filtering or summarization.%

\textbf{Models.} To generate our own captions or annotations, we use variants of PaLI-X pretrained specifically for captioning or visual question answering. Both variants consist of a ViT-22B \cite{dehghani2023scaling} vision model and 32B UL2 \cite{tay2022unifying} language backbone, totaling 55B parameters. For material prediction, we also run BLIP-2 T5 XL \cite{li2023blip} as a baseline. All models are run zero-shot, one input image at a time, and output an autoregressive distribution over language tokens. The likelihood of any sampled text can be computed during the VLM sampling process (e.g., beam search) without any additional cost. None of our methods or results are specific to PaLI or BLIP.

\subsection{Related Work}

We present a literature review in Appendix \ref{sec:literature_review}. Though applying VLMs to 3D domains remains under-explored, the following papers are close to certain aspects of our work:

\begin{enumerate}[nolistsep]
    \item \citet{zhu2023chatgpt} propose a VQA-based approach to caption 2D images. They use an LLM (GPT) to generate questions about image contents, a VLM (BLIP-2) to answer them, and finally an LLM to produce a summary caption. CAP3D takes a similar approach with an extra filtering step before summarization.
    \item \citet{o2023approaching} attempt to replicate human-level 3D-shape understanding using multi-view learning objectives. But all tested models fall short of human performance on held-out ShapeNet objects.
    \item \citet{gao2023physically} crowd-source a dataset to explore VLM-based inference of object physical properties in natural images. VLMs like InstructBLIP \cite{instructblip} benefit from fine-tuning on their data.
    \item A method that contends with fusing VLM outputs (but assuming posed RGB-D data) is ConceptGraphs \cite{gu2023conceptgraphs}. Their focus on building scene graphs to map environments is different from our goal of generating object-centric, property-specific annotations.
\end{enumerate}

\section{Type Annotation}  %
\label{sec:semantic_description}

Our first task is to infer the type of each Objaverse object in a zero-shot, open-vocabulary setting. The task is compelling because only $\sim 5\% $ of Objaverse is accompanied by verified category labels. Being able to predict them would help shed light on the rest of the dataset. We also expect asking for the type of an object to be a language-amenable query, and hence a basic test for VLMs.

Despite how simple a task this initially appears, the challenge of captioning a 3D object is evident from Fig~\ref{fig:headline}-A. The current SoTA for 3D captioning (CAP3D) relies on GPT4 to summarize annotations across multiple views of an object. This can produce deeply flawed summaries---the LLM propagates hallucinations or confusions when there's contrasting details across views (see Fig~\ref{fig:headline}-B1). Though it is explicitly instructed that it is given captions of one object, the LLM can interpret them as multiple co-occurring objects. It lacks the visual context to reconcile contradictions.

To address this, we propose an alternative method of aggregating multi-view or multi-query annotations (Fig~\ref{fig:headline}-B2). We describe the algorithm in Section~\ref{sec:aggregation}. We compare baseline sources with captions and type annotations produced by our method in Section~\ref{sec:quantitative_type_evaluation}. We then present a measure of hallucination (Section~\ref{sec:measure_of_hallucination}) to show our method is reliable where CAP3D is not. Finally, we unpack the performance of our aggregation and show how it scales in Section~\ref{sec:why_aggregate}.

\begin{figure*}[!t]
  
  \begin{minipage}{.65\linewidth}
  \centering
   \begin{overpic}[trim={0cm 0.3cm 0cm 0cm},clip,width=\textwidth,percent]{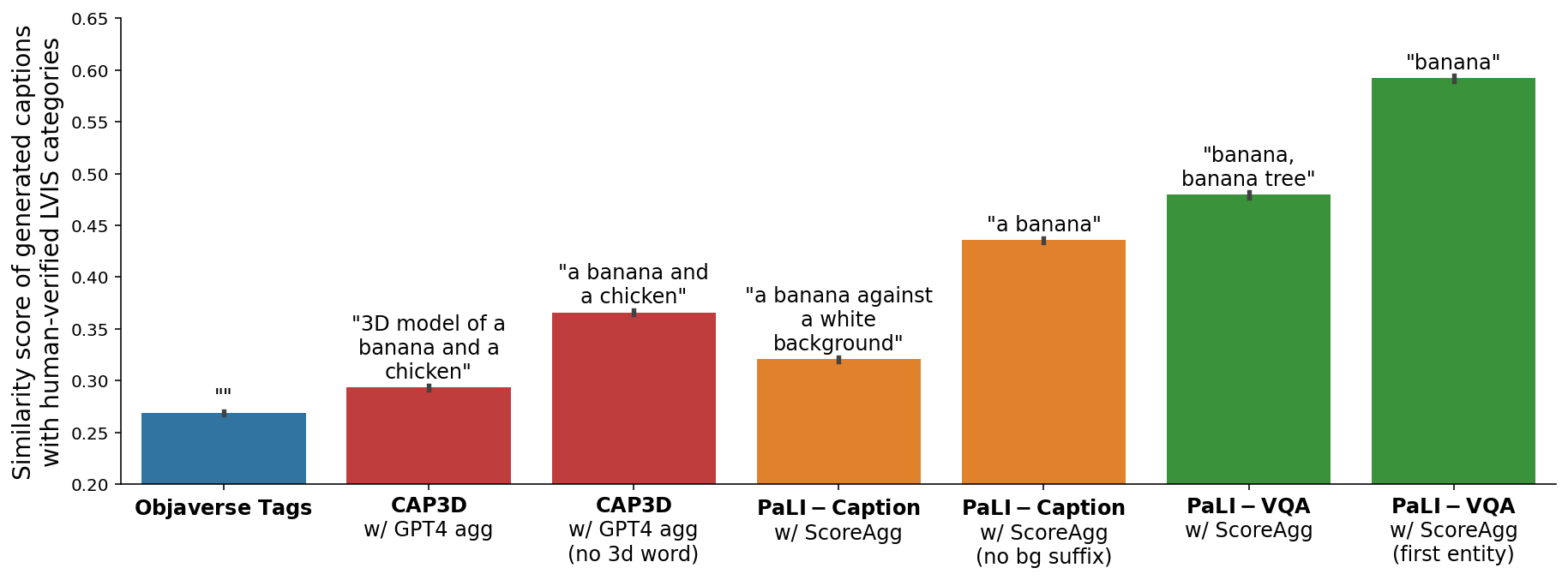}
      \put(10,21){\includegraphics[scale=0.29]{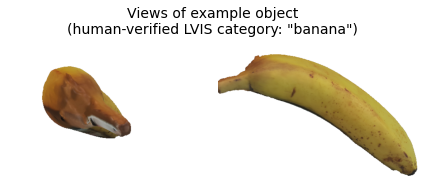}}  %
  \end{overpic}
  \end{minipage}
  \begin{minipage}{.2\linewidth}
    \centering
    \scalebox{0.92}{
    \begin{tabular}{lcc}
    & \multicolumn{1}{c}{Objv. tags} & \multicolumn{1}{c}{PaLI VQA} \\ \hline
    Top-1 acc.        & $0.06 \pm 0.24$       & $0.26 \pm 0.44$  \\
    Top-5 acc.        & $0.13 \pm 0.34$       & $0.52 \pm 0.50$  \\
    Top-$\infty$ acc. & $0.15 \pm 0.36$       & $0.67 \pm 0.47$  \\
    Soft acc.         & $0.04 \pm 0.17$       & $0.19 \pm 0.26$ \\ \hline
    \end{tabular}
    }
  \end{minipage}
  \vspace{-5pt}
\caption{\textbf{Comparison of captions and type annotations generated from different sources/models.}  All results are averaged over Objaverse-LVIS. \textbf{Left}: the bars show text-embedding similarity scores ($\uparrow$) for the aggregate/top per-object descriptions from each source. We show an example caption/type annotation above each bar; these correspond to a fixed object shown in the upper left corner from two different views. \textbf{Right}: string-match metrics that assess full predicted distributions from two sources. We report top-k accuracies (whether the correct type is in the top-k predictions, $\uparrow$) as well as the soft accuracy (probability of the correct type in the output distribution, $\uparrow$).
}
\label{fig:type_comparison}
\end{figure*}

\subsection{Visually Grounded Score-Based Aggregation}
\label{sec:aggregation}

We introduce a method for aggregating VLM outputs across multiple queries that relies on the log-likelihoods or scores of the sampled outputs. Say we run $I$ queries (spanning different views/prompts) to get $J$ responses per query, denoted $\{r_{i,j}\}$. This forms our support for the aggregate output distribution, i.e., we have $IJ$ possibilities for the final response.

An expensive approach to estimate the compatibility between all responses and queries would use the VLM likelihood $p(r | x, q, \theta)$ to score $\{r_{i,j}\}$ with respect to each image $x$ and question/prompt $q$. This would give us a dense response-score matrix, sized $IJ \times I$. We could then marginalize across the $I$ columns to compute global log-likelihoods for all $IJ$ responses. Obtaining the dense matrix, however, would require not only $\mathcal{O}(IJ)$ forward passes through the VLM sampler (to obtain the $IJ$ candidates), but also a subsequent scoring cost of $\mathcal{O}(I^2J)$. The quadratic factor $I^2$ would make it harder to run more queries (multiple views/prompts) to ensure the final output is reliable.%

\textbf{ScoreAgg:} We posit that when VLM queries are correlated (e.g., views of the same object or paraphrased questions), we'll get recurring responses across queries. We also exploit the fact that we can obtain one score per candidate response for free. Paying only the initial $\mathcal{O}(IJ)$ VLM sampling cost, we obtain a sparse response-score matrix containing $IJ$ response-score pairs $\{(r_{i,j}, s_{i, j})\}$. Let $f$ be a map to postprocess strings and reduce them to a canonical form. The following aggregation helps identify responses $r$ that occur frequently while accounting for the model's confidence in each occurrence. $\forall r \in \{r_{i,j}\}$: \begin{align}
    & s_i(r) \defeq \sup \{s_{i,j} \mid f(r_{i,j}) = r \textrm{ and } j = 1, 2, ..., J\} \label{eq1} \\
    & s_{agg}(r) \defeq \log \sum_{i} \exp(s_i(r)) \label{eq2} \\
    & \tilde{p}(r | \{r_{i,j}, s_{i,j}\}) \defeq \exp(s_{agg}(r)) / \sum_{r'} \exp(s_{agg}(r')) \label{eq3} %
\end{align}
Eq \ref{eq1} deduplicates and re-scores responses for a given query $i$. The string processor $f$ determines when $r_{i,j}$ is treated equivalent to $r$, and can be customized per VLM. This is useful when responses are identical up to punctuation, case, or uninformative tokens. Since these are undesirable duplicates, we want to avoid accumulating their scores, so we take the supremum instead. Note that $s_i(r)$ can be $-\infty$ if no $r$ equivalent occurs in the $J$ responses for query $i$. Eq \ref{eq2} then aggregates scores across occurrences of $r$ in distinct queries. These are desirable duplicates (over distinct images or prompts) that merit reinforcing. Finally, Eq \ref{eq3} computes an aggregate probability distribution over responses by taking a softmax over the aggregate scores. See Fig \ref{fig:headline}-B2 for a visual overview of the algorithm.

By assuming some overlap across the $I$ queries, our approach avoids the quadratic scoring cost of the strawman approach. Our final complexity is just $\mathcal{O}(IJ)$. (This in fact helps us run more queries to ensure overlapping responses.) %
To work with the fact that we expect fewer than $I$ scores for each de-duplicated response $r$, we cannot add log-likelihoods in Eq \ref{eq2}---a sum would make the aggregate score smaller for responses that occur frequently (since the log-likelihoods are negative), while less frequent responses avoid that treatment. Instead we need an aggregation function that is non-decreasing in the number of available scores. Log-sum-exp and max are two choices (which we compare in Section \ref{sec:why_aggregate}, showing the former works better). Other choices would require manipulating the scores by adding a positive offset or scaling by a positive constant, but those cannot be chosen in a principled way.

Compared to model-based summarization (e.g., using an LLM), ScoreAgg requires a simple numerical computation. %
Whereas an LLM needs a task-specific prompt, our algorithm can aggregate over arbitrary VLM queries (e.g., to control factors other than the viewpoint). 
While an LLM produces a point estimate, our method outputs a distribution over possible responses.
Although our method inherits potential flaws in a given VLM's scoring, it would be straightforward to decouple the two VLM outputs if desired---we could score $\{r_{i,j}\}$ using a different model and replace $\{s_{i,j}\}$ before Eq~\ref{eq1}. $f$ could also be used to implement arbitrarily complex string matching or textual entailment \cite{Korman2018-KORDTE} to aggregate similar responses.

\subsection{Evaluation w.r.t. Human Labels}
\label{sec:quantitative_type_evaluation}

We collect four sets of semantic descriptions for Objaverse:

\begin{enumerate}[leftmargin=*,nolistsep]
    \item \textbf{Objaverse tags (baseline)}: these were uploaded by the creator of each 3D asset. They are inherently noisy and inconsistent between objects. We comma-separate the tags to produce an aggregate string for each object. But when we compute distributional metrics, we treat the tags separately, as ordered but uniformly likely.
    \item \textbf{CAP3D captions (baseline)}: these were generated and released by \citet{luo2023scalable}. A post-processed version removes the frequent prefix, ``3D model of''. We compare both versions.
    \item \textbf{PaLI captions (ours)}: using a captioning variant of PaLI and simple prompt (``A picture of ''), we generated descriptive captions similar to CAP3D's first stage. We then applied ScoreAgg to summarize $J=5$ responses across $I=8$ views per object. We compare results with and without a post-processing map $f$ (Eq \ref{eq1}) to ignore suffixes of the form ``on/against a white background.'' %
    \item \textbf{PaLI VQA annotations (ours)}: we used four VQA prompts to probe for the type of each object: (i)~``What is this?'' (ii)~``What type of object is this?'' (iii)~``What is in the image?'' (iv)~``Describe the object in the image''. This produced 4 sets of top-5 responses per view ($I=4*8=32,\ J=5$). The responses are typically WordNet entities \cite{miller1995wordnet} that group synonyms or related terms in a comma-separated list. We deduplicate responses by taking the first such term per response. This post-processing map is also ablated.
\end{enumerate}

We compare outputs from these sources to human-verified object categories from Objaverse-LVIS. For our sources, we take the likeliest output from each aggregate distribution. We then embed all text using an independent language encoder, the Universal Sentence Encoder (v4) \cite{cer2018universal} from TensorFlow-Hub. Finally, we compute cosine similarities between the embedded outputs and verified categories. With 512-dim embeddings, the sentence encoder allows comparisons that are invariant to sentence length.

Fig~\ref{fig:type_comparison}-L shows that all VLM pipelines outperform tags from the original dataset. PaLI captions (using ScoreAgg) outperform CAP3D substantially. The quantitative improvement is due to less hallucination and increased accuracy from view-aggregation, as we show in Sections~\ref{sec:measure_of_hallucination} and~\ref{sec:why_aggregate}.

PaLI VQA annotations perform significantly better than the rest. They match ground-truth string labels on a large fraction of validation data without being trained for the task: our output distributions contain the exact expected type on two-thirds of Objaverse-LVIS (Fig~\ref{fig:type_comparison}-R). The soft accuracy (19\%) is significant considering there are up to $IJ=160$ unique responses in each aggregate output distribution.

\subsection{An Unsupervised Measure of Hallucination}
\label{sec:measure_of_hallucination}

\begin{figure}
    \centering
    \includegraphics[trim={0.3cm 1.5cm 0cm 0cm},clip,width=\columnwidth]{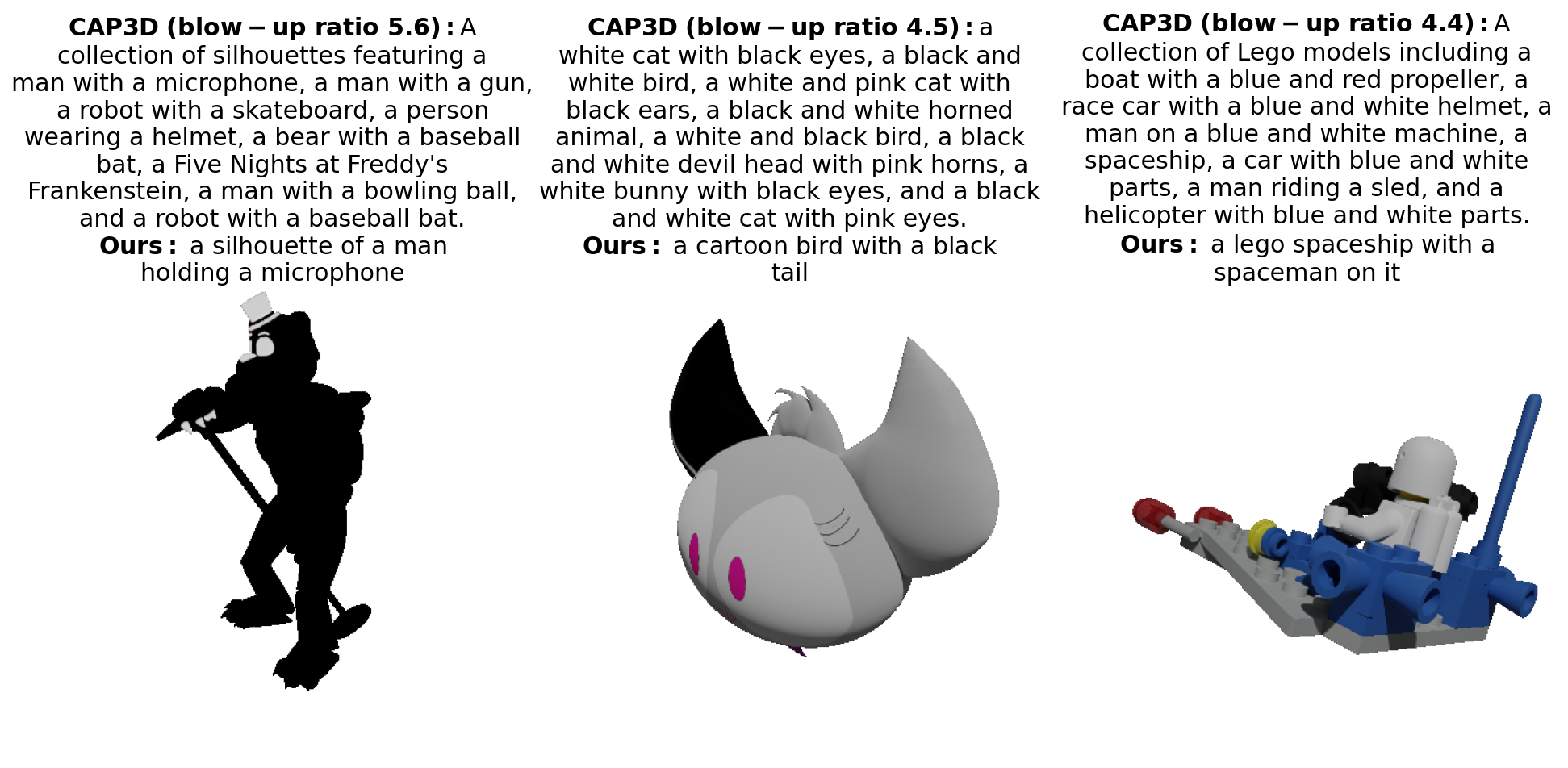}
    \caption{\textbf{Objects with the largest caption blow-up ratio for CAP3D.} We compare their aggregate captions with ours.}
    \vspace{-5pt}
    \label{fig:hallucination_main}
\end{figure}

To explain the performance gap between our approach and CAP3D, we develop a measure to identify cases where CAP3D hallucinates. The goal is to systematically compare those cases with our captions without cherry-picking. We observe that CAP3D's aggregated outputs are often longer than the per view captions, because GPT4 naively adds up descriptions from different views. This blow-up in caption size due to the aggregation step can be measured as follows: \begin{align}
\vspace{-10pt}
\text{blow-up ratio}(r) & \defeq \dfrac{\text{word\-count}(r)}{\max_{i,j} \text{word\-count}(r_{i,j})} \label{eq:hallucination_measure}
\vspace{-5pt}
\end{align}
Eq \ref{eq:hallucination_measure} divides the word count of an output summary by the maximum word count across all single-view captions for the same object. For CAP3D, we find caption blow-up ratios as high as 5.6, implying that GPT4 accumulated the words of at least 5 single-view captions for that particular object. On the other hand, if we computed caption blow-up ratios for ScoreAgg, we would always get a ratio of 1.0, because our final caption is always one of the single-view captions.

We visualize objects with the highest CAP3D blow-up ratios in Fig \ref{fig:hallucination_main} (see also Fig \ref{fig:cap3d_vs_ours_by_hallucination} in the Appendix, where we show more objects). Across the board, our captions are more concise and accurate. We find groups of similar objects emerge when they are ranked, suggesting systematic CAP3D errors (e.g., three ``child's drawings'', two ``silhouette'' figures, and two ``cartoon birds'' in the top 20). While a high blow-up ratio is a precise indicator of hallucination, it has low recall, because even shorter CAP3D summaries can contain contradictions (e.g., ``a banana and a chicken'' in Figure \ref{fig:type_comparison}). We present such examples in Fig \ref{fig:cap3d_vs_ours}. %

\subsection{Explaining ScoreAgg's performance}
\label{sec:why_aggregate}

To show how ScoreAgg works, we examine (i)~the benefit of multi-view/multi-prompt aggregation over individual VLM queries; (ii)~an alternative for the log-sum-exp function in Eq~\ref{eq2}; and (iii)~the effect of hyperparameters $I$ and $J$, which determine the VLM compute budget. We use the same similarity score as Section~\ref{sec:quantitative_type_evaluation} for these analyses.

\begin{figure}[h!]
    \centering
    \begin{subfigure}{\columnwidth}
    \centering
    \includegraphics[trim={0cm 0.3cm 0cm 0cm},clip,width=0.95\textwidth]{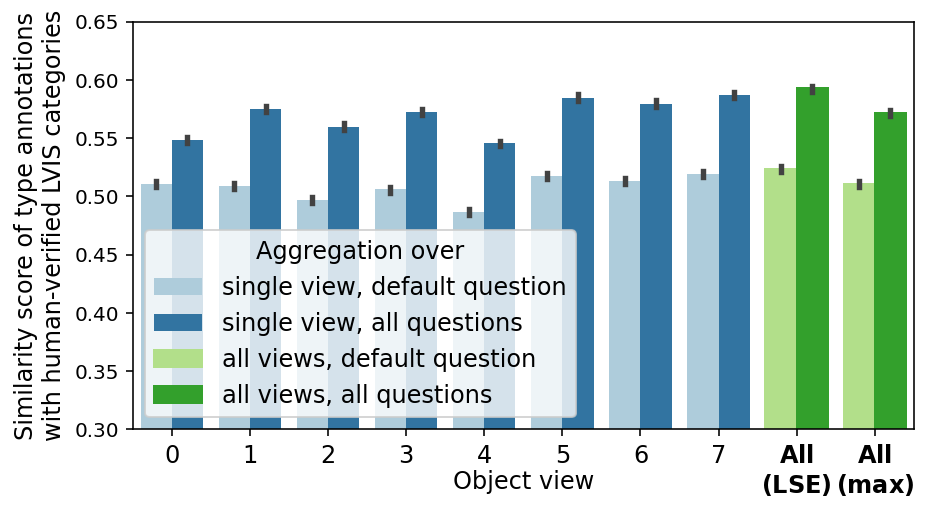}
    \caption{Score-based aggregation across views and across questions. Each bar aggregates a different subset of VLM responses. We compare single-view predictions (labeled 0-7) with aggregated predictions over all views, while highlighting the gap between asking a single or multiple questions.%
    }
    \label{fig:individual_views_and_questions} %
    \end{subfigure}
    
    \begin{subfigure}{0.9\columnwidth}
    \centering
    \includegraphics[width=0.9\textwidth]{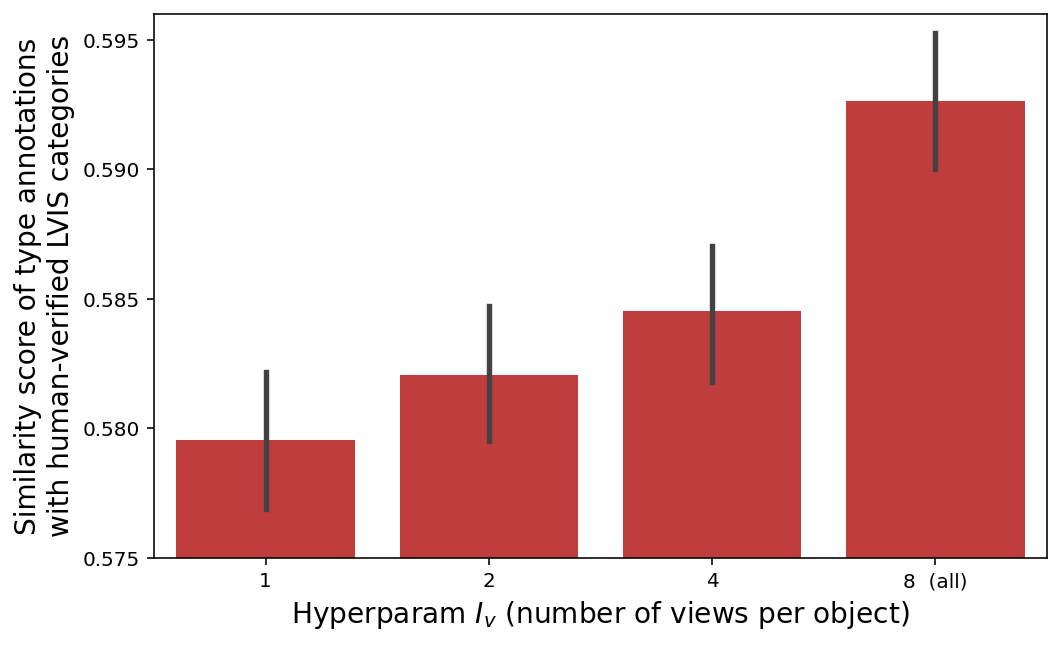}
    \caption{Effect of aggregating across a variable number of object views, $I_v$.}
    \label{fig:increasing_i} %
    \end{subfigure}

    \begin{subfigure}{0.9\columnwidth}
    \centering
    \includegraphics[width=0.9\textwidth]{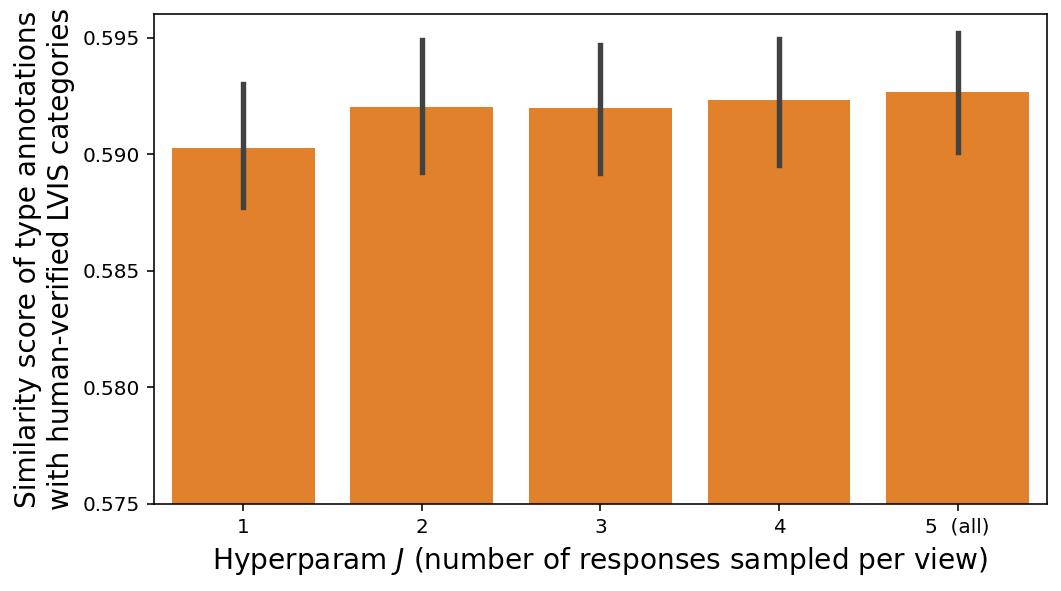}
    \caption{Effect of aggregating across a variable number of VLM responses sampled per probe, $J$.}
    \label{fig:increasing_j} %
    \end{subfigure}
    \caption{\textbf{Explaining ScoreAgg's performance.} We apply ScoreAgg on different subsets of VLM probes ($I_v=8$ object views, $I_q=4$ VQA prompts) and VLM responses per probe (up to $J=5$). Aggregate outputs are scored on Objaverse-LVIS as before.
    }
    \label{fig:views_and_questions} %
\end{figure}

Fig~\ref{fig:individual_views_and_questions} shows the accuracy of individual object views versus all-view ScoreAgg responses. The log-sum-exp (LSE) aggregate performs better than any individual view, underscoring why our captions are more reliable than CAP3D's. Fig~\ref{fig:individual_views_and_questions} also compares the LSE aggregation with the simpler choice of taking the maximum score. The LSE outperforms the max-score aggregation by a small but significant margin---the latter performs worse than several views individually, because overconfident responses might dominate the aggregate. This validates our algorithmic choice.

\begin{figure*}[h]
    \centering
    \includegraphics[width=\textwidth]{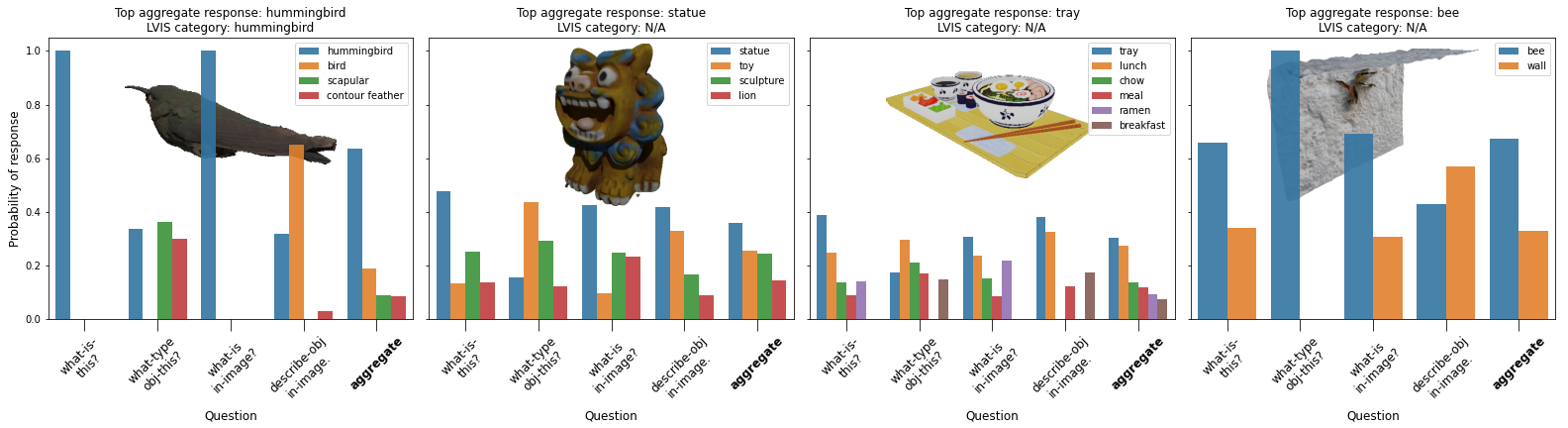}
    \caption{\textbf{PaLI VQA responses per question and after aggregation} for fixed views of a selection of objects. To reduce visual clutter, we filtered responses with scores below a fixed threshold (-1.2). Each subplot legend lists the possible responses sorted by aggregate probabilities. For comparison, we show the object's LVIS category where available.
    }
    \label{fig:type_response_distributions}
\end{figure*}

Prompt-aggregation further boosts the accuracy of type prediction, as we show qualitatively in Fig~\ref{fig:type_response_distributions}. Using multiple questions smoothens the ScoreAgg response distribution and widens the support. We see that prompt-aggregation helps avoid mode collapse in bimodal cases (such as the bee on the wall). It also reduces question-specific biases (e.g., questions that include the word ``object'' make the VLM likelier to say ``toy,'' while remaining questions are likelier to elicit ``statue'' or ``lion.'')

Finally, we examine how ScoreAgg scales with respect to additional VLM computation (which may yield inconsistent responses). Figures~\ref{fig:increasing_i} and \ref{fig:increasing_j} underscore that ScoreAgg benefits from more candidate captions and views. The intuition is that increasing $I$ and $J$ increases the overlap in the response-view score matrix (see Fig \ref{fig:headline}-B2), thus producing a more reliable aggregate score for each candidate caption.

In sum, ScoreAgg is key to improving the accuracy of type annotation. We probe these annotations further on inferring downstream properties in the next section.

\section{Material Inference}
\label{sec:physical_properties}
Our second task is to infer what each Objaverse object is made of. The task is significant because an object's physical composition has immediate implications for how it behaves physically. Whether it will sink, bounce, stretch, or crack is largely determined by its material. There is limited prior work to study whether VLMs can infer material properties \cite{gao2023physically}, mainly due to a lack of validation data.

To address the gap, we collected a test set of 860 objects spanning 58 material classes. All labels are derived from original Objaverse tags (see Appendix \ref{sec:material_test_set} for details). To make the test as challenging as possible, we included:

\begin{enumerate}[leftmargin=0.7cm,label=(\roman*),nolistsep]
    \item classes at different levels of specificity (e.g., `metal' and `aluminum'), because we expect aggregated VLM predictions to place probability mass on both levels (varying based on the VLM's confidence). This is not the same as multi-label classification; we only evaluate on one target label per object.
    \item a wide range of materials with different properties: non-rigid materials (e.g., `tarpaulin', `snow'), organic materials (e.g., `bamboo', `seashell', `bone'), manufactured materials (e.g., `glass', `plastic', `steel'), food ingredients (e.g., `chocolate', `rice'), fabrics (e.g., `wool', `denim'), and natural elements (e.g., `sand', `ice').
\end{enumerate}

\subsection{Evaluation w.r.t. Human Labels}
\label{sec:material_supervised_evaluation}

We devise and evaluate 10 inference scenarios over the following axes. We assume ScoreAgg to generate view-aggregated predictions:
\begin{enumerate}[leftmargin=0.7cm,label=(\roman*),nolistsep]
    \item \textbf{Conditional inference}: We expect an object's material to be harder to infer from images than type; this raises the question whether we can prompt a VLM to reason deeper about material. One way is to equip the VLM with prior inferences such as the object's type. Thus the VLM can make its prediction on the basis of two factors: object type and appearance. We evaluate how well this works by posing questions including or excluding the likeliest type (e.g., ``what material is the spoon made of'' vs ``what material is this made of''). We also run a third probe including the type in the question but without any image input (e.g., ``what material is a spoon made of''). This makes the model operate as an LLM, with the same model weights, and helps measure the accuracy of language-only reasoning. %

    \item \textbf{Choice of type annotations}: When specifying the object's type as part of a question, we can choose to use detailed captions like CAP3D's, or succinct type annotations produced by our VQA pipeline (see Sec \ref{sec:quantitative_type_evaluation}). The comparison is unfair because 43\% of CAP3D captions explicitly contain the target material label, compared to 12\% of PaLI types, mainly due to objects like ``iceberg'' or ``woodcarving''. Nevertheless we study which type annotations perform better. 
    
    \item \textbf{Choice of VLM}: We run two VLMs on all scenarios above–––PaLI-X VQA as before and the smaller BLIP-2 T5 XL (used in CAP3D). This helps ensure our results are not specific to a model class or size.
    
\end{enumerate}

\textbf{Results.} Table \ref{tab:material_prediction} reveals that class-conditional inference can boost material prediction abilities in both VLMs (PaLI-X and BLIP-2). Although the effect is stronger in (the significantly larger) PaLI-X, using a type annotation as well as the object’s appearance generally outperforms using one or the other. We conjecture two possible reasons: (i) fixing a value for an upstream property, from the VLM’s full range of predictions, helps mitigate confusion; and (ii) access to previous computations helps the VLM avoid redundant processing and attend to the downstream task. 

Predictions from text alone (see ``From Type'' subcolumns) confirm that CAP3D captions contain more material information than PaLI-VQA types. Yet PaLI types are on par or better than CAP3D captions when we do use the object’s appearance (see ``From Type and Appearance'' subcolumns). This is likely explained by hallucinations or specious details in CAP3D captions which hinder VLM reasoning. See Table \ref{tab:material_prediction_examples} for more examples of predictions from all inference scenarios across the material test set. %

\begin{table*}[]
\centering
\caption{\textbf{Material inference with two VLMs: PaLI-X and BLIP-2.}. The models are provided either an object type annotation or image as inputs or both. We report top-k accuracies and soft accuracies averaged over the material test set; we also show top-2 predictions on an example object (right) under each inference scenario. VLM-mode responses are view aggregated (via ScoreAgg). The predicted distributions contain up to J=5 alternatives in LLM mode or IJ=40 in VLM mode.}
\scalebox{0.7}{
\begin{tabular}{cl|cc|c|cc|c}
\cline{3-7}
\multicolumn{2}{l|}{\multirow{2}{*}{
\includegraphics[width=3cm]{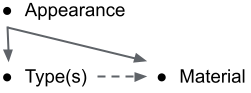}
}}  
& \multicolumn{2}{c|}{\begin{tabular}[c]{@{}c@{}} From Type\\ (LLM mode)\end{tabular}} & \multicolumn{1}{c|}{\begin{tabular}[c]{@{}c@{}}From Appearance\\ (VLM mode)\end{tabular}}  & \multicolumn{2}{c|}{\begin{tabular}[c]{@{}c@{}}From Type and Appearance\\ (VLM mode)\end{tabular}} & \multirow{5}{*}{\shortstack[c]{Example object\\Material label: \hl{``porcelain''}. CAP3D\\caption: ``small white \hl{porcelain}\\ vase with colorful floral designs\\ on it''. PaLI-VQA type: ``inkwell''.}}  \\ \cline{3-7} 
\multicolumn{2}{l|}{} & \multicolumn{1}{c|}{\begin{tabular}[c]{@{}c@{}}CAP3D\\ captions\end{tabular}} & \multicolumn{1}{c|}{\begin{tabular}[c]{@{}c@{}}PaLI-VQA\\ types\end{tabular}} & \multicolumn{1}{c|}{\begin{tabular}[c]{@{}c@{}}No caption/type\\ information\end{tabular}} & \multicolumn{1}{c|}{\begin{tabular}[c]{@{}c@{}}CAP3D\\ captions\end{tabular}} & \multicolumn{1}{c|}{\begin{tabular}[c]{@{}c@{}}PaLI-VQA\\ types\end{tabular}} & \\ \cline{1-7}

\multicolumn{1}{|c|}{\multirow{5}{*}{\shortstack{PaLI-X\\55B\\VQA}}}   & Top-1 acc. & \multicolumn{1}{c|}{$0.46 \pm 0.50$} & $0.33 \pm 0.47$ & $0.56 \pm 0.50$  & \multicolumn{1}{c|}{$\mathbf{0.61 \pm 0.49}$} & $0.60 \pm 0.49$ & \\ \cline{2-7} 
 \multicolumn{1}{|c|}{} & Top-3 acc. & \multicolumn{1}{c|}{$0.73 \pm 0.44$} & $0.58 \pm 0.49$    & $0.83 \pm 0.37$      & \multicolumn{1}{c|}{$\mathbf{0.87 \pm 0.34}$} & $0.86 \pm 0.35$   & \\ \cline{2-7} 
\multicolumn{1}{|c|}{} & Soft acc.  & \multicolumn{1}{c|}{$0.36 \pm 0.29$} & $0.25 \pm 0.28$    & $0.41 \pm 0.28$      & \multicolumn{1}{c|}{$\mathbf{0.44 \pm 0.27}$} & $\mathbf{0.44 \pm 0.29}$ & \multirow{7}{*}{\includegraphics[trim={2.5cm 0cm 2.5cm 2.5cm},clip,width=0.18\linewidth]{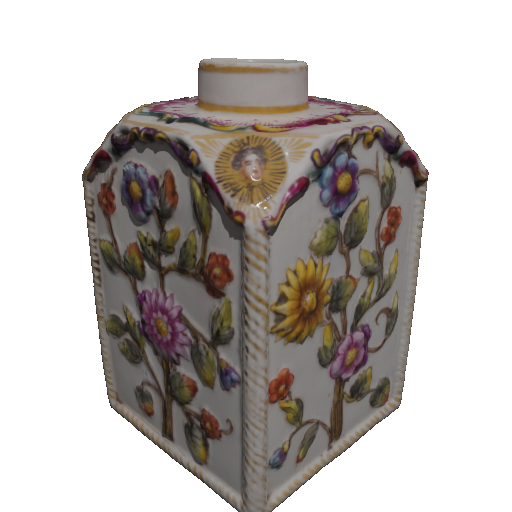}} \\ \cline{2-7} 
\multicolumn{1}{|c|}{} & \multirow{2}{*}{\shortstack[l]{Example\\preds.}}  
    & \multicolumn{1}{c|}{\multirow{2}{*}{\shortstack{\hl{``porcelain''} (0.29),\\``faience'' (0.28)}}} 
    &  \multirow{2}{*}{\shortstack{``glass'' (0.28),\\ \hl{``porcelain''} (0.24)}} 
    & \multirow{2}{*}{\shortstack{``faience'' (0.88),\\ \hl{``porcelain''} (0.06)}} 
    & \multicolumn{1}{c|}{\multirow{2}{*}{\shortstack{``faience'' (0.68),\\ \hl{``porcelain''} (0.15)}}}
    & \multirow{2}{*}{\shortstack{``faience'' (0.71),\\ \hl{``porcelain''} (0.16)}} & \\
 \multicolumn{1}{|c|}{} & & \multicolumn{1}{c|}{} & & & \multicolumn{1}{c|}{} & \\ \cline{1-7}

\multicolumn{1}{|c|}{\multirow{6}{*}{\shortstack{BLIP-2\\T5 XL}}} & Top-1 acc. & \multicolumn{1}{c|}{$0.21 \pm 0.41$} & $0.18 \pm 0.39$ & $\mathbf{0.57 \pm 0.50}$& \multicolumn{1}{c|}{$0.47 \pm 0.50$} & $0.56 \pm 0.50$  \\ \cline{2-7} 

 \multicolumn{1}{|c|}{} & Top-3 acc. & \multicolumn{1}{c|}{$0.24 \pm 0.43$} & $0.22 \pm 0.41$    & $0.68 \pm 0.47$      & \multicolumn{1}{c|}{$0.59 \pm 0.49$} & $\mathbf{0.69 \pm 0.46}$    \\ \cline{2-7} 
\multicolumn{1}{|c|}{} & Soft acc.  & \multicolumn{1}{c|}{$0.19 \pm 0.35$} & $0.16 \pm 0.33$    & $0.50 \pm 0.41$      & \multicolumn{1}{c|}{$0.42 \pm 0.42$} & $\mathbf{0.51 \pm 0.42}$    \\ \cline{2-7} 
\multicolumn{1}{|c|}{} & \multirow{3}{*}{\shortstack[l]{Example\\preds.}}  
    & \multicolumn{1}{c|}{\multirow{3}{*}{\shortstack{``China'' (0.58),\\``ceramic'' (0.24)}}}
    &  \multirow{3}{*}{\shortstack{``metal'' (0.93),\\``metal or\\plastic'' (0.06)}}
    & \multirow{3}{*}{\shortstack{\hl{``porcelain''} (0.99),\\``white\\porcelain'' (0.01)}}
    & \multicolumn{1}{c|}{\multirow{3}{*}{\shortstack{\hl{``porcelain''} (0.80),\\``china'' (0.12)}}}
    & \multirow{3}{*}{\shortstack{\hl{``porcelain''} (0.94),\\``china'' (0.04)}} \\
 \multicolumn{1}{|c|}{} & & \multicolumn{1}{c|}{} & & & \multicolumn{1}{c|}{} & \\ 
 \multicolumn{1}{|c|}{} & & \multicolumn{1}{c|}{} & & & \multicolumn{1}{c|}{} & \\ \cline{1-7}

\end{tabular}
}
\label{tab:material_prediction}
\end{table*}

\subsection{An Unsupervised Metric for Downstream Inference}
\label{sec:unsupervised_properties}

We now probe our material annotations through an unsupervised lens. The motivation is---while VLMs can perform arbitrary question answering tasks, collecting labeled data to validate their responses remains a bottleneck. What we do have are some intermediate inferences which are already validated, and can be used for conditional inference. Using these, we develop a metric to characterize the quality of downstream VLM responses, and to highlight interesting or problematic cases without supervision.

As in Sec~\ref{sec:material_supervised_evaluation}, we run VLMs with and without a visual input by supplying a text value $z$ for the type property in both cases. The question $q_z$ changes slightly to be type-specific in LLM mode, $q'_z$, rather than instance-specific in VLM mode. Let $\{r_{i,j}\}$ and $\{r'_{j}\}$ respectively denote the sampled responses (where $i$ indexes object views and $j$ indexes VLM responses for a fixed query).
We compare outputs from running in VLM and LLM modes using a probability metric, the Hellinger distance $H$, which for discrete distributions (like the output of ScoreAgg) is identical to the Euclidean distance between two square root probability vectors: \begin{align}
\label{eq:visual_sensitivity}
\vspace{-5pt}
\text{visual sensitivity}(q_z) \defeq H( \tilde{p}(r | \{r_{i, j}\}), \tilde{p}(r | \{r'_{j}\} ))
\vspace{-5pt}
\end{align}

\begin{figure}[h!]
    \centering
    \begin{subfigure}{0.9\columnwidth}
    \includegraphics[width=0.95\textwidth]{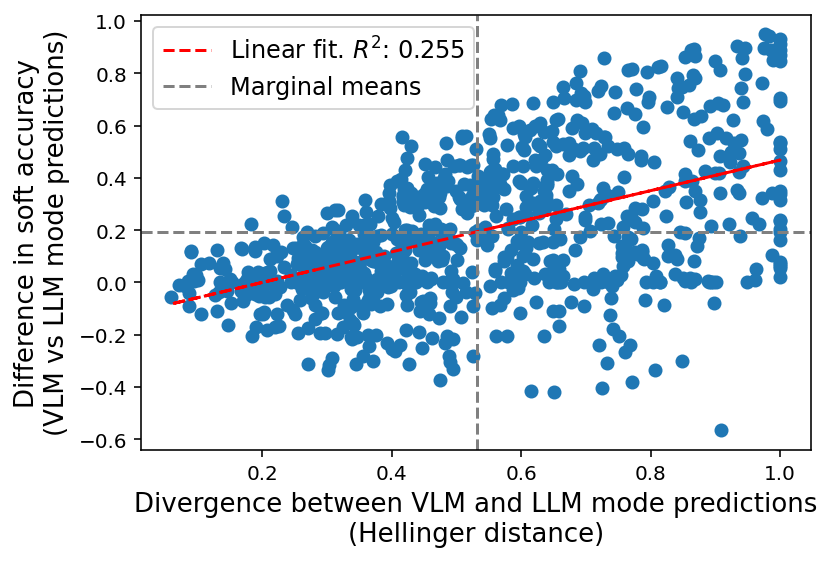}
    \caption{Gains in material prediction accuracy when the VLM diverges from underlying language-only predictions. We show all datapoints from the material validation set, with a linear fit to highlight the correlation.}
    \label{fig:material_accuracy_vs_distance} %
    \end{subfigure}
    
    \begin{subfigure}{0.98\columnwidth}
    \includegraphics[trim={0cm 0.3cm 0cm 0cm},clip,width=\textwidth]{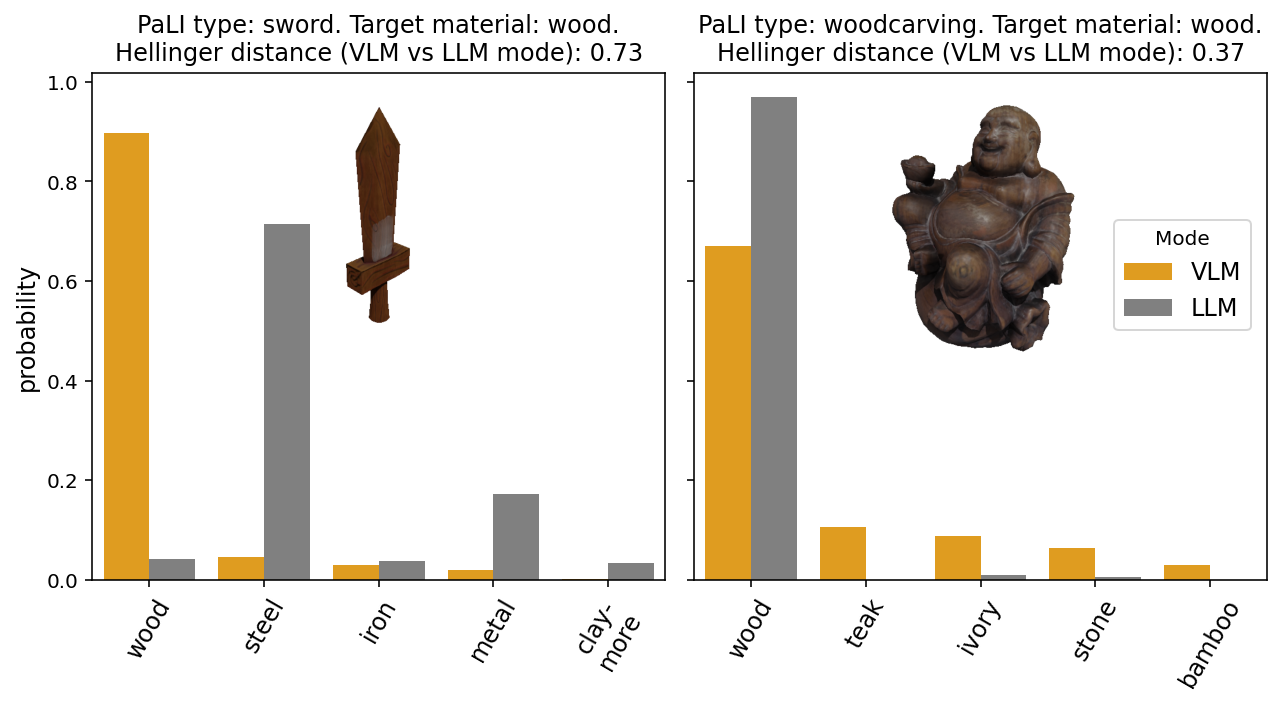}
    \caption{Material predictions. We show outputs over 2\% prob.
    }
    \label{fig:material_qualitative_samples} %
    \end{subfigure}

    \begin{subfigure}{\columnwidth}
    \includegraphics[trim={0cm 0.3cm 0cm 0.2cm},clip,width=\textwidth]{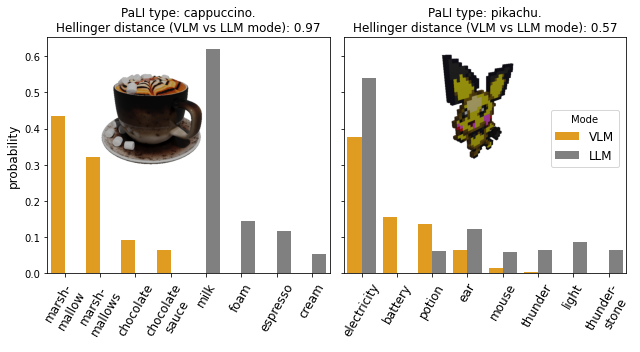}
    \caption{Containment predictions. We show outputs over 5\% prob.
    }
    \label{fig:containment_qualitative_samples}
    \end{subfigure}

    \caption{\textbf{Visual sensitivity measured as the divergence between VLM and LLM mode predictions.} VLM mode results are view-aggregated using ScoreAgg. %
    }
\end{figure}

While such a distance can be computed for any VLM in any unsupervised case, the question that arises is whether the distance correlates with predictive performance. Since the contribution of a VLM's visual branch is generally residual (via cross-attention from the language branch), we posit that when answers differ between the visionless and vision-based conditions, the latter is likely more accurate. We assess this hypothesis in Figure \ref{fig:material_accuracy_vs_distance} using PaLI-X predictions on the labeled material test set (from Section~\ref{sec:material_supervised_evaluation}). We find significant correlation between the visual sensitivity metric and gains in (supervised) soft accuracy. The correlation is likely underestimated due to noise in the material labels---the vision-based material predictions are often justifiably multi-modal (see Figure \ref{fig:material_qualitative_samples}) whereas our labels are one-hot.

Having shown its usefulness on material, we compute our visual sensitivity metric when querying for other object properties (Table \ref{tab:unsupervised_metric}). We infer (i) binary properties such as fragility or lift-ability, (ii) open-vocabulary properties such as color and affordance, and (iii) relations between objects such as what a given object might contain. We find that the standard deviation of Hellinger distances for any given question is indicative of the size of the output space (e.g., binary-response questions have the lowest spread). We also find that the mean of Hellinger distances grows as expected with the difficulty of answering questions in visionless mode (e.g., color benefits the most from VLM mode). Lastly, we find that providing more information in the question (both material and type rather than type alone) consistently reduces the gap (i.e., mean Hellinger distance) between VLM and LLM mode responses.

Our analysis is useful not only to compare questions in aggregate, but also to highlight individual cases which merit attention. For instance (in Figures \ref{fig:material_qualitative_samples} \& \ref{fig:containment_qualitative_samples}), it flags the atypical material of a wooden sword or unexpected objects contained in a cappuccino (marshmellows), which were evident only from visual context. Thus, the analysis could be instrumental in scaling VLM annotation to unsupervised cases beyond the scope of human-powered validation.

\newcommand{\typePlaceholder}{T}
\newcommand{\materialPlaceholder}{M}

\begin{table}[t!]
\centering
\caption{\textbf{Object properties assessed without validation, via visual ablation of PaLI-X responses.} Placeholders \typePlaceholder\ and \materialPlaceholder\ are filled in with a prior type or material inference. The visual sensitivity metric is averaged over 44K objects. (See also Fig \ref{fig:all_properties_fixed_objects} where we showcase these predictions qualitatively on a fixed set of objects.)
}
\scalebox{0.85}{
    \begin{tabular}{|p{0.14\columnwidth}|p{0.49\columnwidth}|p{0.23\columnwidth}|}
    \hline
    Question type & Question in LLM/VLM mode & Hellinger vis. sensitivity \\ \hline
    \multirow{2}{0.14\columnwidth}{Fragility}         
        & Is a/this \typePlaceholder\ fragile?  & $0.110 \pm 0.058$ \\ 
        & Is a/this \materialPlaceholder\ \typePlaceholder\ fragile?   & $0.103 \pm 0.054$ \\ 
        \hline
    \multirow{2}{0.14\columnwidth}{Lift-ability}
        & Can a human lift a/this \typePlaceholder?  & $0.133 \pm 0.063$ \\
        & Can a human lift a/this \materialPlaceholder\ \typePlaceholder? & $0.124 \pm 0.055$    \\
        \hline
    Afford-ance & How is a/the \typePlaceholder\ typically used? & $0.575 \pm 0.223$ \\ \hline
    \multirow{3}{0.14\columnwidth}{Contain-ment} 
        & What might a/the \typePlaceholder\ contain? & $0.690 \pm 0.260$\\
        & What is something that typically goes into a/the \typePlaceholder? & $0.570 \pm 0.235$ \\
        & What items or substance might a/the \typePlaceholder\ contain?        & $0.731 \pm 0.236$ \\
        \hline  
    \multirow{2}{0.14\columnwidth}{Color} 
        & What color is a/this \typePlaceholder? & $0.894 \pm 0.163$ \\
        & What color is a/this \materialPlaceholder\ \typePlaceholder? & $0.875 \pm 0.173$ \\
        \hline  
    \end{tabular}
}\vspace{-10pt}
\label{tab:unsupervised_metric}
\end{table}

\section{Conclusions}
\label{sec:conclusions}

We explored the design space of VLM pipelines for captioning and open-vocabulary classification. We evaluated what VLMs are sensitive to, including changes in object view, question wording, prior inferences specified in the prompt, and access to the object's appearance. We introduced an algorithm (ScoreAgg) to marginalize over some of these factors---akin to how humans might arrive at an inference by examining an object from multiple angles. If the aggregation is not visually grounded, we showed it hallucinates using a simple indicator. 

Score-aggregated annotations serve as reliable representations for downstream inference. Our unsupervised visual sensitivity metric helps address the bottleneck of validation data in scaling VLM pipelines. Our outputs for Objaverse are available on our project page, and promise to serve a variety of applications (from retrieval to 3D generation and physical simulation). We hope our evaluations and insights also help shape VLM annotation pipelines in other contexts.

\subsection{Limitations and Future Work} 

ScoreAgg's reliability is based on using numerous overlapping queries. When there's disjoint coverage across queries, it will likely drop infrequent details to produce a globally consistent caption. If we applied it to non-overlapping photos of a room (e.g., taken from the center) we can only expect a high-level caption rather than a combination of details across views. This limitation could be mitigated by increasing the field of view to ensure overlap (e.g., taking photos from the boundary of the room). We saw in our usage that ScoreAgg benefits from aggregating over more VLM queries (Figure~\ref{fig:increasing_i}). Hence, extra VLM compute could be deployed to boost query overlap and final accuracy.

Another limitation concerns the aggregation function in Eq~\ref{eq2}. The log-sum-exp (LSE) helps balance contradictory signals and produce a globally reliable response. %
It would need to be modified if the goal was different---say if we wanted to detect whether a certain feature occurs in any view of an object. In such cases, it might be appropriate to replace the LSE function with a max.

Future generations of pretrained models might include VLMs capable of processing multi-view images simultaneously to produce a global response. While this would be an interesting direction, there is still value in approaches like ours: the ScoreAgg algorithm is transparent and helps mitigate black-box hallucination. %

Before this work can be applied to object identification using arbitrary images (e.g., from a mobile camera), it would be useful to develop a measure of the marginal value of specific new views or object complexity as a whole. Nonetheless, we hope our pipeline optimization approach will transfer to scene understanding beyond digital 3D objects.

\section*{Impact Statement}
Our work allows using pretrained VLMs more accurately for annotation tasks. It may reduce the need for compute resources spent on fine-tuning such models, a positive societal consequence. On the flip side:

While increasing the reliability of our annotations was a key focus for us, any automated captioning system will exhibit biases. Given the scale of the annotations we produce (350 million open-vocabulary responses), they may contain all kinds of content.

The concern is heightened from our use of Objaverse---a relatively new, under-explored set of object assets. They were originally created and uploaded by 100K artists to the \href{https://sketchfab.com/}{Sketchfab platform}. Given their diverse origin, the assets may contain all kinds of content. Our annotations (for 764K objects) will verbalize and reflect what is in the assets. 

We include some of our annotations in the supplementary materials to help reviewers assess them (see Appendix \ref{sec:outline} for an outline). We ran the outputs through the \href{https://perspectiveapi.com/}{Perspective API} to measure toxicity. Although we did not find any concerning examples, the API may have missed them.

If our annotations do get popularly used, we would caution users and highlight the need to review the captions. This could be undertaken in a community-driven way, but remains impossible for individuals. 

\ifdefined\CAMERAREADY
\section*{Acknowledgements}
We are grateful to Murray Shanahan, Drew Hudson, Jovana Mitrović, Martin Engelcke, and Radu Soricut for their comments and assistance.

\fi

%% file: X_suppl.tex
\clearpage
\appendix

\section{Outline}
\label{sec:outline}

The Appendix is structured as follows---we provide model and dataset details in Sections \ref{sec:model_details} and \ref{sec:dataset_details} respectively. We present a literature review in Sec~\ref{sec:literature_review}. Then we present the following extended results:
\begin{enumerate}[nolistsep]
    \item Section~\ref{sec:appearance}: VLM sensitivity to image and lighting conditions or object texture.
    \item Section~\ref{sec:extra_hallucination_examples}: statistics, word clouds, and more examples comparing CAP3D captions with ours (including two objects that the CAP3D paper highlighted as failure cases).
    \item Section~\ref{sec:extended_results_material_prediction}: predictions from all material inference scenarios on 24 example objects.
    \item Section~\ref{sec:qualitative_assessment}: probes on a fixed set of objects to draw qualitative insights on what factors affect VLM predictions for a series of properties.
\end{enumerate}

\section{Model Details}
\label{sec:model_details}

We used the following VLMs off the shelf without tweaking:

\subsection{PaLI-X} 
The model \cite{chen2023palix} is based on the \href{https://github.com/google/flaxformer}{flaxformer} transformer \cite{vaswani2017attention} library and \href{https://github.com/google-research/t5x}{t5x} training/evaluation infrastructure \cite{roberts2022t5x}, both written and released in jax \cite{frostig2018compiling}. The captioning- and VQA-tuned variants have a common architecture. Checkpoints and configs for the 20B UL2 \cite{tay2022unifying} language backbone are separately available \href{https://github.com/google-research/google-research/tree/master/ul2}{here}. The language backbone relies on a SentencePiece tokenizer \cite{kudo2018sentencepiece} with vocab size 250K available \href{https://github.com/google/sentencepiece}{here}. The visual backbone for PaLI-X (a ViT-22B \cite{dehghani2023scaling}) includes an additional OCR-based classification pretraining task on WebLI images \cite{chen2023pali} beyond the original JFT-3B \cite{sun2017revisiting,zhai2022scaling} image classification task. 

During VLM training, the captioning and VQA variants diverge in their image resolutions ($672^2$ versus $756^2$) and training task mixtures. While the VQA variant was partly trained using the Object-Aware method \cite{piergiovanni2022pre} to detect or list object classes on the OpenImages V4 dataset \cite{kuznetsova2020open}, we are not aware of any other reason PaLI-X would be predisposed to predict object labels accurately, especially on the long-tailed distribution of Objaverse.

PaLI scoring is length normalized as originally described in Eq 14 of \cite{wu2016google} or coded in t5x \href{https://github.com/google-research/t5x/blob/main/t5x/decoding.py#L693}{here}. This is to help ensure that longer outputs are not disadvantaged. We kept the length norm parameter fixed at $\alpha=0.6$ and used default beam searching sampling with 5 parallel decodings.

\subsection{BLIP-2} 
As CAP3D did, we use BLIP-2 from \href{https://github.com/salesforce/LAVIS}{LAVIS} \cite{li2022lavis}, which is based on the widely used PyTorch \href{https://huggingface.co/docs/transformers/index}{transformers} library. The model is appealing because it was shown to perform better than 54x larger models on VQA and image captioning. Its FlanT5 encoder-decoder backbone \cite{chung2022scaling} was pretrained using the span corruption objective (introduced in T5 \cite{raffel2020exploring}), then instruction-finetuned for various language-based question answering. The image encoder is a ViT-g/14 model, as trained by EVA-CLIP \cite{fang2023eva}.

Unlike PaLI-X components, BLIP-2's ViT was directly evaluated for object detection and instance segmentation on LVISv1.0 \cite{gupta2019lvis}. The authors highlighted emergent capabilities on object-level instance segmentation; their ViT classified 1200 LVIS categories as accurately as the 80 COCO categories on which it was trained \cite{fang2023eva}. We expect BLIP-2 to exhibit some transfer to Objaverse-LVIS. So it is surprising that PaLI-X outperforms BLIP-2 nonetheless. Besides model size, a reason for the performance gap could be that BLIP-2 operates with images of size $224^2$.

We tweaked BLIP-2's code slightly to (i)~generate outputs with accompanying scores, using existing functionality in the transformers library; and (ii)~optionally run in LLM mode, by removing visual tokens from inputs to the transformer. We attach the diff for these changes, totaling 20 lines, in \path{blip2-code_diff.txt}.

\section{Dataset Details}
\label{sec:dataset_details}

\subsection{Objaverse Rendering}

We downloaded 798,759 Objaverse GLB files and rendered them using Blender $3.4$ \cite{blender}. We dropped animated objects while rendering (to avoid misrepresenting them by generating static-object captions). This produced 763,844 objects rendered from 8 different views, including 44,199 with LVIS categories.

\textbf{Rendering process.} We placed each object at the origin and scaled its maximum dimension to 1. We then rotated the camera at a fixed height and distance to the origin, rendering images at azimuthal intervals of 45 degrees. To determine the camera height, we swept over a few values of the polar angle $\theta$ w.r.t. the z-axis. We presented this sweep and other rendering hyperparameters (such as lighting conditions) in Table \ref{tab:appearances} in the main text.

\textbf{Reproducing CAP3D views.} CAP3D uses a distinct image rendering pipeline. While we render images at a fixed camera height, CAP3D images include top-down and bottom views of the object. Although we did not have access to their original images, we compared rendered images and approximated their camera poses from Fig 23 in the CAP3D paper. By default, only views 1, 3, 5, and 7 (zero-indexed) from our pipeline are comparable to 4, 3, 5, and 2 from the CAP3D pipeline. We remap CAP3D captions to align with our view indices, and focus on the compatible viewpoints when comparing captions.

\subsection{Material Test Set}
\label{sec:material_test_set}

\begin{figure*}[h]
    \centering
    \includegraphics[width=\textwidth]{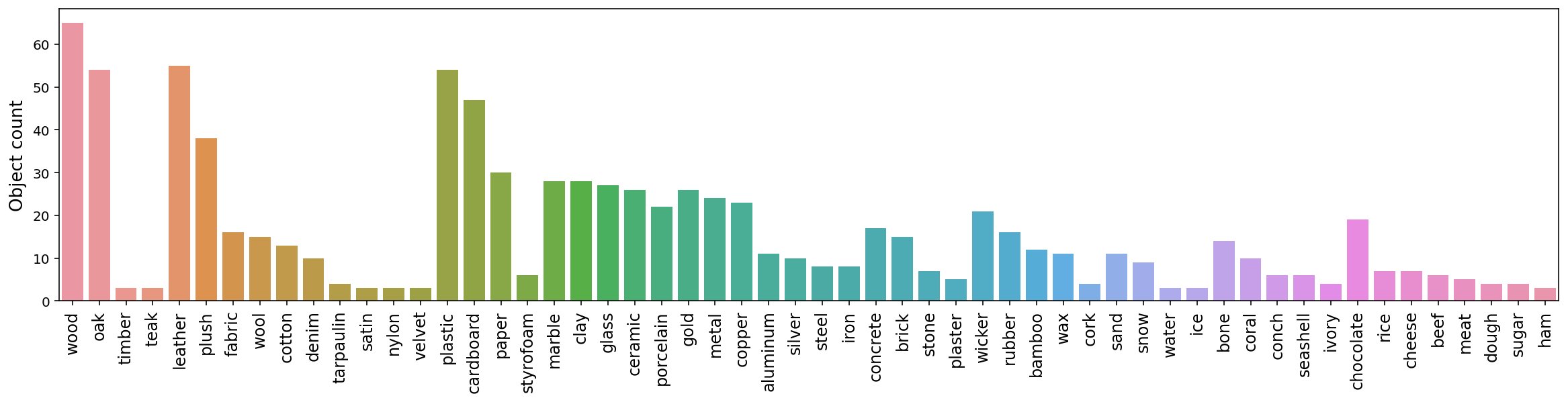}
    \caption{\textbf{Our long-tailed material test set comprising 58 material classes (on the x-axis) and 860 objects.}}
    \label{fig:material_test_set_histogram} %
\end{figure*}

We took inspiration from the long-tailed distribution of LVIS categories \cite{gupta2019lvis} to enumerate material classes for our test set. To make the test as difficult as possible, we included a range of materials with different properties. We included labels at different levels of specificity. We also included labels which are easily conflated (e.g., `coral', `seashell', and `conch').

\textbf{Recipe.} Rather than label objects on our own, we searched through object tags from Objaverse for a superset of 122 material labels. 
The initial matches were noisy---the tags often contain spurious materials (because artists can add arbitrary tags to optimize search engine visibility for their objects). 
So we used a custom web app to accept or reject object-label pairs from the initial matches. This helped preserve the natural, long-tailed distribution of material tags in the dataset (that also reflects the real world). 
We dropped query terms which returned too few/poor quality matches\footnote{Full list of material tags we searched for, but dropped due to insufficient, ambiguous, or poor quality matches: `alcohol',
 `alloy',
 `aluminium foil',
 `ash',
 `ashes',
 `buckram',
 `canvas',
 `carbon fiber',
 `carbon fibre',
 `carbon-fiber',
 `carbon-fibre',
 `cashmere',
 `cellulose',
 `chenille',
 `chitin',
 `cloth',
 `corduroy',
 `corn',
 `egg',
 `eggs',
 `eggshell',
 `feather',
 `feathers',
 `felt',
 `fiberglass',
 `flax',
 `flour',
 `flowers',
 `fur',
 `granite',
 `graphite',
 `grass',
 `knitwear',
 `lace',
 `laminate',
 `leaf',
 `leaves',
 `limestone',
 `mahogany',
 `milk',
 `oil',
 `papyrus',
 `parchment',
 `pasta',
 `pewter',
 `platinum',
 `ply',
 `plyboard',
 `pvc',
 `pyrex',
 `resin',
 `suede',
 `shell',
 `silk',
 `slate',
 `tartan',
 `teflon',
 `textile',
 `titanium',
 `veggie',
 `veggies'.}.
In some cases, we had to resolve multiple tag matches per object. Although this was an opportunity to test multi-label prediction (which our aggregate VLM predictions do facilitate), for now we chose to focus on the primary/dominant material of each object. 
Finally, we merged some near-duplicate labels that we included initially to increase matches\footnote{Full list of materials we merged into other labels: `metallic', `woollen', `aluminium', `tarp', `shell'.}.

The final test set contains 860 objects spanning 58 classes, with at least 3 objects per class, as shown in Figure \ref{fig:material_test_set_histogram}. See the attached file \path{material_test_set_objects.pdf} for images of all objects in the test set. 

\textbf{Evaluation metric.} Unlike in Sec \ref{sec:quantitative_type_evaluation}, we cannot rely on similarity in text embedding space because materials can appear close even when they are different (e.g. "wood" and "metal" have a cosine similarity score of 0.408). So we look for exact string matches in the VLM responses.

See Appendix \ref{sec:extended_results_material_prediction} for examples of PaLI-X and BLIP-2 predictions.

\section{Literature Review}
\label{sec:literature_review}

\subsection{Classic object recognition} 

\citet{besl1985three} wrote an influential survey scoping the field of 3D object recognition. They defined the task as follows: given a ``list of distinguishable objects'' and some ``digitized sensor data corresponding to one particular, but arbitrary, field of view'', to be able to answer whether any object appears in the sensor data (and infer its location, count, and 3D orientation). They assumed either range or intensity data (i.e., depth or RGB). The methods they surveyed primarily relied on feature extraction to characterize objects (e.g., dimensions, extreme points, and spatial relationships between object components), surfaces (e.g., curvature), or edges (e.g., parallel edge relationships). 

Even modern methods tend to assume/retrieve a ``list of distinguishable objects''---for instance, \citet{zareian2021open} explored novel object detection using sparse bounding box annotations, but relied on extensive image-caption data. Our approach does not require any specified object classes---we emphasize the ability to work with novel objects. (That said, one might argue that our list of distinguishable objects is baked into the VLM's pretrained knowledge.) 

Our task is also different from the classical definition in our emphasis on multi-view consistency. Compared to classical methods such as feature extraction and graphical representation, using a VLM does seem very black-box; but it is much less black-box than end-to-end learning, which would process all inputs (e.g., multi-view images) together.

\subsection{Dense object-centric tasks}

Given real-world visual inputs, the task typically shifts to segmentation or bounding box detection. \citet{he2021deep} survey a variety of deep learning approaches for 3D segmentation from point clouds, voxels, meshes, or RGB-D. A recent work \cite{koo2022partglot} showed that part segmentation emerged using human text annotations to discriminate between related shapes; the work is evocative of image-text contrastive learning which powers VLM training. Even in 2D, object detection and tracking is an important problem for applications in robotics, augmented reality, or autonomous driving \cite{yao2020video}. These applications typically benefit from slow changes in the view, object motion relative to the scene, or availability of camera parameters.

Our task is salient in that it isolates an object and attempts to generate annotations that are consistent w.r.t. all views of the object. Our focus on global annotation is different from the local processing/dense spatial emphasis of segmentation, tracking, or bounding box detection.

\subsection{Language-based 3D tasks} 

Captioning or question answering in 3D are nascent but growing fields. A substantial amount of work involves navigating environments to answer questions \cite{gordon2018iqa} or to find specified objects (e.g., ObjectNav \cite{batra2020objectnav}). For observational settings, ScanQA \cite{azuma2022scanqa} is one dataset which tests object understanding via question answering. But the dataset is limited to 800 rooms and 41k questions. It assumes RGB-D or point cloud inputs rather than multi-view images.

Captioning 3D objects or scenes is also infrequently explored. One approach \cite{han2020shapecaptioner} to caption 3D shapes detected parts of an object across multiple views, then translated a sequence of view-aggregated part features into a caption using an RNN. \citet{chen2023end} proposed an approach for dense captioning that aims ``to generate multiple captions per scene localized with their associated object regions.'' While there's extensive work on 2D captioning using scene graphs \cite{yang2019auto, Chen_2020_CVPR}, and we have also seen methods for 3D scene graph inference \cite{armeni20193d, 3DSSG2020}, the combination has not been explored for 3D captioning to our knowledge. 

\textbf{Pretrained large models:} With the advent of LLMs and VLMs \cite{li2019visualbert,radford2021learning,jia2021scaling,li2022blip,alayrac2022flamingo,chen2023pali}, it became possible to derive interpretable solutions for arbitrary image-processing tasks. VISPROG \cite{Gupta_2023_CVPR} used in-context learning to produce Python code to invoke off-the-shelf vision models and image processing APIs. Their method worked well for 2D question answering. ViperGPT \cite{suris2023vipergpt} also showed gains in reasoning at the level of object attributes by decomposing queries into executable image-processing subroutines. 

But the use of foundation models remains limited in 3D domains. \citet{hong20233d} is one work which contends with question answering and captioning in 3D. They propose to train 3D VLMs by projecting 3D feature maps to 2D and bootstrapping from a pretrained 2D VLM. Their focus on training LLMs to ingest  point clouds as inputs is orthogonal to our focus on pushing the performance of pretrained models without fine-tuning.

In conclusion, applying pretrained models to 3D domains remains under-explored. CAP3D \cite{luo2023scalable} was the only suitable baseline for our work.

\section{Extended Results}
\label{sec:extended_results}

\begin{table*}[h]
\centering
\caption{\textbf{CAP3D captions versus ours, compared on the frequency of satisfying undesirable criteria ($\downarrow$).} We use the aggregate, post-processed captions in each case---CAP3D drops prefixes like "3D model of," whereas we remove suffixes like "on a white background" from PaLI captions using Eq \ref{eq1}. We count captions that meet the listed criteria, then normalize by the full size of Objaverse (798,759). We use case-insensitive keyword searches unless marked with an asterisk *.%
}
\begin{tabular}{|l|c|l|}
\cline{1-3}
Captions that... & Cap3D & \multicolumn{1}{c|}{PaLI captions}                                                 \\ \hline
\multicolumn{1}{|l|}{Are missing/empty} & 17.17\% & \textbf{4.37\%} \\ \hline
\multicolumn{1}{|l|}{\begin{tabular}[c]{@{}l@{}}\emph{Contain undesirable keywords:}\\ "object"\\ "model" / "3D"\\ "royalty-free" / "royalty free"\\ "download" / "sale"\\ * "OBJ" / "FBX" / "C4D" / "Blender" / "Maya" (but not "Mayan")\end{tabular}} & \begin{tabular}[c]{@{}c@{}}\\\textbf{2.95\%}\\ 2.90\%\\ 0.42\%\\ 0.19\%\\ 0.14\%\end{tabular} & \begin{tabular}[c]{@{}l@{}}\\9.45\%\\ \textbf{0.89\%}\\ \textbf{0.00\%}\\ \textbf{0.00\%}\\ \textbf{0.00\%}\end{tabular} \\ \hline
\multicolumn{1}{|l|}{Start with a word other than "a" / "an" / "the" / "two"}                                                                                                                                                                & 32.59\%                                                                            & 5.79\%                                                                             \\ \hline
\multicolumn{1}{|l|}{End with "background" or "background."} & 0.45\% & \textbf{0.15\%} \\ \hline
\end{tabular}
\label{tab:keyword_caption_comparison}
\end{table*}

\subsection{Ablating Image and Lighting Conditions}
\label{sec:appearance}

So far we have assumed a given set of images per object. That is a limiting assumption for user-driven, real-time object annotation. In this section we explore VLM sensitivity to image and camera settings (such as lighting, poses) as well as changes to the object's appearance (such as untextured rendering). See Table \ref{tab:appearances}.

First, we are interested in how much VLMs and their underlying ViTs rely on texture to recognize objects versus shape \cite{NEURIPS2020_db5f9f42, NEURIPS2021_c404a5ad, dehghani2023scaling}. We render all objects without colors and lighting (using Blender's Workbench engine) to compare with the default appearances. The untextured images do hurt PaLI's type prediction performance but the effect is small enough to suggest that PaLI is largely shape-driven.

Another reason for the reduced impact of untextured rendering is that a number of Objaverse models are already untextured. We suspect that lighting conditions will make a difference in recognizing these objects, because they are prone to losing detail with brightness. We render all objects with three different scene lighting choices: (i) eight area lights and one ceiling light surrounding the object, (ii) a stationary light placed higher than the camera's initial position to enhance shadows, (iii) a moving backlight that follows the camera as we move it. (i) and (iii) ensure symmetry of lighting across the eight views we render, whereas (ii) makes an object look darker from ``behind.'' We find these intuitions do translate to slight performance gains in recognizing objects. And for the same reasons, placing objects on dark backgrounds (rather than a constant white) also helps.

Lastly, we revisit our choice of camera poses for image collection. Our aim was to encourage overlap in VLM responses to ensure the most reliable responses can ``win'' during aggregation. So we took images from regular yaw intervals on a circle around the object. But there are other possible schemes, e.g., one might maximize the information content of each view by prioritizing atypical object views. CAP3D took this alternative approach, varying camera height simultaneously with yaw to include top and bottom views of the object. Though we didn't have access to the original images from CAP3D, we reverse-engineered their camera poses, and found that our choice worked better with our method. We also varied the free parameter in our choice, i.e., the polar angle at which the camera is placed facing the object. This didn't make much difference.

On the whole, we found PaLI to be quite robust to the visual and image settings which affect object appearance.

\begin{table}[h]
\caption{\textbf{VLM sensitivity to image conditions and object appearance.} We collect and score PaLI VQA type annotations using cosine similarity on Objaverse-LVIS. All results are view-aggregated.}
\begin{tabular}{|l|c|} \hline
Hyperparameters & \begin{tabular}[c]{@{}c@{}}PaLI VQA\\type similarity score\end{tabular} \\ \hline
\begin{tabular}[c]{@{}l@{}}\emph{Object appearance}\\ Textured (Cycles)\\Untextured (Workbench) \end{tabular} & \begin{tabular}[c]{@{}l@{}}\\ $\bf{0.593 \pm 0.290}$\\$0.549 \pm 0.289$\end{tabular}\\ \hline
\begin{tabular}[c]{@{}l@{}}\emph{Scene lighting}\\ Surround area lights\\ Stationary point light\\ Camera backlight\end{tabular} & \begin{tabular}[c]{@{}l@{}}\\ $0.580 \pm 0.291$ \\ $0.586 \pm 0.291$ \\ $\bf{0.593 \pm 0.290}$ \end{tabular}\\ \hline
\begin{tabular}[c]{@{}l@{}}\emph{Image background}\\ Black (0, 0, 0)\\ Dark grey (100, 100, 100)\\ White (255, 255, 255)\end{tabular}                                         & \begin{tabular}[c]{@{}l@{}}\\ $0.603 \pm 0.296$\\$\bf{0.612 \pm 0.294}$\\$0.593 \pm 0.290$\end{tabular}\\ \hline
\begin{tabular}[c]{@{}l@{}}\emph{Camera poses}\\ CAP3D (w/ top \& bottom views)\\Ours (constant polar angle)\end{tabular}                                         & \begin{tabular}[c]{@{}l@{}}\\$0.588 \pm 0.290$\\$\bf{0.593 \pm 0.290}$\end{tabular} \\ \hline
\begin{tabular}[c]{@{}l@{}}\emph{Camera polar angle $\theta$}\\ 64 degrees\\68 degrees\\72 degrees\end{tabular}                                         & \begin{tabular}[c]{@{}l@{}}\\$0.591 \pm 0.291$ \\ $0.593 \pm 0.290$\\$\bf{0.593 \pm 0.290}$\end{tabular}\\ \hline
\end{tabular}
\label{tab:appearances}
\end{table}

\subsection{Detailed Comparison with CAP3D}
\label{sec:extra_hallucination_examples}

Figure \ref{fig:cap3d_vs_ours_by_hallucination} extends Figure \ref{fig:hallucination_main} from the main text, showing the next 18 objects with the highest CAP3D caption blow-up. 

As discussed, the caption blow-up ratio is a precise indicator but lacks recall. To focus on hallucination cases which are not picked up by the blow-up ratio, we show more examples in Figure \ref{fig:cap3d_vs_ours}. This figure visualizes single-view captions from both pipelines to illustrate where the aggregates come from. The last two rows of the figure present two specific examples presented as failure cases in the CAP3D paper.

\begin{figure}
    \centering
    \begin{subfigure}{\columnwidth}
    \includegraphics[width=\textwidth]{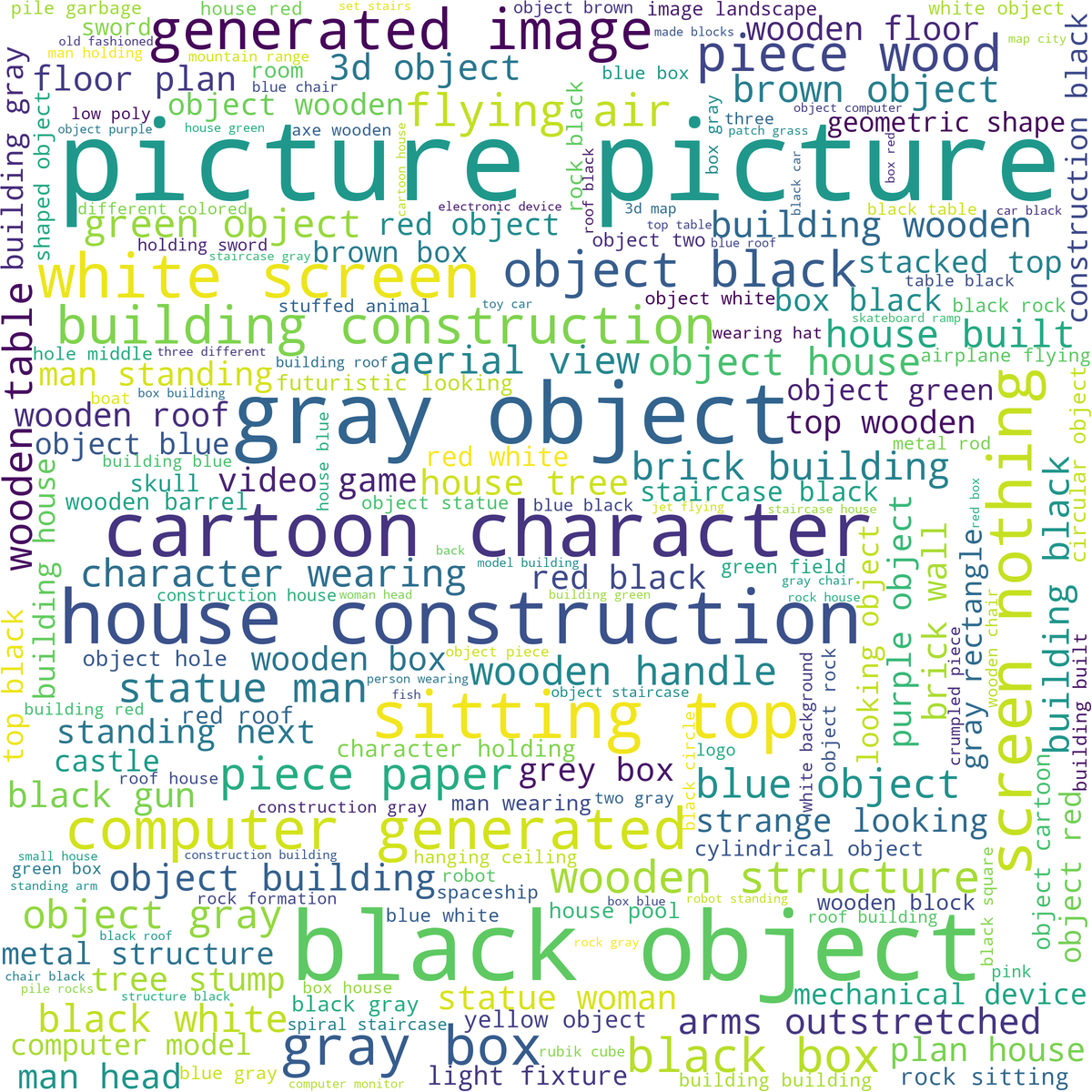}
    \caption{PaLI aggregate top captions.}
    \label{fig:pali_wordcloud} %
    \end{subfigure}
    
    \begin{subfigure}{\columnwidth}
    \includegraphics[width=\textwidth]{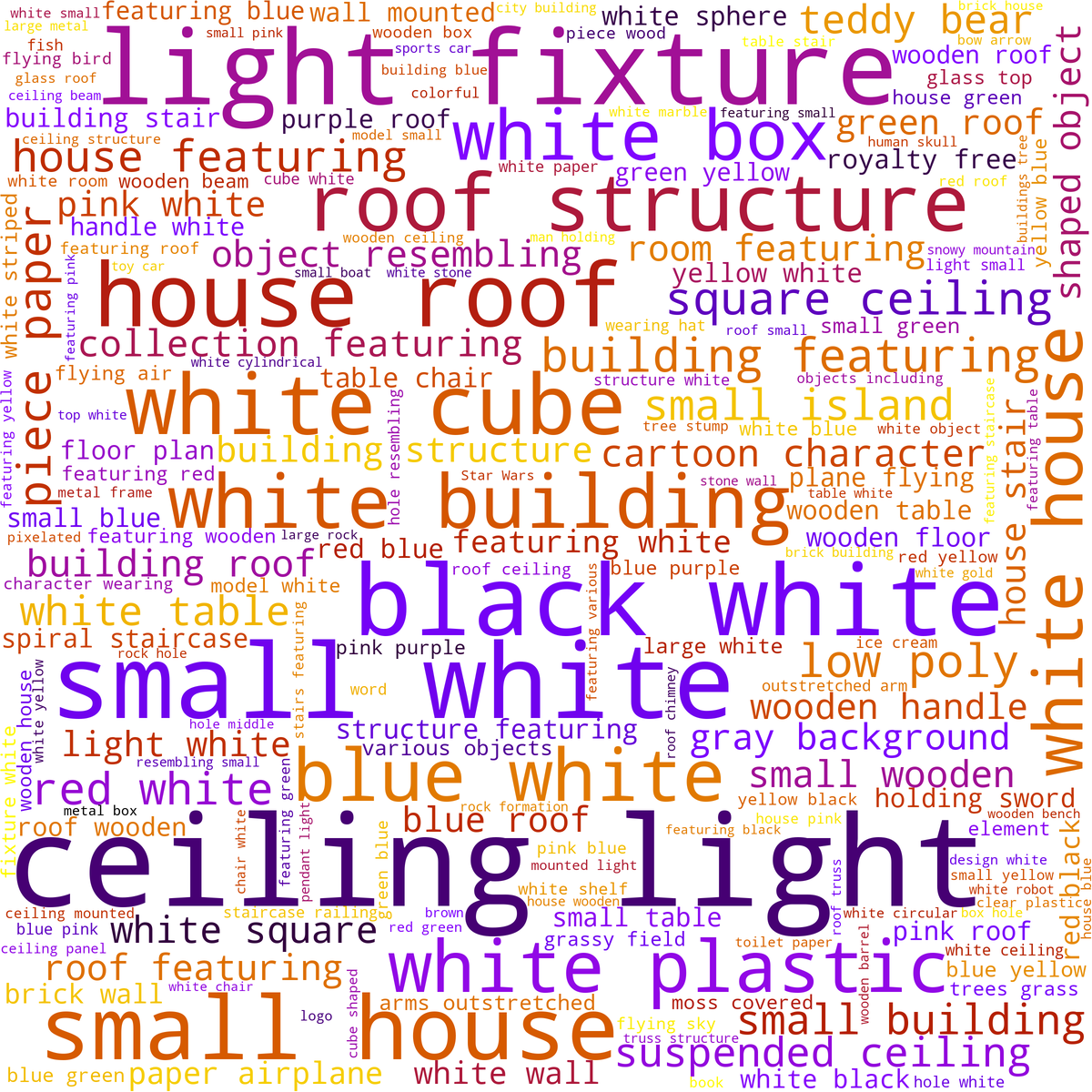}
    \caption{CAP3D aggregate captions.}
    \label{fig:cap3d_wordcloud} %
    \end{subfigure}

    \caption{\textbf{Word clouds comparing PaLI and CAP3D captions based on word frequency.} Articles and prepositions are dropped.}
    \label{fig:word_clouds}
\end{figure}

Finally, we plot word frequency clouds in Figure \ref{fig:word_clouds} and compute aggregate statistics in Table \ref{tab:keyword_caption_comparison} for both sets of captions. While our pipeline is missing captions for 4\% of objects (mostly animations) which we dropped while rendering, CAP3D is missing captions for 17\% of Objaverse. CAP3D's reason for dropping a significant fraction of objects was that they lacked ``sufficient camera information for rendering'' (see Sec 3.2 of \citet{luo2023scalable}). Nevertheless, both pipelines have better coverage than artist-written tags or descriptions in the original dataset---those are empty for 38\% and 37\% of all objects respectively. %

\begin{figure*}[h]
    \centering

    \begin{subfigure}{0.95\textwidth}
        \centering
        \includegraphics[width=\textwidth]{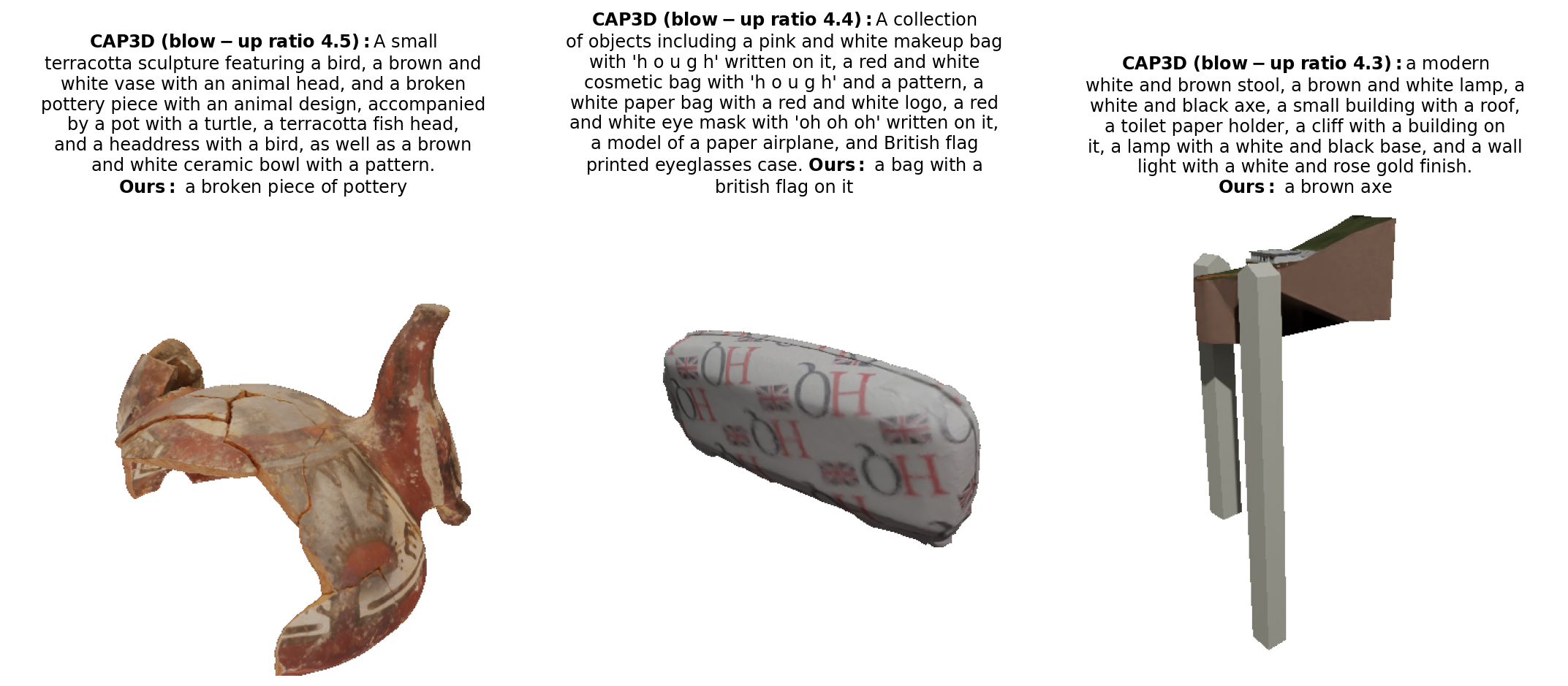}
    \end{subfigure} \hrule

    \begin{subfigure}{0.95\textwidth}
        \centering
        \includegraphics[width=\textwidth]{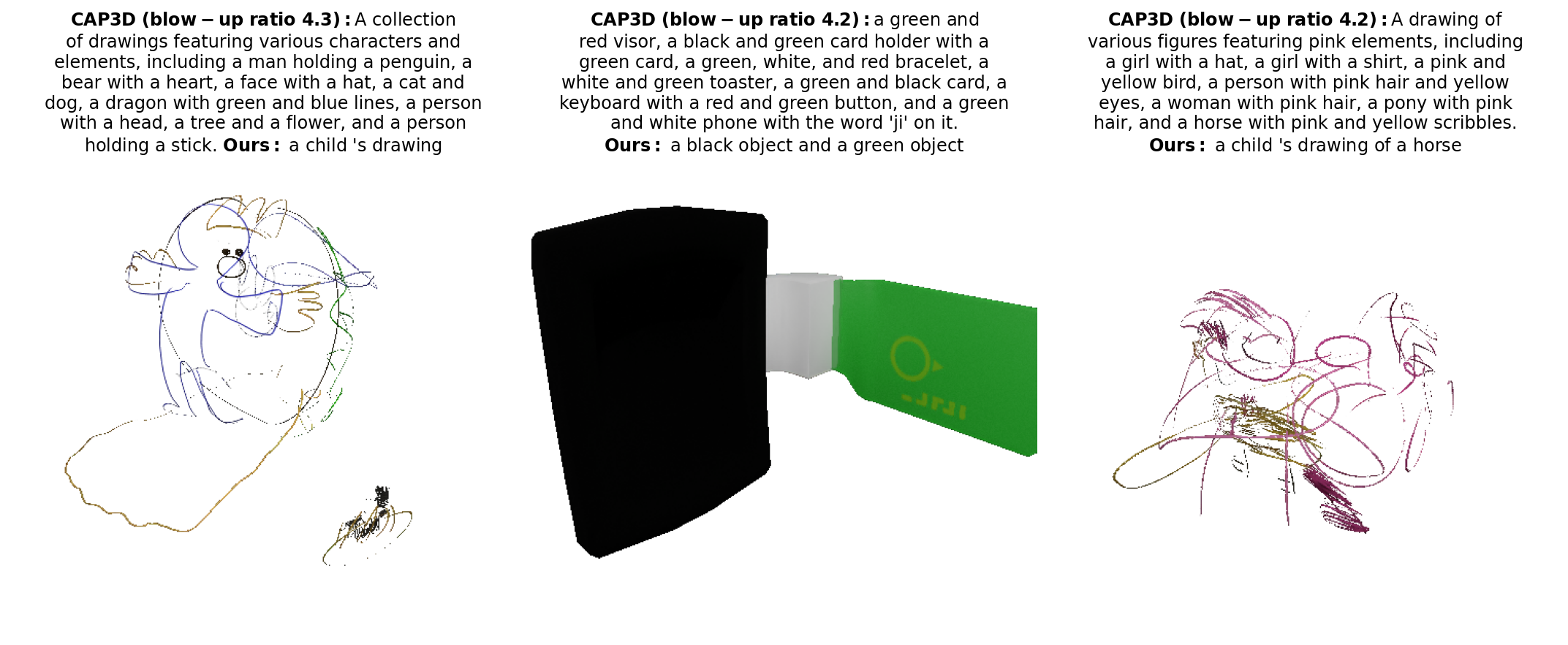}
    \end{subfigure} \hrule

    \begin{subfigure}{0.95\textwidth}
        \centering
        \includegraphics[width=\textwidth]{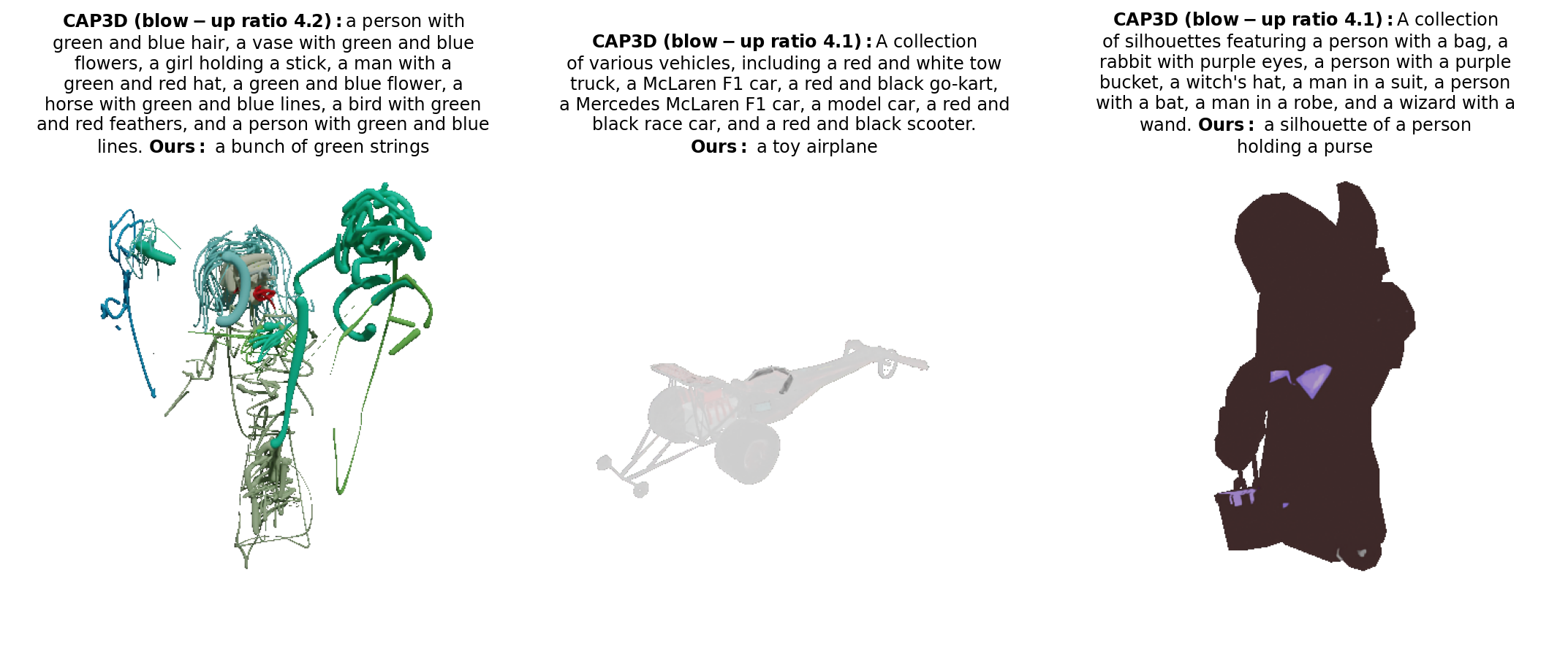}
    \end{subfigure} \hrule
    
    \caption{\textbf{Comparison of view-aggregated captions from our pipeline versus the current SoTA, CAP3D, on objects with the highest CAP3D caption blow-up ratios.} We skip the three objects already presented in Fig \ref{fig:hallucination_main}.
    }
    \label{fig:cap3d_vs_ours_by_hallucination}
\end{figure*}

\begin{figure*}
    \ContinuedFloat

    \begin{subfigure}{0.95\textwidth}
        \centering
        \includegraphics[width=\textwidth]{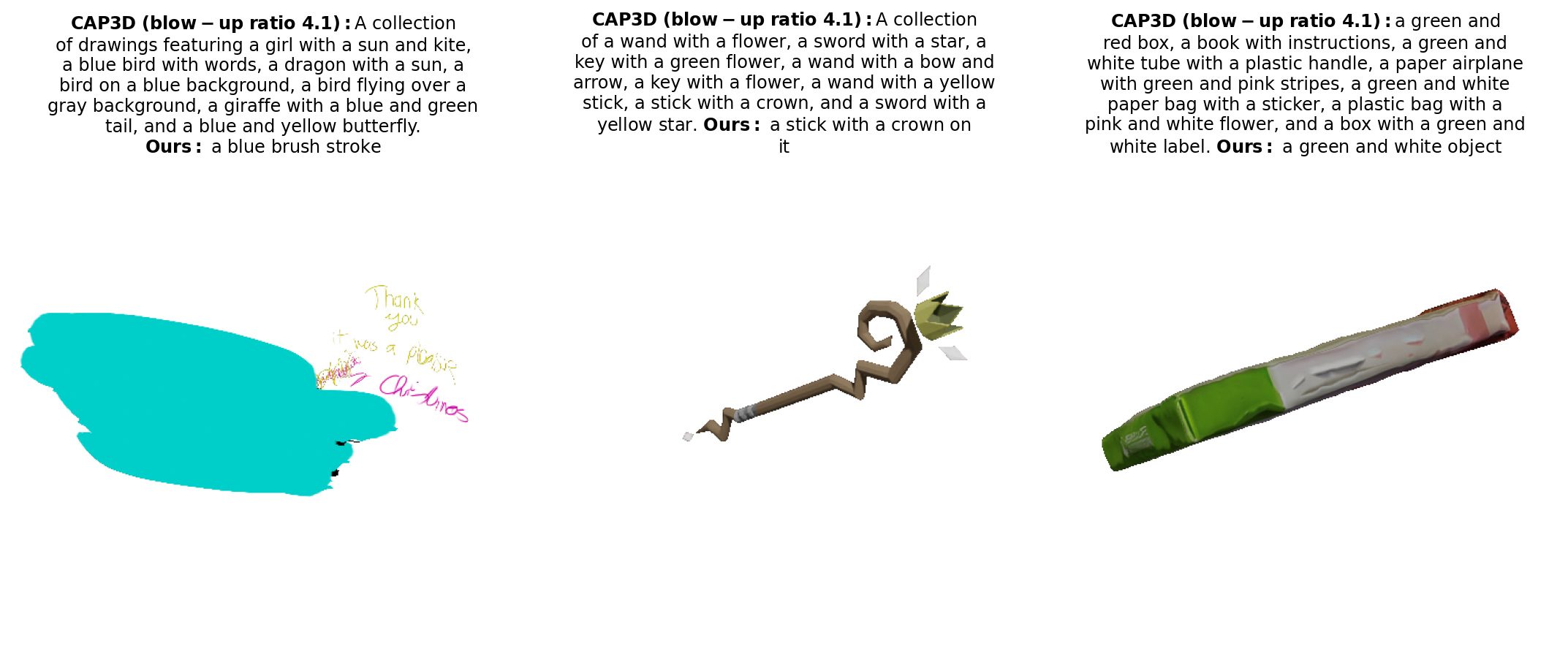}
    \end{subfigure} \hrule

    \begin{subfigure}{0.95\textwidth}
        \centering
        \includegraphics[width=\textwidth]{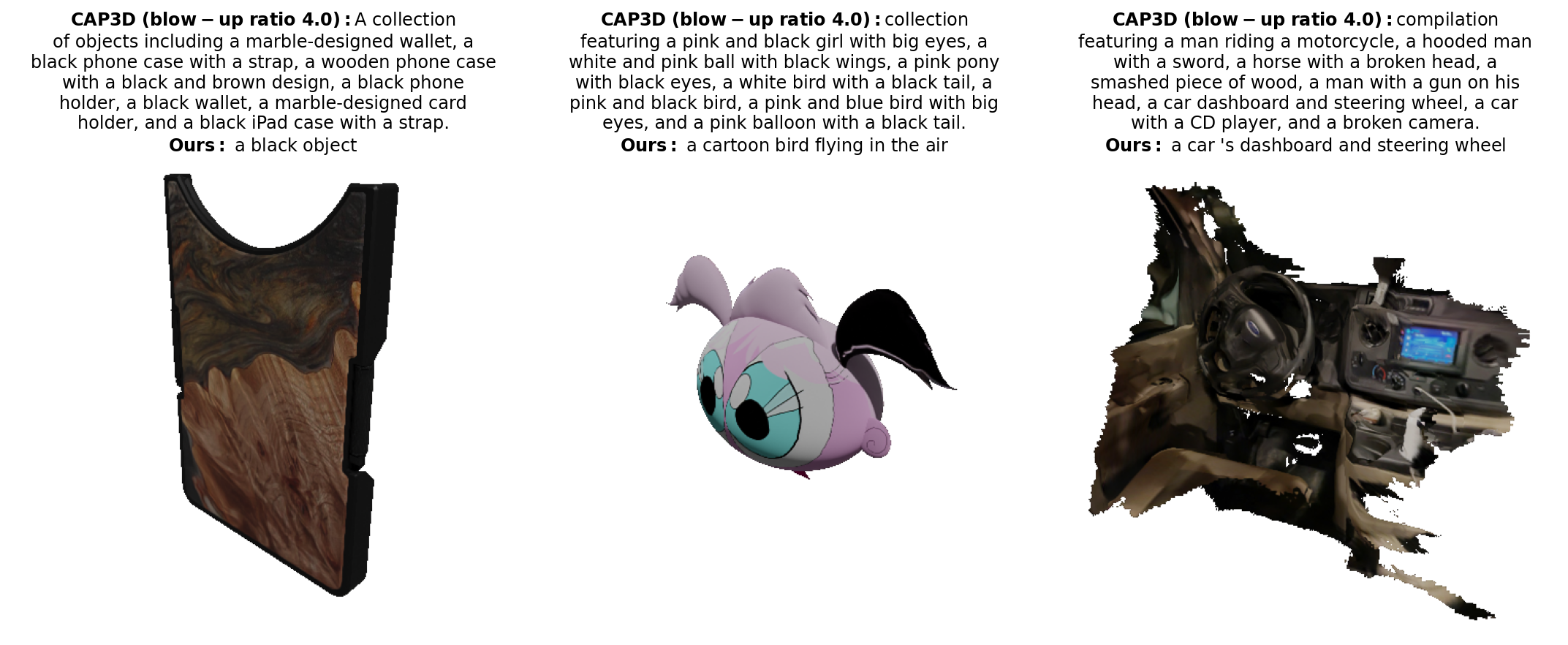}
    \end{subfigure} \hrule

    \begin{subfigure}{0.95\textwidth}
        \centering
        \includegraphics[width=\textwidth]{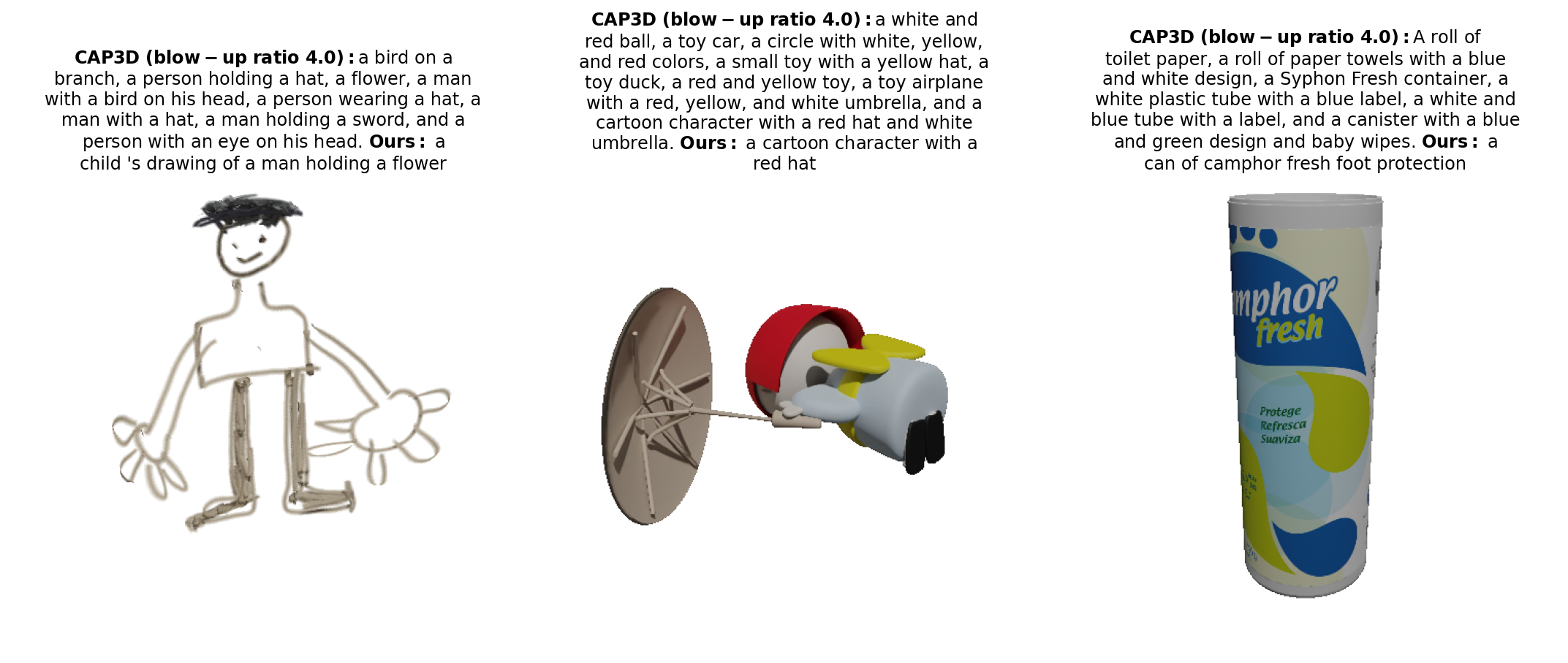}
    \end{subfigure} \hrule
    
    \caption{\textbf{Comparison of view-aggregated captions from our pipeline versus the current SoTA, CAP3D, on objects with the highest CAP3D caption blow-up ratios (contd.)}}
\end{figure*}

\begin{figure*}
    \centering
    \vspace{-5pt}

    \begin{subfigure}{\textwidth}
        \centering
        \includegraphics[width=\textwidth]{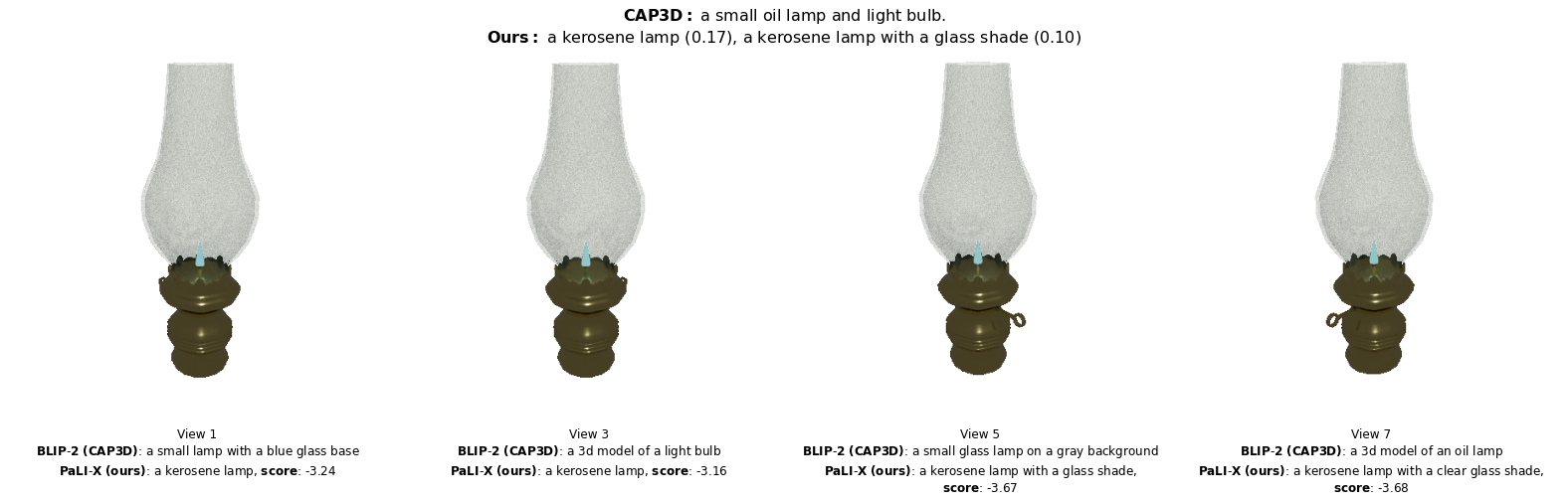}
    \end{subfigure} \hrule

    \begin{subfigure}{\textwidth}
        \centering
        \includegraphics[width=\textwidth]{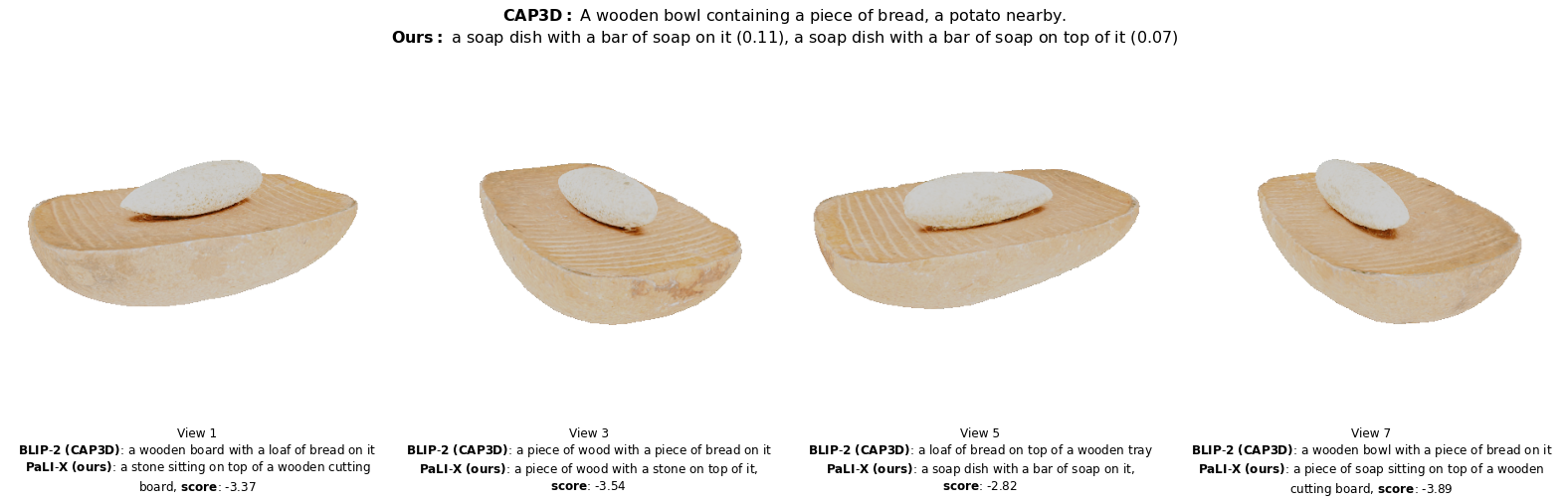}
    \end{subfigure} \hrule

    \begin{subfigure}{\textwidth}
        \centering
        \includegraphics[width=\textwidth]{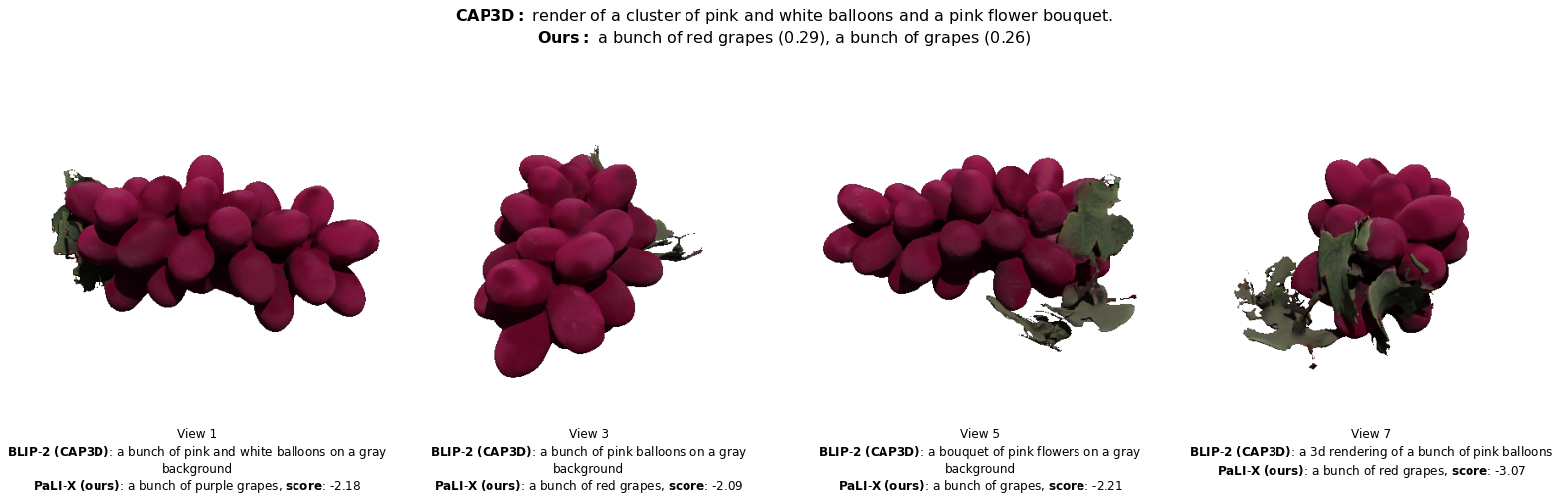}
    \end{subfigure} \hrule

        \begin{subfigure}{\textwidth}
        \centering
        \includegraphics[width=\textwidth]{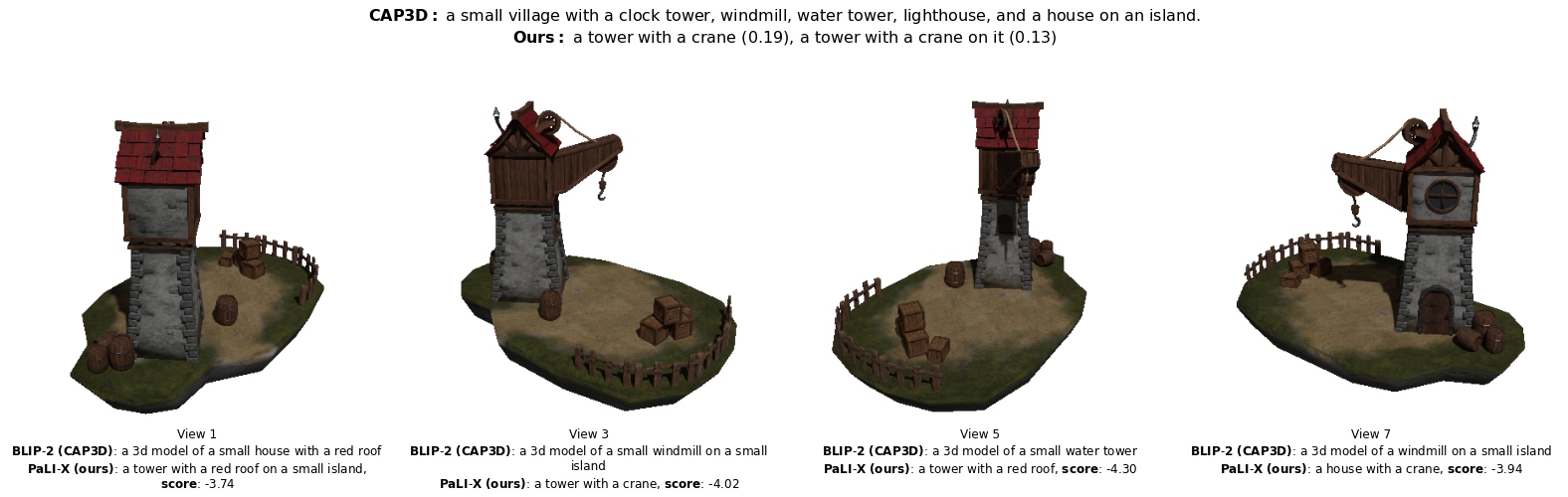}
    \end{subfigure} \hrule

    \caption{\textbf{Comparison of captions from our pipeline versus the current SoTA, CAP3D.} Besides the aggregates, we also show view-specific captions from the underlying VLMs (PaLI-X and BLIP-2). %
    }
    \label{fig:cap3d_vs_ours}
\end{figure*}

\begin{figure*}
    \ContinuedFloat

    \vspace{-5pt}
    \begin{subfigure}{\textwidth}
        \centering
        \includegraphics[width=\textwidth]{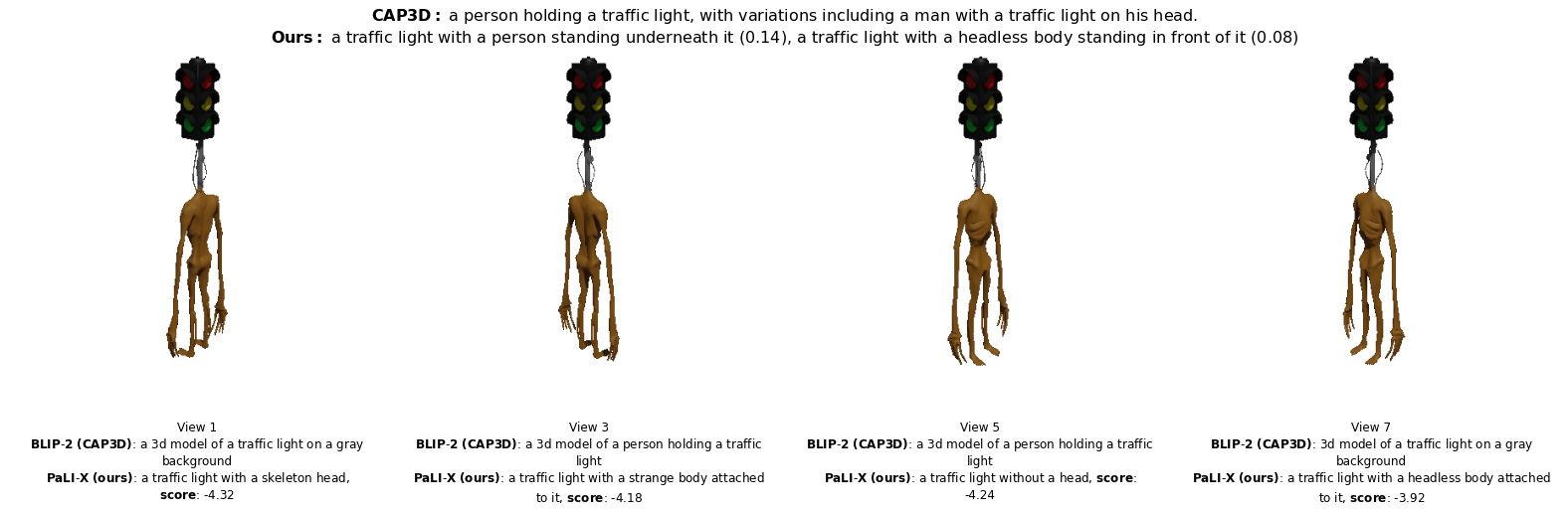}
    \end{subfigure} \hrule
    
    \begin{subfigure}{\textwidth}
        \centering
        \includegraphics[width=\textwidth]{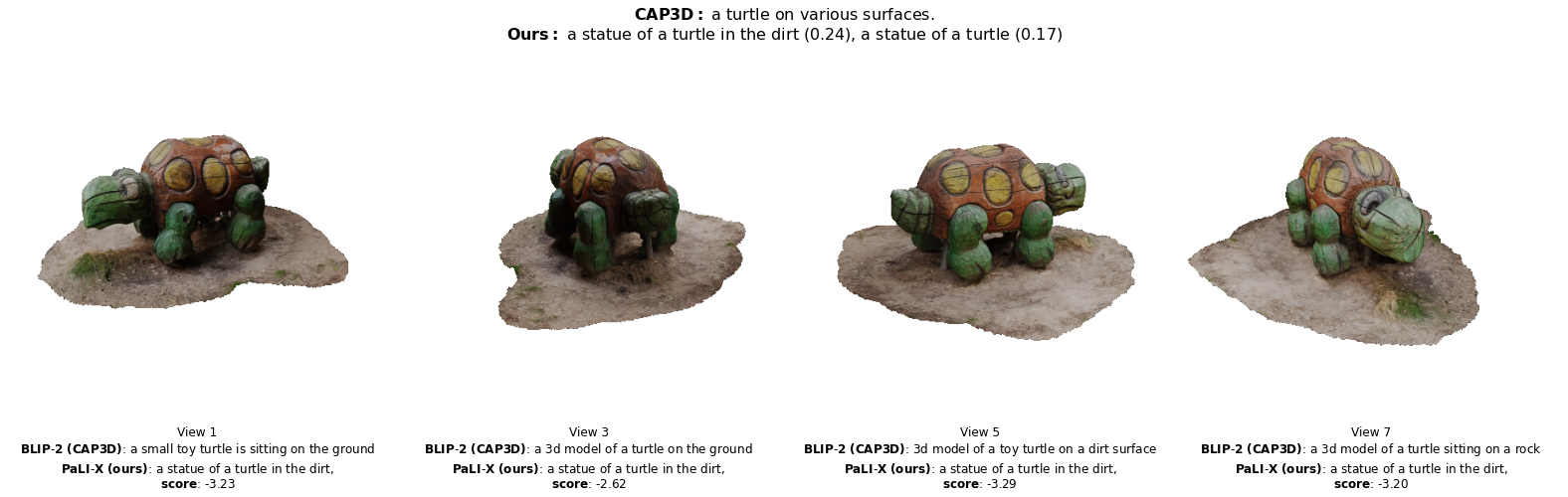}
    \end{subfigure} \hrule

    \begin{subfigure}{\textwidth}
        \centering
        \includegraphics[width=\textwidth]{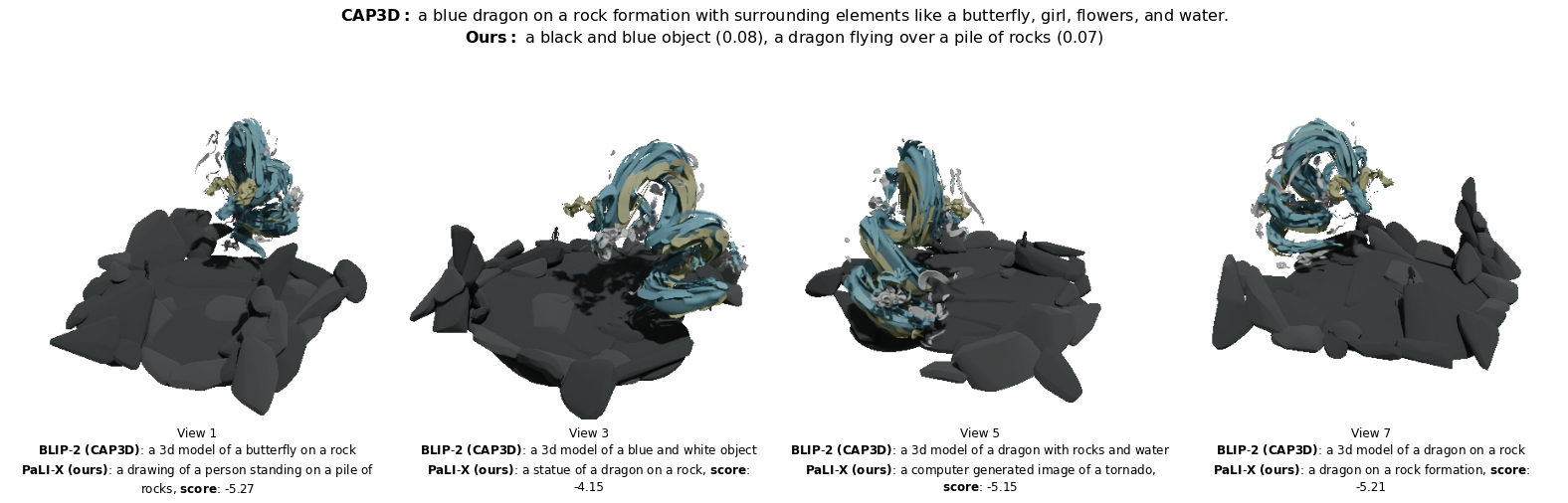}
    \end{subfigure} \hrule

    \begin{subfigure}{\textwidth}
        \centering
        \includegraphics[width=\textwidth]{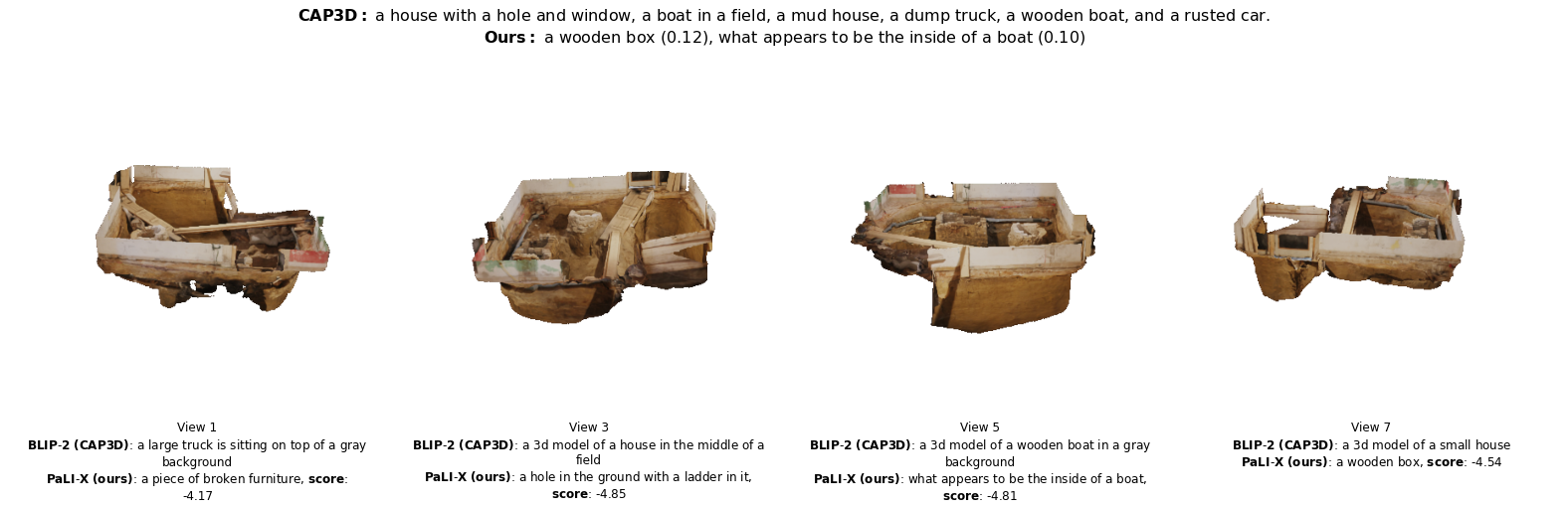}
    \end{subfigure} \hrule

    \caption{\textbf{Comparison of captions from our pipeline versus the current SoTA, CAP3D (contd.)} The last two rows were described as failure cases for CAP3D in that paper.}
\end{figure*}

\subsection{Material Inference}
\label{sec:extended_results_material_prediction}

We show material predictions under all inference scenarios for various objects from our test set in Table \ref{tab:material_prediction_examples}. For aggregate statistics, see Table \ref{tab:material_prediction} in the main text.

\subsection{What Factors Matter for Different Properties}
\label{sec:qualitative_assessment}

We take a qualitative look at the effect of varying the question, view, or appearance of an object when predicting various properties. We also observe an effect of object size though we kept it fixed in our pipeline.

To examine the role of the question, we fix a set of objects and probe them for all properties discussed so far. We avoid aggregating across object views in this section. For each property we show PaLI responses to question variants separately, then the effect of aggregating across questions using ScoreAgg. We cover changes in question wording and what prior inferences are specified. 

\textbf{Open-vocabulary properties.} Figure \ref{fig:all_properties_fixed_objects} starts with type and material inference. We observe PaLI-X shows varying confidence in its responses on different objects. This could provide signal for when we need to refine predictions (e.g., by asking more questions). 

We find PaLI-X surprisingly capable of describing what an object might contain, even when the contained object is hidden or hypothetical (e.g., ``money'' in a ``wallet'' or ``people'' in a ``boat''). It is unclear whether the variance across questions here is due to significant changes in wording, or the complexity of the knowledge we're probing. We ask a single question on object affordance---the space of possible responses and entropy of predictions are large even under a single question. The model appears to understand use cases for all our objects. 

\textbf{Physical behavior.} When it comes to physical properties, PaLI makes more mistakes. It knows that a ``leather wallet'' can be molded but not crushed; that a ``brick bakery'' would be hard to deform; that a rock can only be crushed. On the other hand, it expects a ``wood boat'' or ``snow globe`` to be moldable by hand. It takes most objects to be fragile (including a ``soda can'' or ``remote control'') and incapable of floating in water (with the obvious exception of a ``boat''). Interestingly, whether or not an object contains something significantly changes the likelihood of its floating or sinking.

\textbf{Lift-ability.} When predicting if objects are liftable, we run into the consequences of not controlling for object size. The model is uncertain even in obvious cases like a ``clay toy'' and ``aluminum soda can.'' These are both objects for which the height was the maximum dimension; they take up more vertical image space and possibly appear abnormally large to the model. Though we could try aggregating/marginalizing over varying-size renderings of an object, a better solution might specify an object's scale (if available) to the model explicitly.

\textbf{Color and count.} The model can express colors in words correctly. It offers multi-color responses when there's a mix of colors (e.g., ``yellow and blue'' for the clay toy). When it comes to counting, PaLI can separate single objects (count=1) with high precision. It also does reasonably for disjoint objects. For busy scenes, it seems to get the scale right---we even see the variance of its numerical responses increase (e.g. the ``banquet'').

\textbf{Pose.} An object appears in different poses to the model across our rendered images (i.e., changes in view). We test whether the model can tell the front of an object from its back or side (Fig \ref{fig:pose_inference}). This seems readily possible for asymmetric objects, perhaps helped by the presence of facial features in the case of the ``lion''. For more symmetrical objects like the ``wallet'', the model can tell side views from (squarely) front or back views, but is somewhat confused.

\textbf{Shininess.} Whether an object is shiny is a physical property that follows from its type and material. This offers an opportunity to assess whether VLMs are more sensitive to their visual inputs or prompts. We take an object with ambiguous shine, then vary the lighting conditions, camera angle, and background image color (Fig \ref{fig:shininess_inference}). We ask four question variants specifying the object's type or material in all combinations. We find almost no effect of varying the prompt in comparison to the effect of varying the image settings. The ``soda can'' is described as shiny and dull in equal measure across the appearance-varying probes.

\textbf{Conclusion.} Running multiple VLM probes and aggregating across them can be a powerful technique to uncover/deal with VLM uncertainty. We can meaningfully marginalize over views of an object (camera pose) or variations in the question. But VLMs can be thrown off by visual features (e.g., lighting, contrast, or object size) especially when they’re not relevant to the query. Such queries should be decoupled from object appearance either by specifying more relevant information in the prompt, or failing that, perhaps using a visionless model.

\input{material_predictions_figure}

\begin{figure*}

    \begin{subfigure}{\textwidth}
        \centering
        \includegraphics[width=\textwidth]{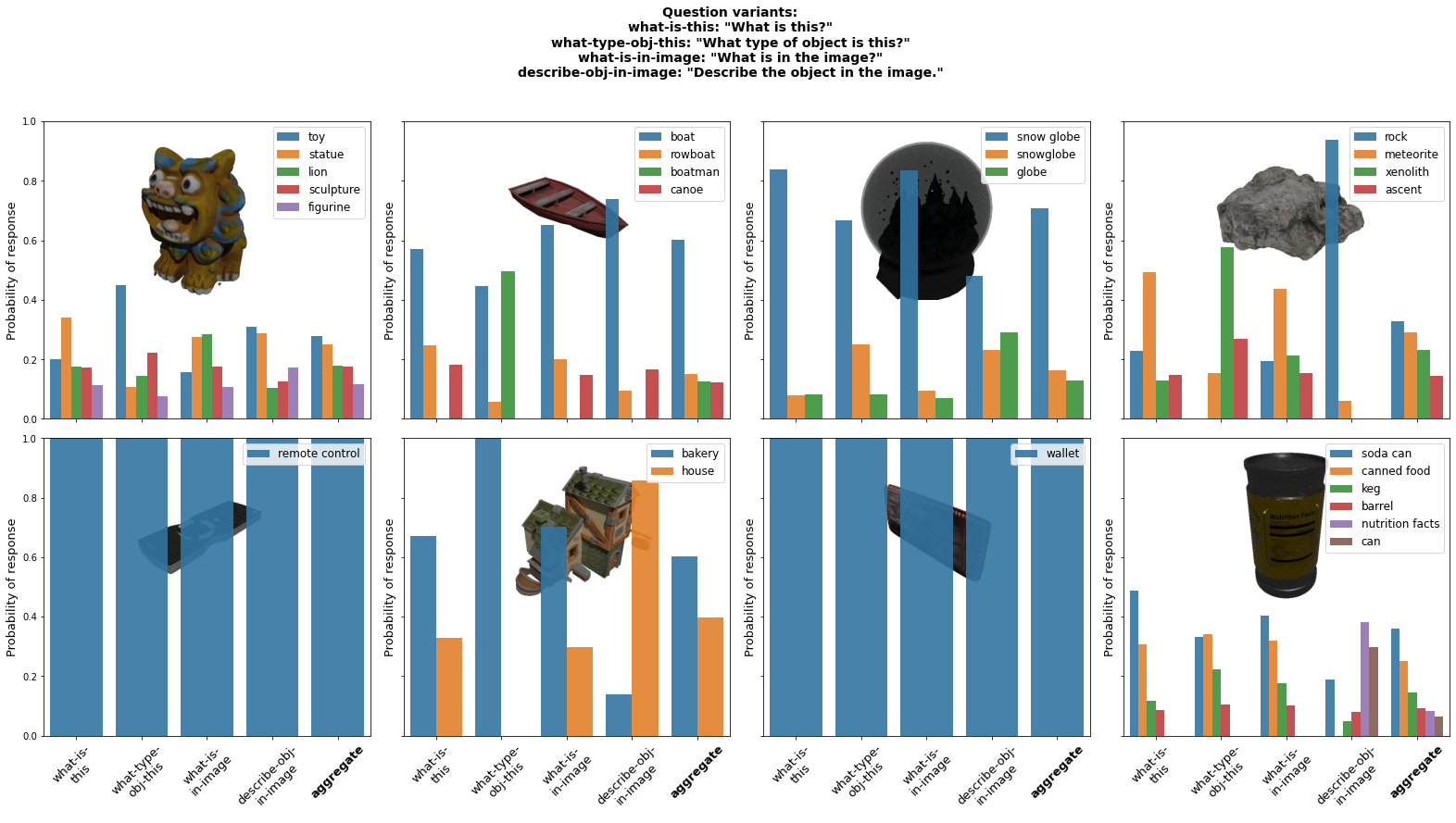}
    \end{subfigure} \hrule
    
    \begin{subfigure}{\textwidth}
        \centering
        \includegraphics[width=\textwidth]{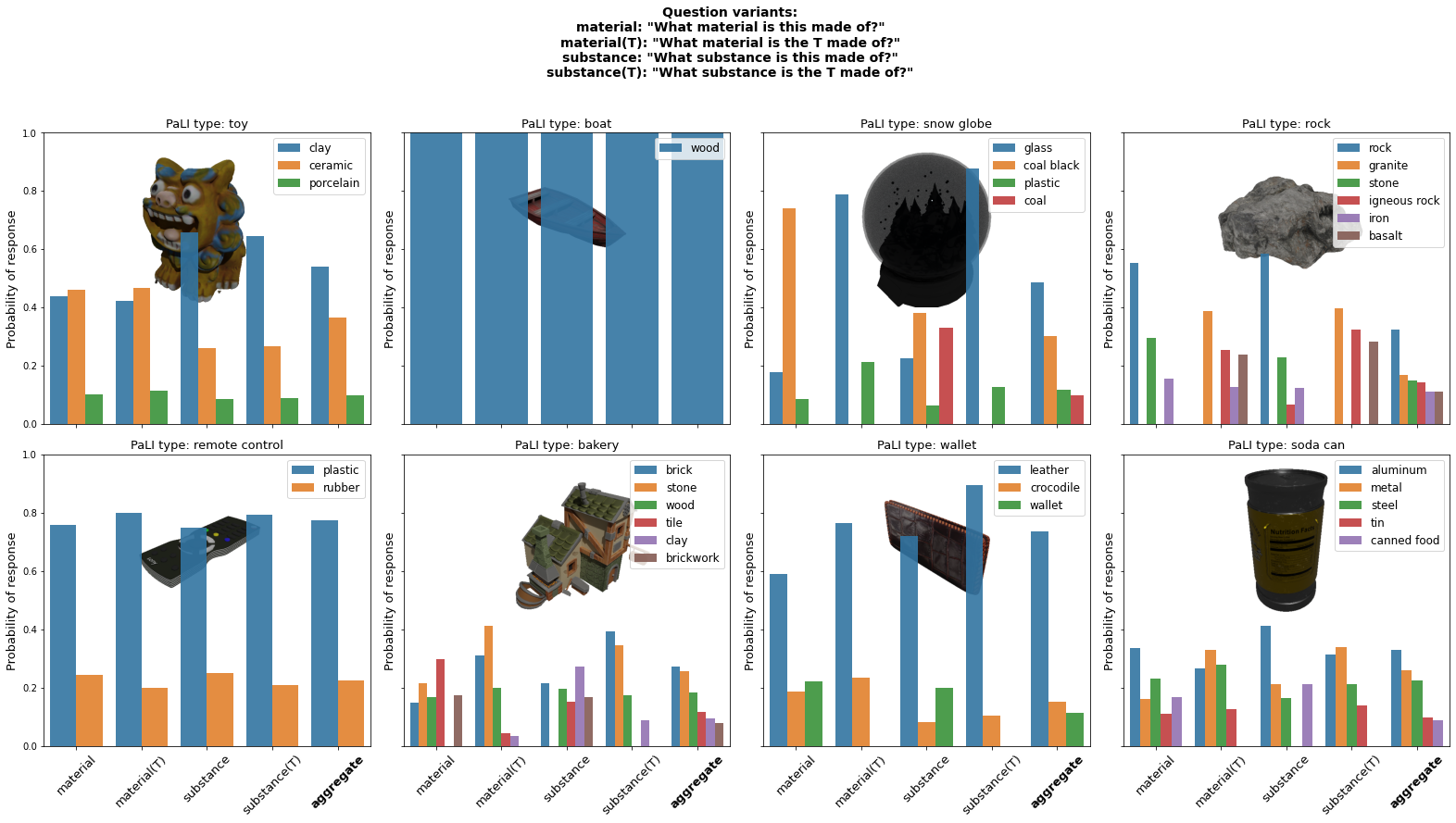}
    \end{subfigure} \hrule

    \caption{\textbf{Responses to various questions, on a fixed set of objects, for a series of properties.} Top: type. Bottom: material.}
    \label{fig:all_properties_fixed_objects}
\end{figure*}

\begin{figure*}
    \ContinuedFloat

    \begin{subfigure}{\textwidth}
        \centering
        \includegraphics[width=\textwidth]{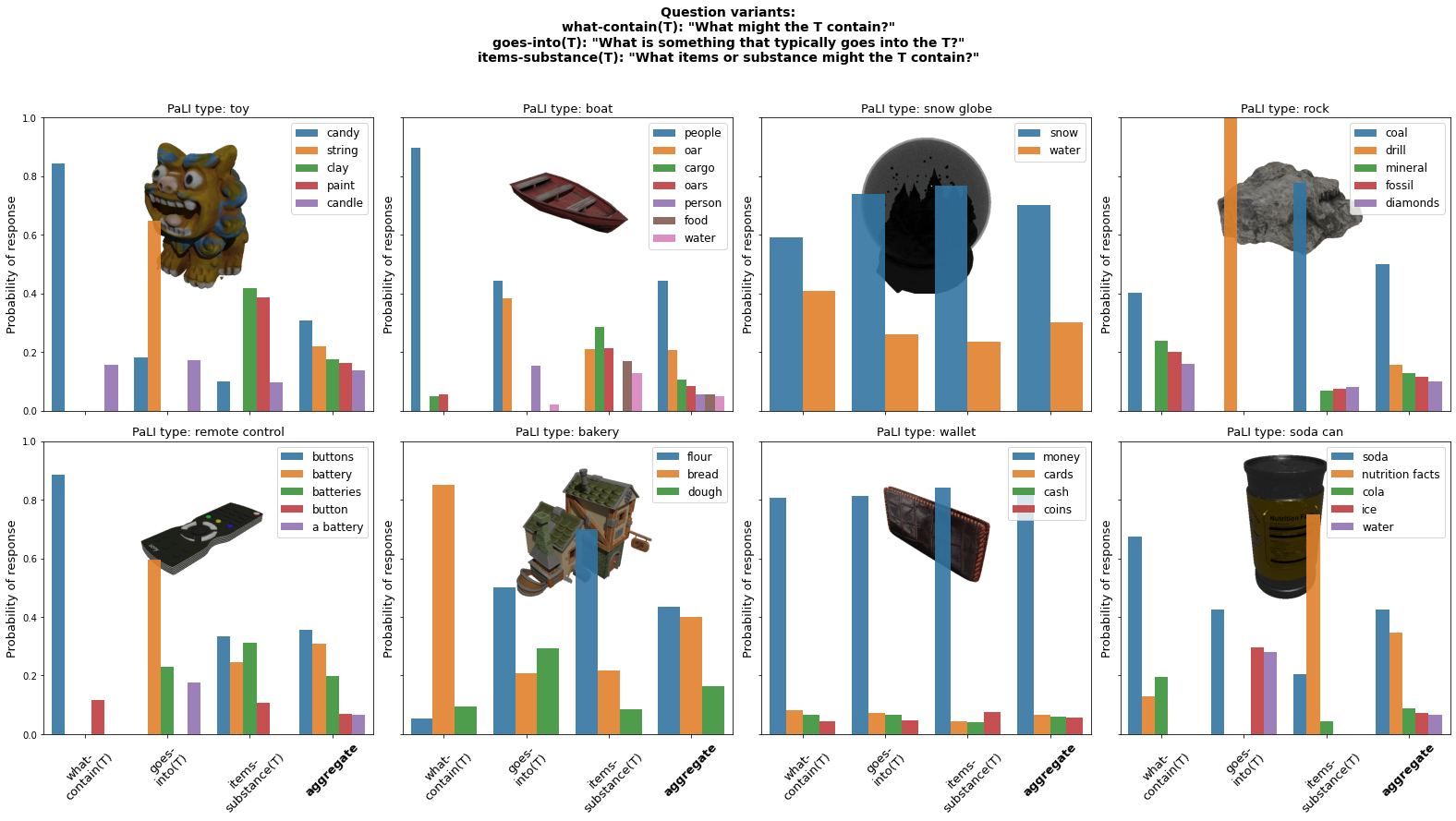}
    \end{subfigure} \hrule
    
    \begin{subfigure}{\textwidth}
        \centering
        \includegraphics[width=\textwidth]{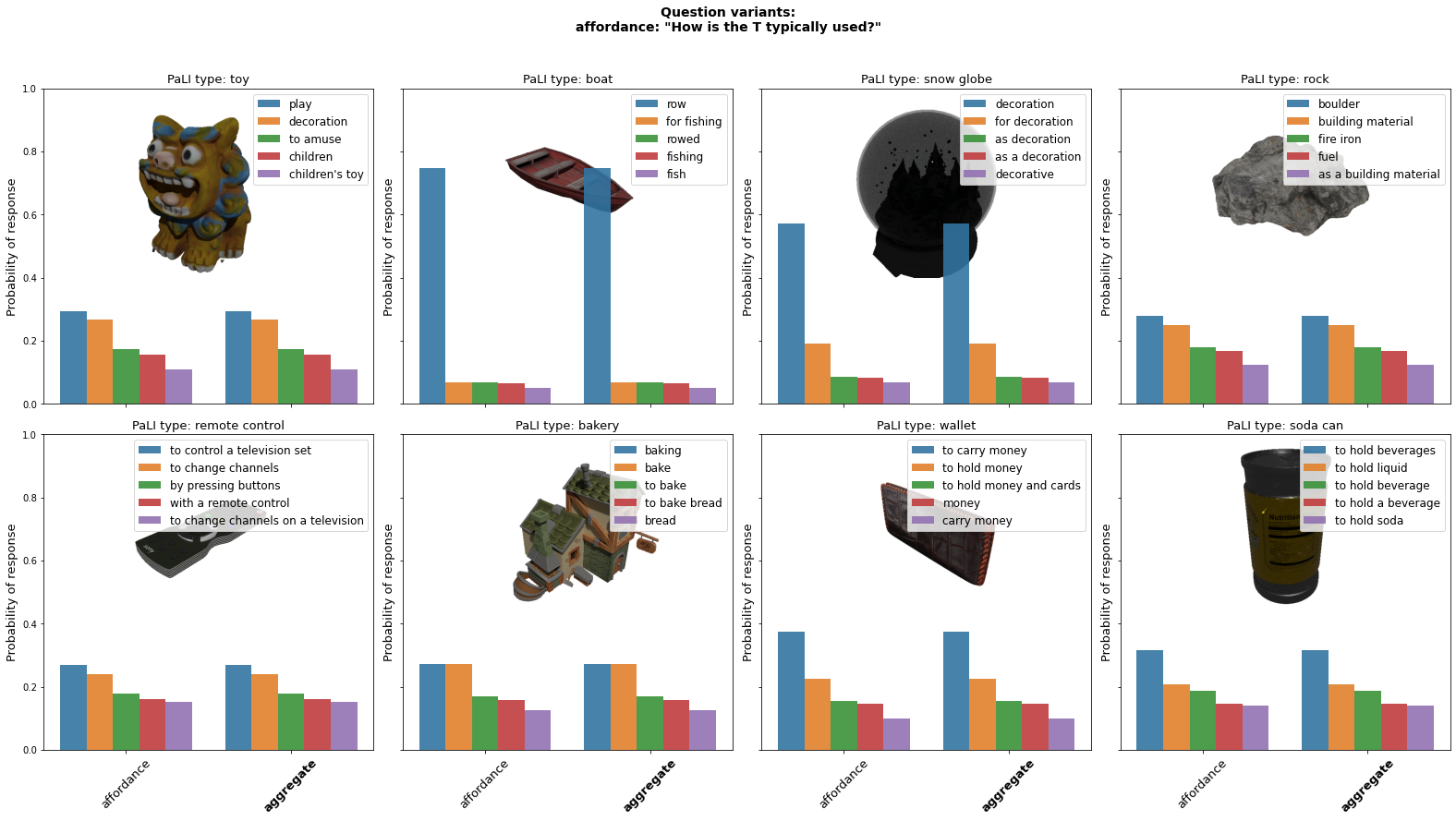}
    \end{subfigure} \hrule

    \caption{\textbf{Variations of questions per property for a fixed set of objects (contd.)} Top: containment. Bottom: affordance.}
\end{figure*}

\begin{figure*}
    \ContinuedFloat
    
    \begin{subfigure}{\textwidth}
        \centering
        \includegraphics[width=\textwidth]{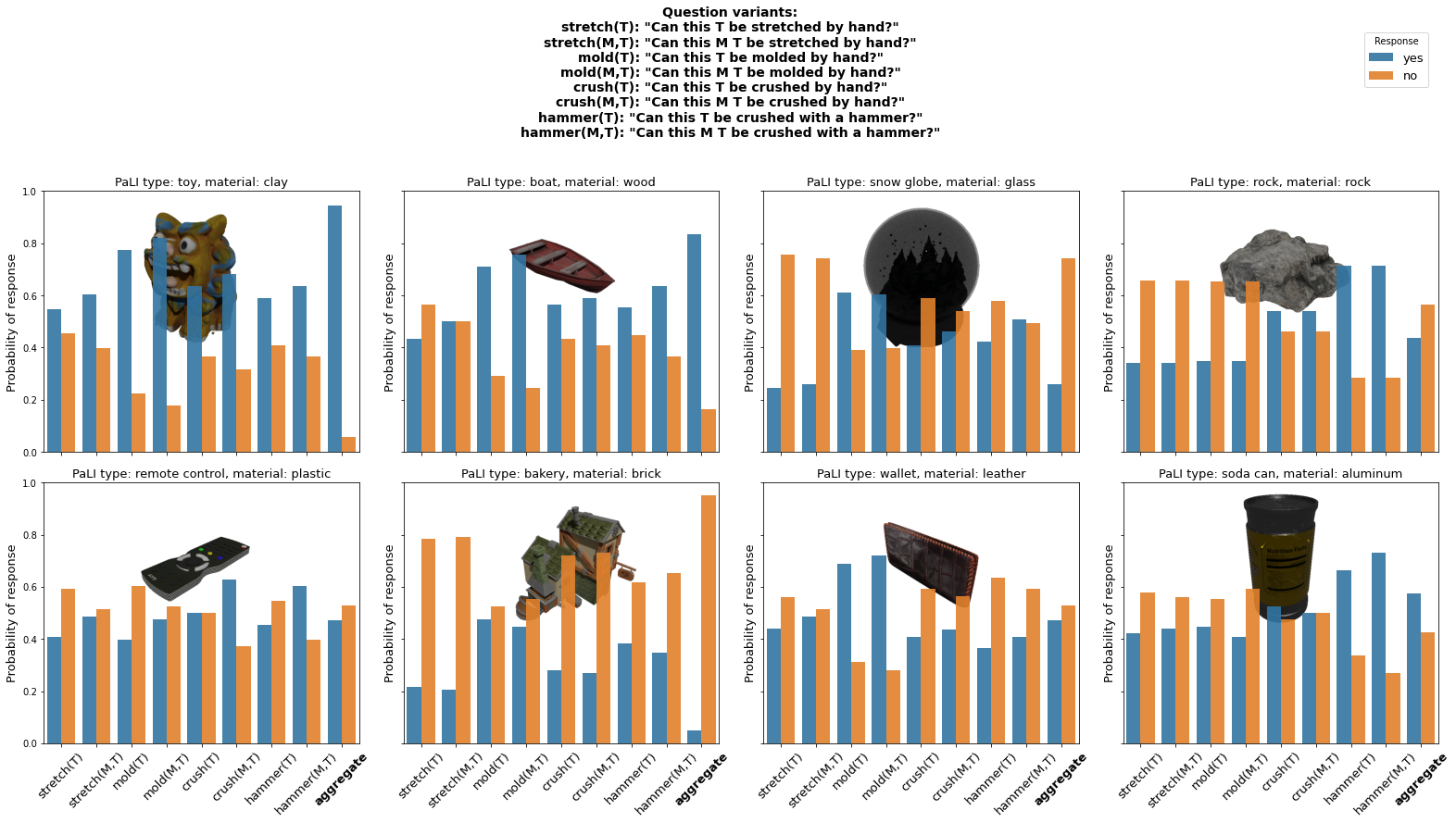}
    \end{subfigure} \hrule

    \begin{subfigure}{\textwidth}
        \centering
        \includegraphics[width=\textwidth]{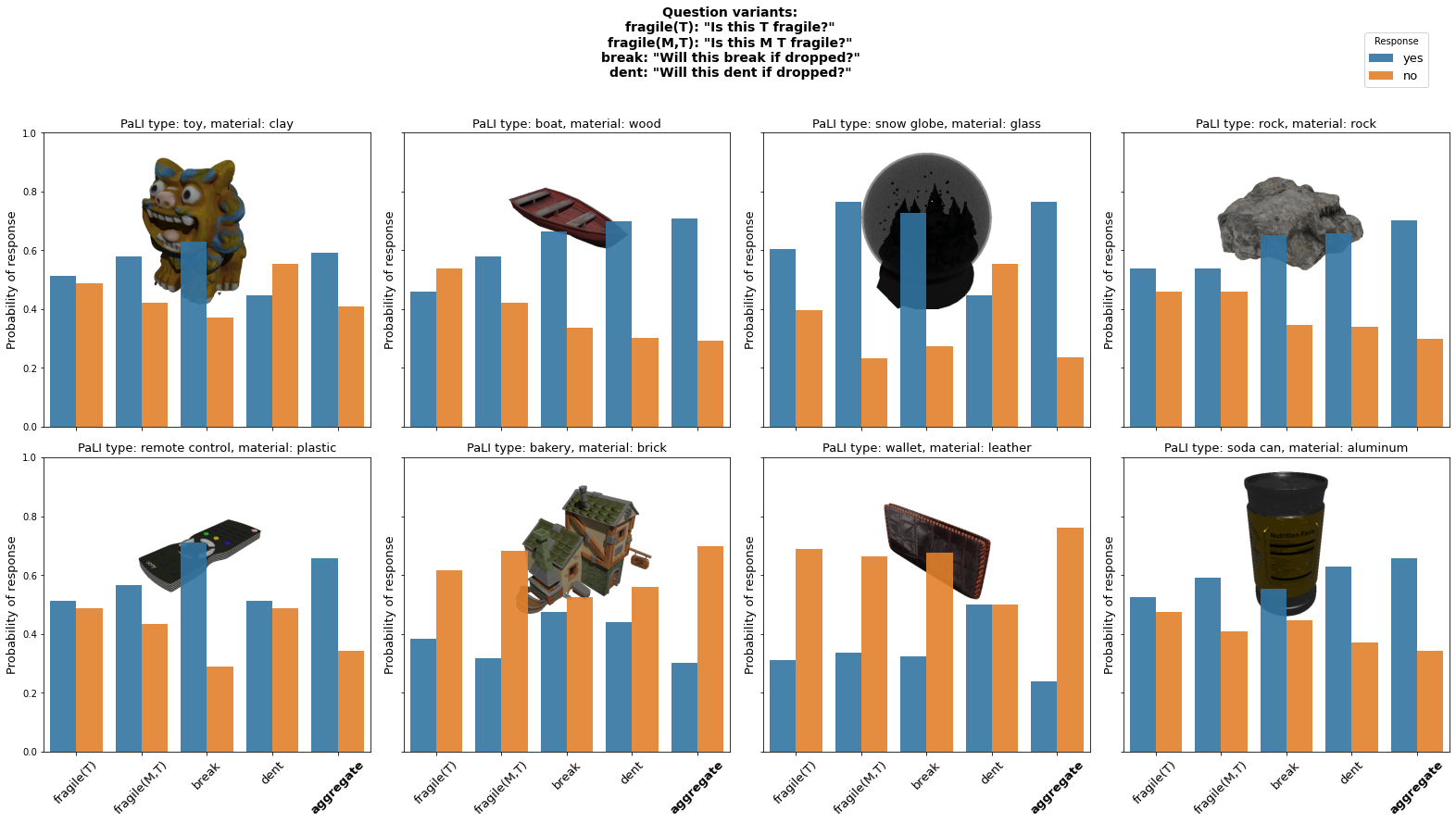}
    \end{subfigure} \hrule

    \caption{\textbf{Responses to various questions, on a fixed set of objects, for a series of properties (contd.)}  Top: deformability. Bottom: fragility.}
\end{figure*}

\begin{figure*}
    \ContinuedFloat

    \begin{subfigure}{\textwidth}
        \centering
        \includegraphics[width=\textwidth]{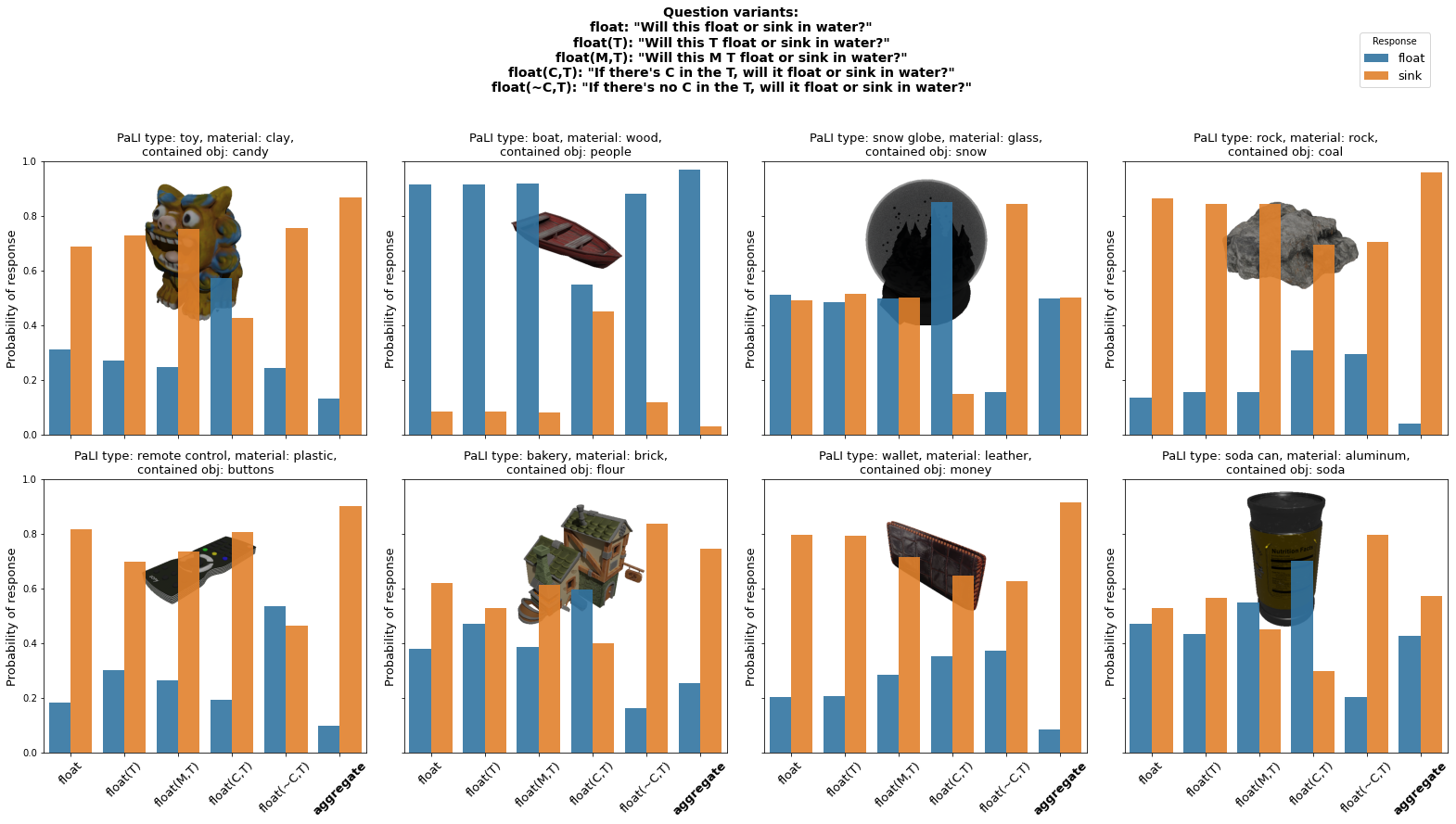}
    \end{subfigure} \hrule
    
    \begin{subfigure}{\textwidth}
        \centering
        \includegraphics[width=\textwidth]{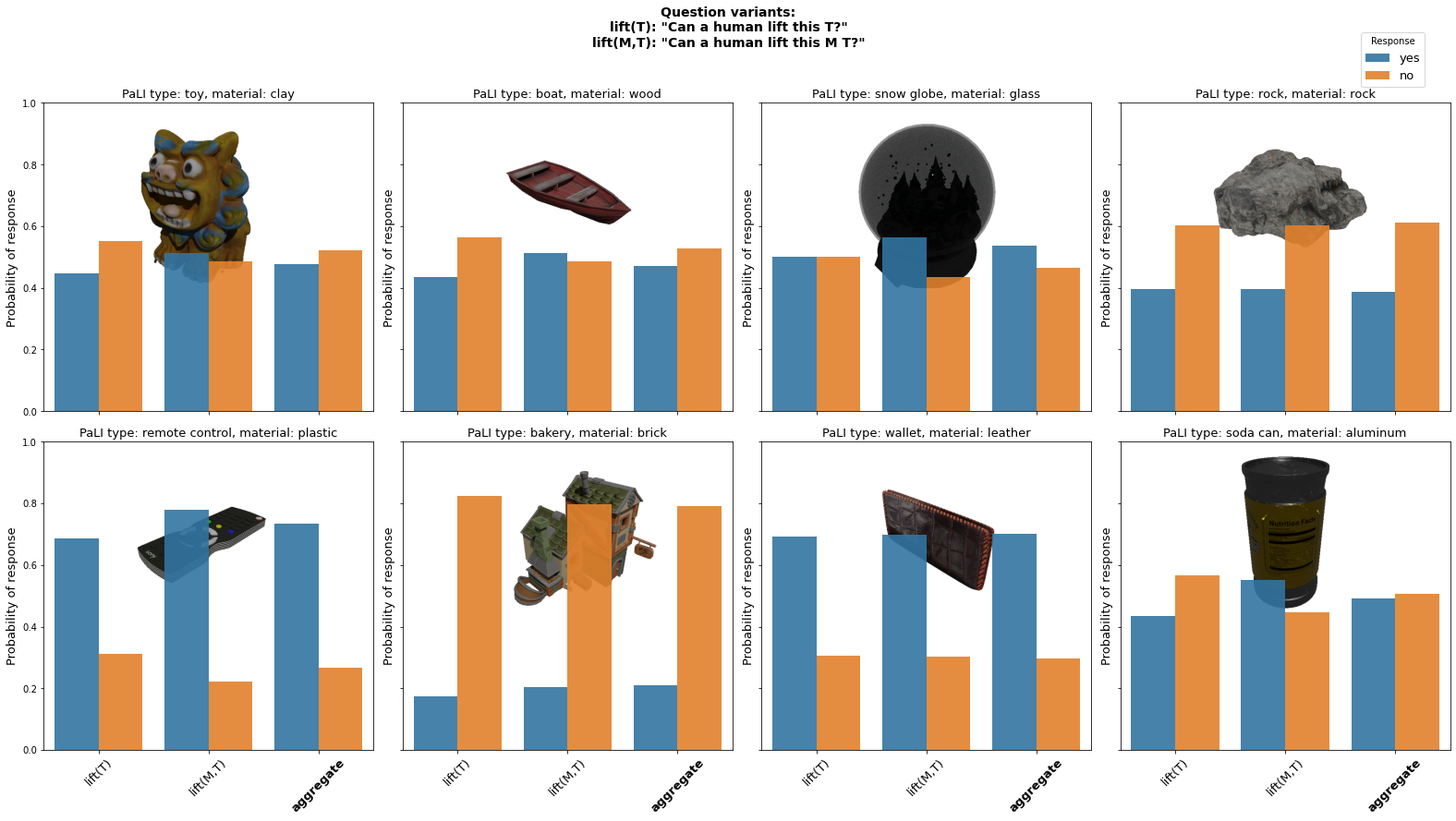}
    \end{subfigure} \hrule

    \caption{\textbf{Responses to various questions, on a fixed set of objects, for a series of properties (contd.)} Top: float-ability. Bottom: lift-ability.}
\end{figure*}

\begin{figure*}
    \ContinuedFloat

    \begin{subfigure}{\textwidth}
        \centering
        \includegraphics[width=\textwidth]{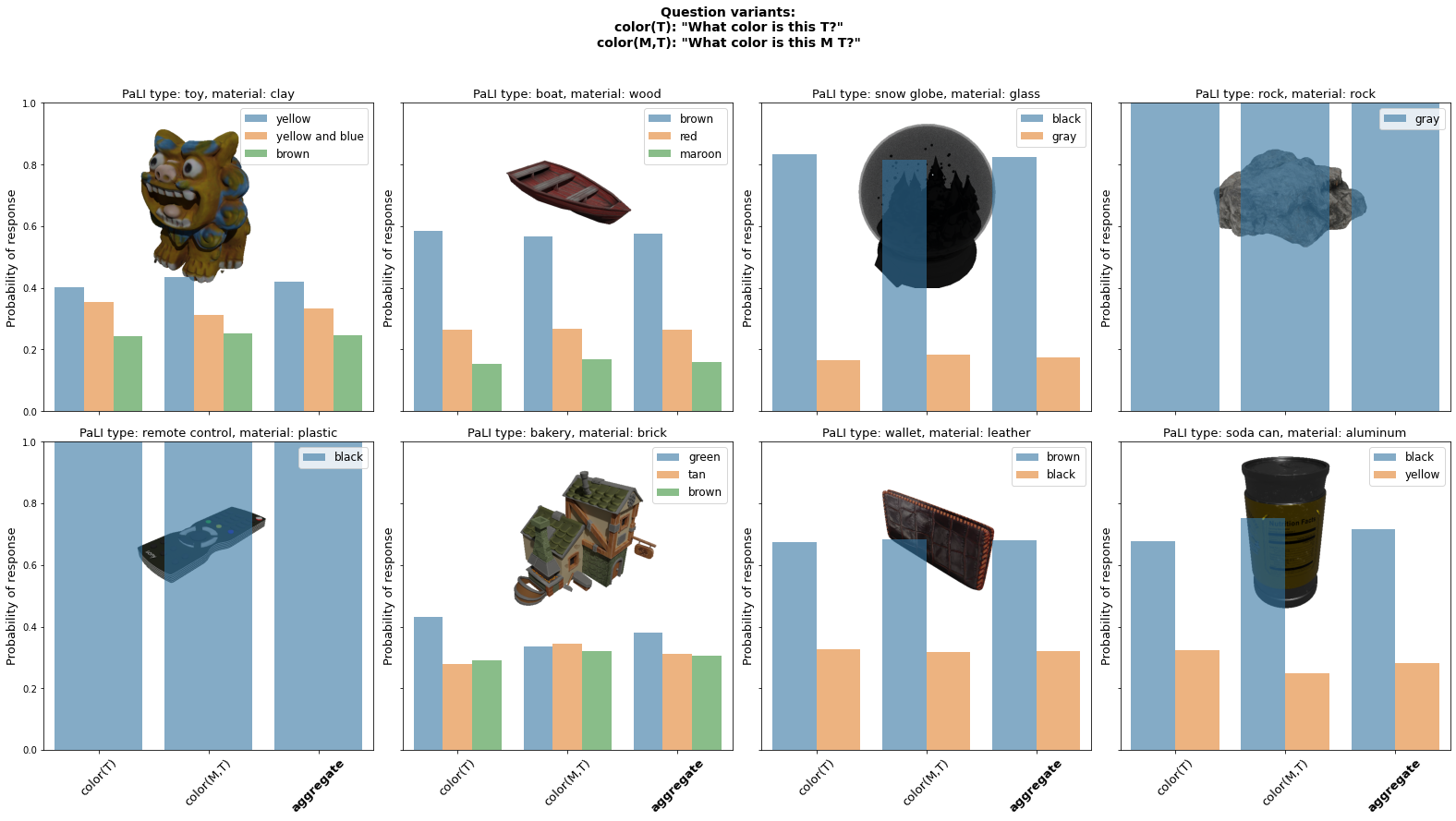}
    \end{subfigure} \hrule

    \caption{\textbf{Responses to various questions, on a fixed set of objects, for a series of properties (contd.)}}
\end{figure*}

\begin{figure*}

    \begin{subfigure}{\textwidth}
        \centering
        \includegraphics[width=\textwidth]{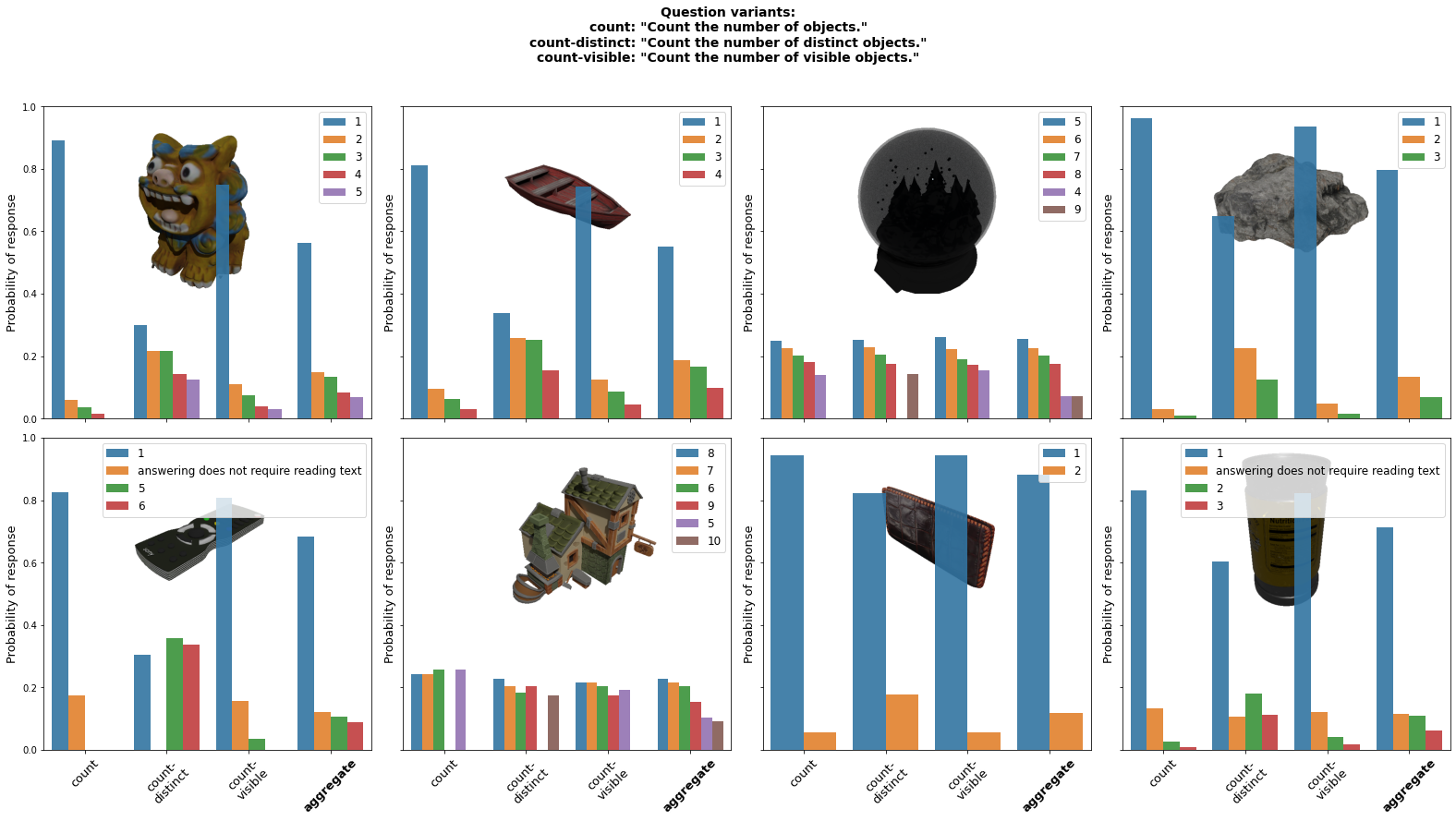}
    \end{subfigure} \hrule

    \begin{subfigure}{\textwidth}
        \centering
        \includegraphics[width=\textwidth]{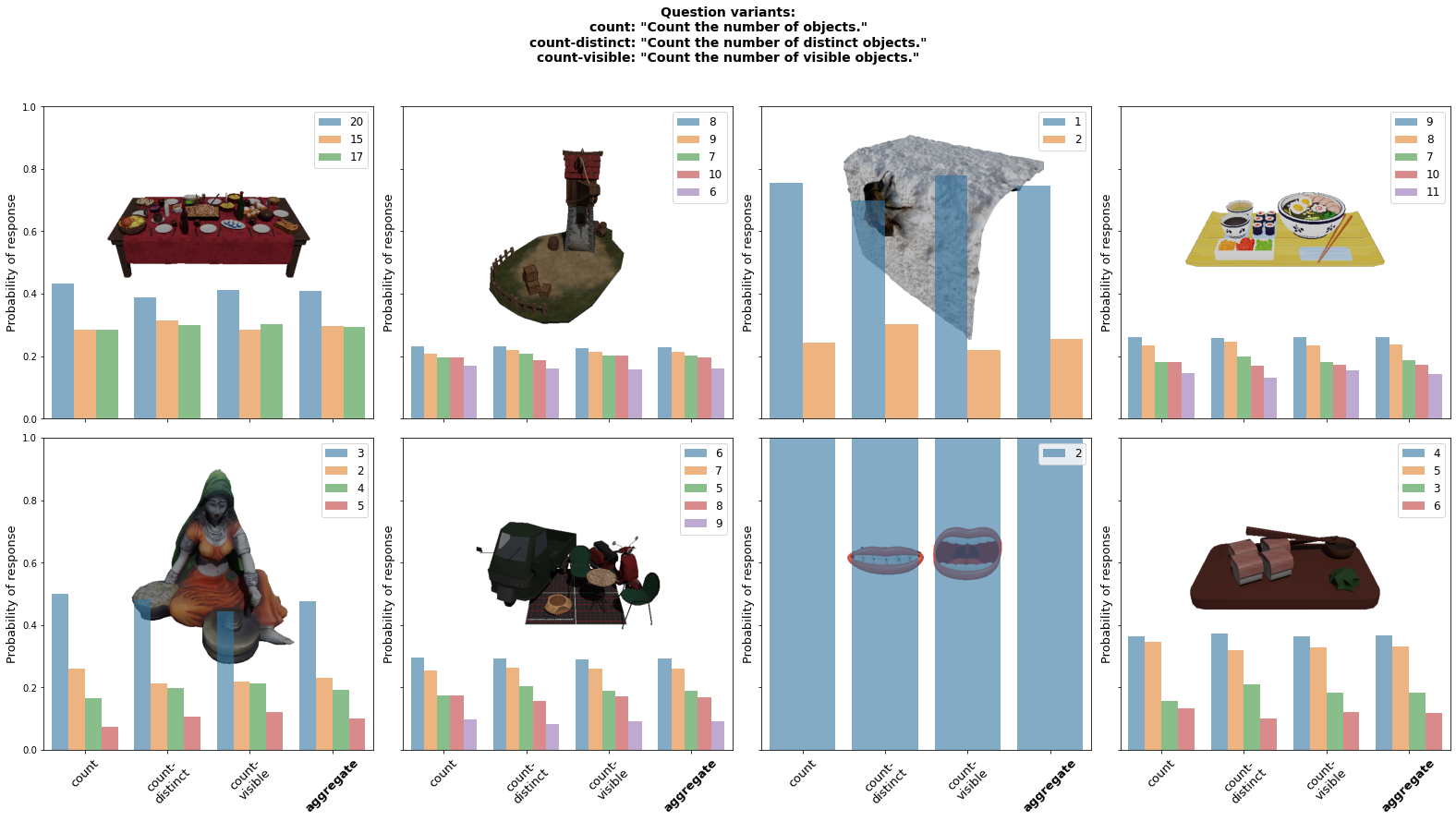}
    \end{subfigure} \hrule
    
\caption{\textbf{Variations of questions to infer object count.} We show the prior set of objects (above), then a set of multi-object scenes (below).}
\end{figure*}

\begin{figure*}

    \begin{subfigure}{\textwidth}
        \centering
        \includegraphics[width=\textwidth]{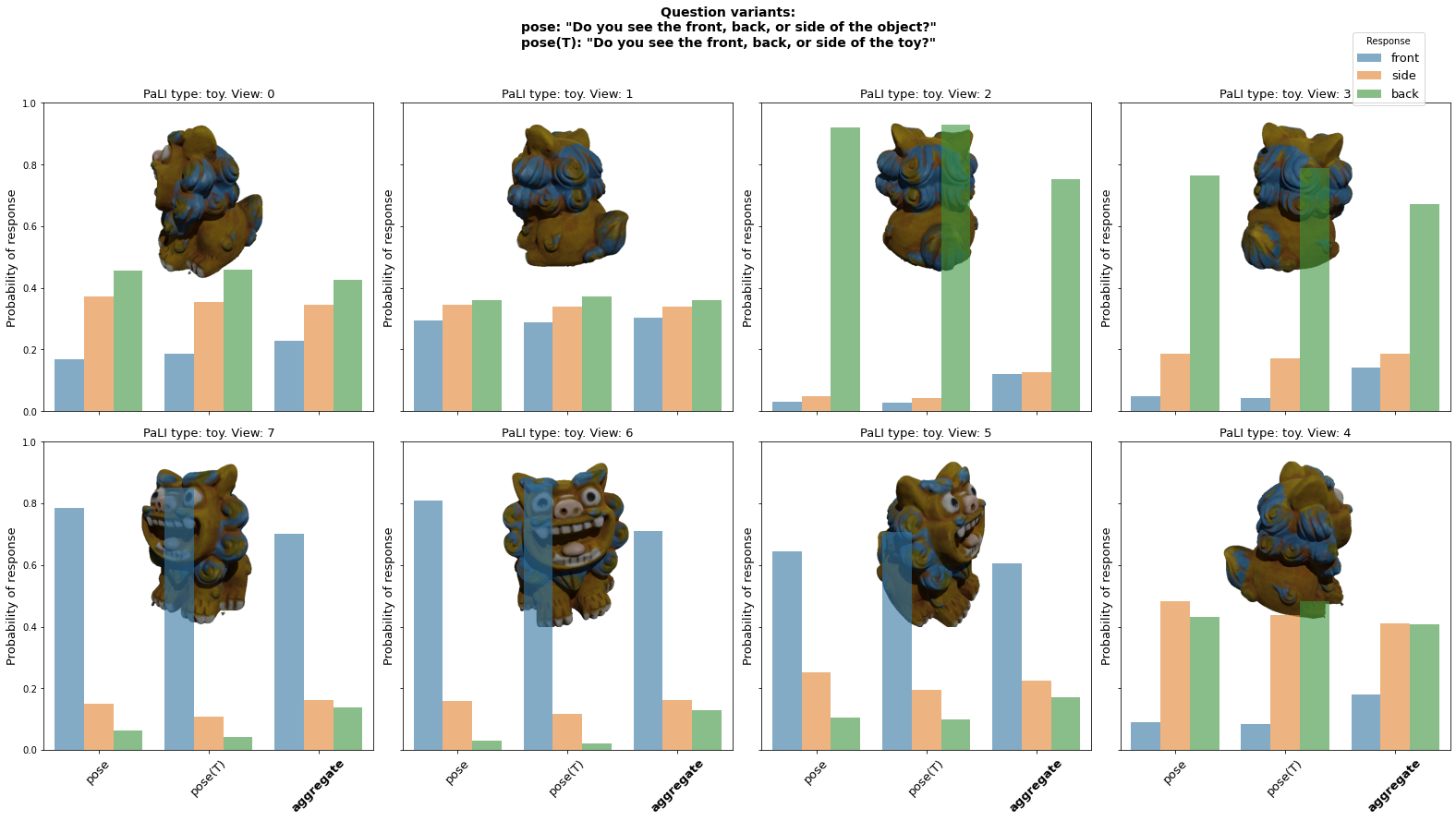}
    \end{subfigure} \hrule

    \begin{subfigure}{\textwidth}
        \centering
        \includegraphics[width=\textwidth]{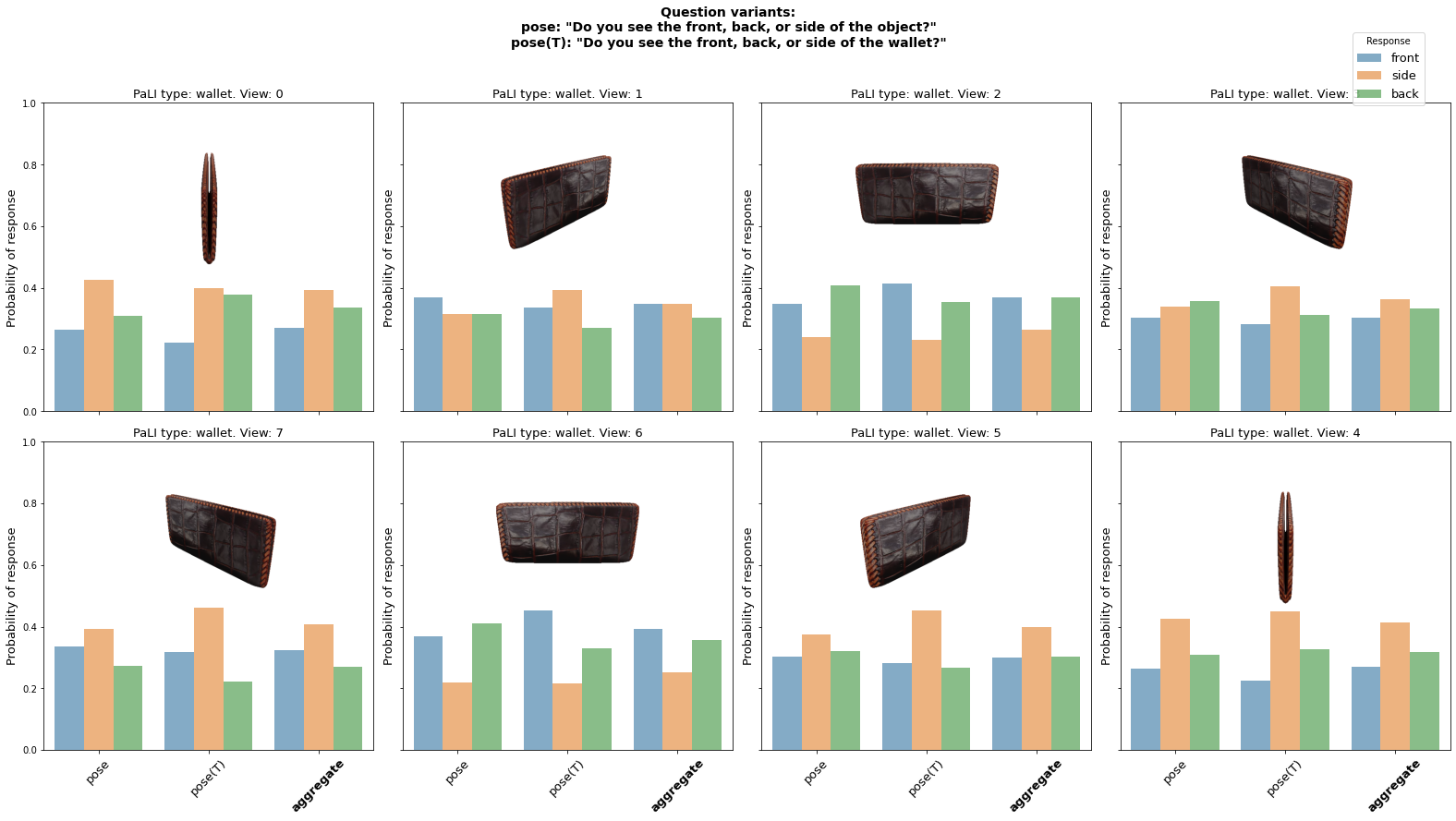}
    \end{subfigure} \hrule
    
    \caption{\textbf{Pose inference.} We show eight views per object to the VLM, asking if it knows which side of the object is visible in each view.}
    \label{fig:pose_inference}
\end{figure*}

\begin{figure*}

    \begin{subfigure}{\textwidth}
        \centering
        \includegraphics[width=\textwidth]{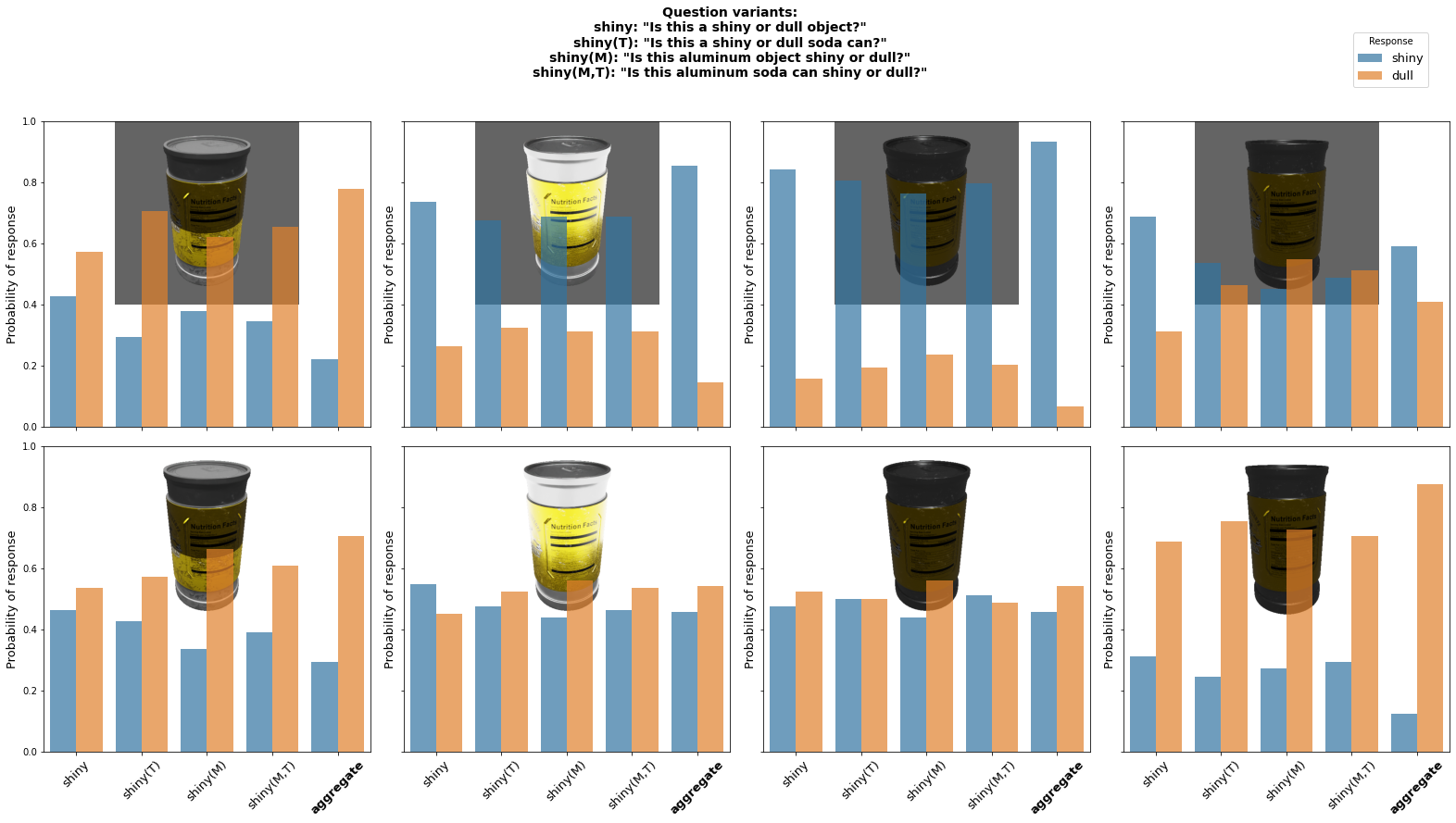}
    \end{subfigure} \hrule

    \caption{\textbf{How lighting/rendering settings affect the inference of shininess.} For a fixed object, we vary the lighting or camera height across columns, keeping the image background color fixed. The first column uses an area light under the object. The second column uses surround lighting. The third and fourth column use the same camera backlight, but vary in camera height.}
    \label{fig:shininess_inference}
\end{figure*}

%% file: material_predictions_figure.tex
\clearpage
\onecolumn

\begin{longtable}{llp{8cm}}
    \caption{\textbf{Material prediction examples on 12 categories from our custom test set.} We show predicted distributions from both VLMs (PaLI-X and BLIP-2) and all five sets of inputs described in Sec \ref{sec:physical_properties}. For brevity, only the top two outputs from each predicted distribution are presented, along with their probabilities in parentheses. We use $t_{cap3d}$ or $t_{pali}$ to denote the type annotations,  $A$ to denote all object views, $p_{vlm}(\hat{m} | .)$ to denote a predicted distribution, and $m$ to denote the true material.}\\
    \endfirsthead
    \caption*{\textbf{Material prediction examples on each category from our custom test set (contd).}}
    \endhead
    \label{tab:material_prediction_examples}
    
    \multirow{13}{*}{\includegraphics[width=0.25\linewidth]{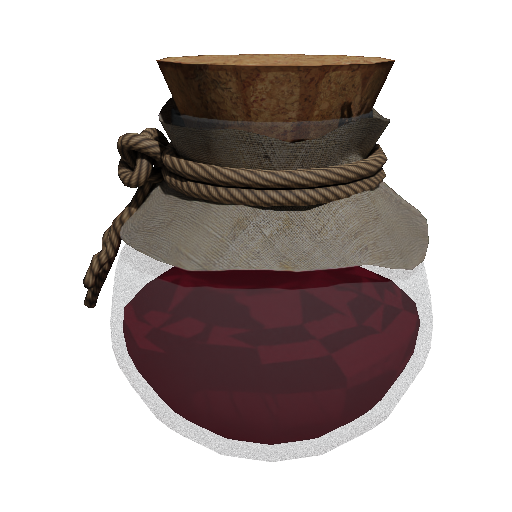}} & $m$ & ``glass'' \\ &$t_{cap3d}$ & ``hat and a jar, both with ropes tied around them'' \\ &$t_{pali}$ & ``potion'' \\ &$p_{pali}(\hat{m} | t_{cap3d})$ & ``cotton'' (0.64), ``can't tell'' (0.36) \\ &$p_{pali}(\hat{m} | t_{pali})$ & ``potion'' (0.35), ``glass'' (0.27) \\ &$p_{pali}(\hat{m} | A)$ & ``cork'' (0.45), ``glass'' (0.19) \\ &$p_{pali}(\hat{m} | t_{cap3d}, A)$ & ``burlap'' (0.44), ``canvas'' (0.30) \\ &$p_{pali}(\hat{m} | t_{pali}, A)$ & ``glass'' (0.67), ``cork'' (0.17) \\ &$p_{blip}(\hat{m} | t_{cap3d})$ & ``straw'' (0.49), ``plastic'' (0.33) \\ &$p_{blip}(\hat{m} | t_{pali})$ & ``a tainted potion made of a tainted potion and a tainted potion'' (0.77), ``a tainted potion made of a tainted potion, and a tainted poti'' (0.14) \\ &$p_{blip}(\hat{m} | A)$ & ``wood'' (0.83), ``rope'' (0.10) \\ &$p_{blip}(\hat{m} | t_{cap3d}, A)$ & ``wood'' (0.68), ``leather'' (0.13) \\ &$p_{blip}(\hat{m} | t_{pali}, A)$ & ``wood'' (0.95), ``stone'' (0.04) \\ \hline
\multirow{13}{*}{\includegraphics[width=0.25\linewidth]{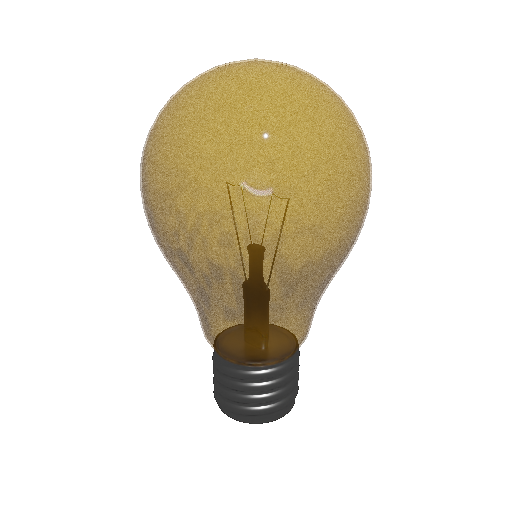}} & $m$ & ``glass'' \\ &$t_{cap3d}$ & ``light bulb'' \\ &$t_{pali}$ & ``light'' \\ &$p_{pali}(\hat{m} | t_{cap3d})$ & ``glass'' (0.77), ``filament'' (0.11) \\ &$p_{pali}(\hat{m} | t_{pali})$ & ``glass'' (0.58), ``light-emitting diode,LED'' (0.13) \\ &$p_{pali}(\hat{m} | A)$ & ``glass'' (0.41), ``brass'' (0.19) \\ &$p_{pali}(\hat{m} | t_{cap3d}, A)$ & ``glass'' (0.60), ``porcelain'' (0.13) \\ &$p_{pali}(\hat{m} | t_{pali}, A)$ & ``glass'' (0.51), ``filament'' (0.14) \\ &$p_{blip}(\hat{m} | t_{cap3d})$ & ``glass'' (0.52), ``filament'' (0.29) \\ &$p_{blip}(\hat{m} | t_{pali})$ & ``light-emitting diodes'' (0.73), ``light-emitting diodes (LEDs)'' (0.20) \\ &$p_{blip}(\hat{m} | A)$ & ``metal'' (0.30), ``3ds max'' (0.22) \\ &$p_{blip}(\hat{m} | t_{cap3d}, A)$ & ``metal'' (0.84), ``gold'' (0.13) \\ &$p_{blip}(\hat{m} | t_{pali}, A)$ & ``metal'' (0.78), ``gold'' (0.14) \\ \hline
\multirow{13}{*}{\includegraphics[width=0.25\linewidth]{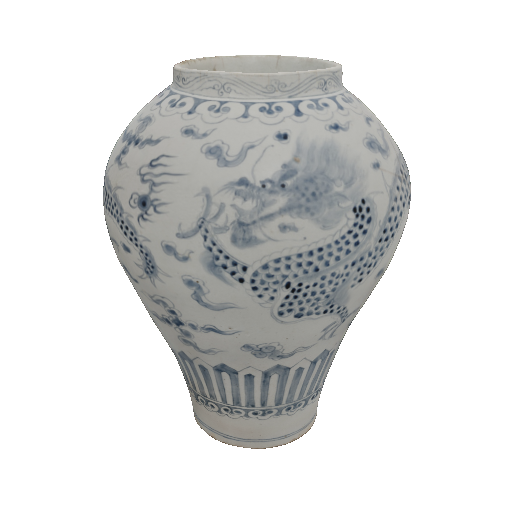}} & $m$ & ``porcelain'' \\ &$t_{cap3d}$ & ``blue and white vase featuring a dragon design'' \\ &$t_{pali}$ & ``vase'' \\ &$p_{pali}(\hat{m} | t_{cap3d})$ & ``ceramic'' (0.38), ``porcelain'' (0.34) \\ &$p_{pali}(\hat{m} | t_{pali})$ & ``ceramic'' (0.35), ``glass'' (0.31) \\ &$p_{pali}(\hat{m} | A)$ & ``faience'' (0.62), ``porcelain'' (0.14) \\ &$p_{pali}(\hat{m} | t_{cap3d}, A)$ & ``ceramic'' (0.38), ``porcelain'' (0.32) \\ &$p_{pali}(\hat{m} | t_{pali}, A)$ & ``faience'' (0.44), ``ceramic'' (0.24) \\ &$p_{blip}(\hat{m} | t_{cap3d})$ & ``porcelain'' (0.65), ``Chinese celadon'' (0.32) \\ &$p_{blip}(\hat{m} | t_{pali})$ & ``Porcelain'' (0.86), ``terracotta'' (0.09) \\ &$p_{blip}(\hat{m} | A)$ & ``porcelain'' (0.83), ``ceramic'' (0.08) \\ &$p_{blip}(\hat{m} | t_{cap3d}, A)$ & ``porcelain'' (0.88), ``china'' (0.12) \\ &$p_{blip}(\hat{m} | t_{pali}, A)$ & ``porcelain'' (0.80), ``china'' (0.07) \\ \hline
\pagebreak \multirow{13}{*}{\includegraphics[width=0.25\linewidth]{figures/material_predictions/43a66abe1d37486d95bd264b04aac7ea.png}} & $m$ & ``porcelain'' \\ &$t_{cap3d}$ & ``small white porcelain vase with colorful floral designs on it'' \\ &$t_{pali}$ & ``inkwell'' \\ &$p_{pali}(\hat{m} | t_{cap3d})$ & ``porcelain'' (0.29), ``faience'' (0.28) \\ &$p_{pali}(\hat{m} | t_{pali})$ & ``glass'' (0.28), ``porcelain'' (0.24) \\ &$p_{pali}(\hat{m} | A)$ & ``faience'' (0.88), ``porcelain'' (0.06) \\ &$p_{pali}(\hat{m} | t_{cap3d}, A)$ & ``faience'' (0.68), ``porcelain'' (0.15) \\ &$p_{pali}(\hat{m} | t_{pali}, A)$ & ``faience'' (0.71), ``porcelain'' (0.16) \\ &$p_{blip}(\hat{m} | t_{cap3d})$ & ``China'' (0.58), ``ceramic'' (0.24) \\ &$p_{blip}(\hat{m} | t_{pali})$ & ``metal'' (0.93), ``metal or plastic'' (0.06) \\ &$p_{blip}(\hat{m} | A)$ & ``porcelain'' (0.99), ``white porcelain'' (0.01) \\ &$p_{blip}(\hat{m} | t_{cap3d}, A)$ & ``porcelain'' (0.80), ``china'' (0.12) \\ &$p_{blip}(\hat{m} | t_{pali}, A)$ & ``porcelain'' (0.94), ``china'' (0.04) \\ \hline
\multirow{13}{*}{\includegraphics[width=0.25\linewidth]{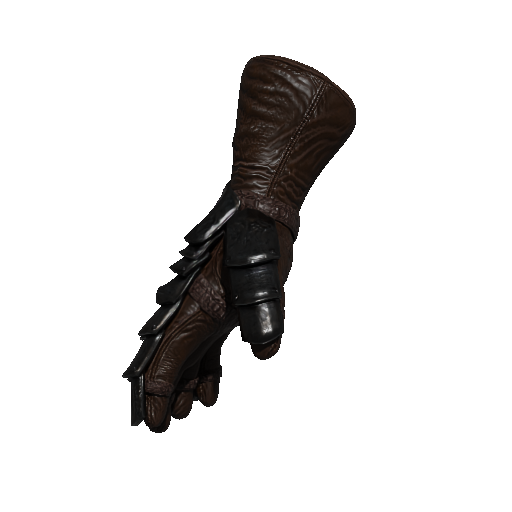}} & $m$ & ``leather'' \\ &$t_{cap3d}$ & ``armored leather gloves and a brown leather boot'' \\ &$t_{pali}$ & ``glove'' \\ &$p_{pali}(\hat{m} | t_{cap3d})$ & ``leather'' (0.83), ``cowhide'' (0.08) \\ &$p_{pali}(\hat{m} | t_{pali})$ & ``leather'' (0.34), ``cotton'' (0.21) \\ &$p_{pali}(\hat{m} | A)$ & ``leather'' (0.69), ``armor plate,armour plate,armor plating,plate armor,plate armour'' (0.08) \\ &$p_{pali}(\hat{m} | t_{cap3d}, A)$ & ``leather'' (0.80), ``cowhide'' (0.07) \\ &$p_{pali}(\hat{m} | t_{pali}, A)$ & ``leather'' (0.84), ``nylon'' (0.04) \\ &$p_{blip}(\hat{m} | t_{cap3d})$ & ``leather'' (1.00) \\ &$p_{blip}(\hat{m} | t_{pali})$ & ``leather'' (0.98), ``neoprene'' (0.02) \\ &$p_{blip}(\hat{m} | A)$ & ``leather'' (1.00) \\ &$p_{blip}(\hat{m} | t_{cap3d}, A)$ & ``leather'' (1.00), ``neoprene'' (0.00) \\ &$p_{blip}(\hat{m} | t_{pali}, A)$ & ``leather'' (1.00), ``neoprene'' (0.00) \\ \hline
\multirow{13}{*}{\includegraphics[width=0.25\linewidth]{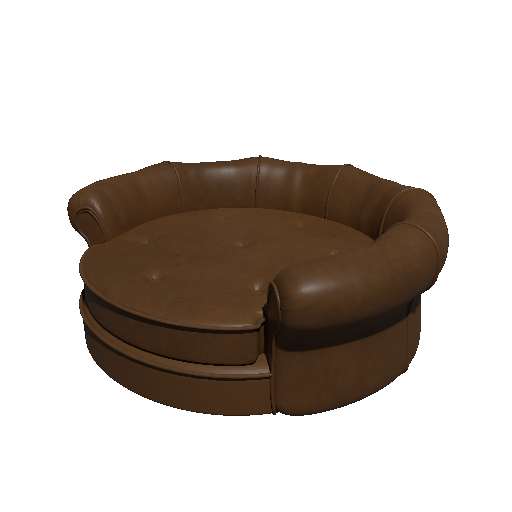}} & $m$ & ``leather'' \\ &$t_{cap3d}$ & ``round tan leather sofa-style dog bed with buttons'' \\ &$t_{pali}$ & ``dog bed'' \\ &$p_{pali}(\hat{m} | t_{cap3d})$ & ``leather'' (0.70), ``suede'' (0.16) \\ &$p_{pali}(\hat{m} | t_{pali})$ & ``foam'' (0.39), ``cotton'' (0.37) \\ &$p_{pali}(\hat{m} | A)$ & ``leather'' (0.81), ``upholstery'' (0.08) \\ &$p_{pali}(\hat{m} | t_{cap3d}, A)$ & ``leather'' (0.87), ``faux leather'' (0.04) \\ &$p_{pali}(\hat{m} | t_{pali}, A)$ & ``leather'' (0.89), ``faux leather'' (0.04) \\ &$p_{blip}(\hat{m} | t_{cap3d})$ & ``faux leather'' (0.87), ``faux-leather'' (0.13) \\ &$p_{blip}(\hat{m} | t_{pali})$ & ``a soft fabric, such as cotton, wool, linen, or a combination of the two'' (1.00), ``a soft fabric, such as cotton, wool, linen, or a synthetic material, such as acetate or polypropylene'' (0.00) \\ &$p_{blip}(\hat{m} | A)$ & ``leather'' (1.00) \\ &$p_{blip}(\hat{m} | t_{cap3d}, A)$ & ``leather'' (0.79), ``3d model'' (0.12) \\ &$p_{blip}(\hat{m} | t_{pali}, A)$ & ``leather'' (1.00) \\ \hline
\pagebreak \multirow{13}{*}{\includegraphics[width=0.25\linewidth]{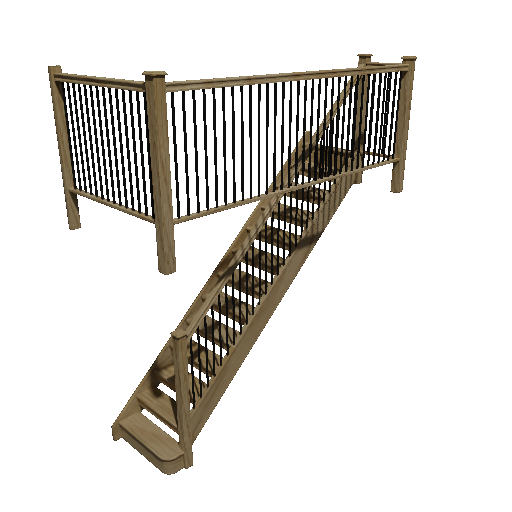}} & $m$ & ``oak'' \\ &$t_{cap3d}$ & ``wooden staircase with metal railings'' \\ &$t_{pali}$ & ``bannister'' \\ &$p_{pali}(\hat{m} | t_{cap3d})$ & ``wood'' (0.46), ``steel'' (0.19) \\ &$p_{pali}(\hat{m} | t_{pali})$ & ``wood'' (0.80), ``marble'' (0.07) \\ &$p_{pali}(\hat{m} | A)$ & ``wood'' (0.78), ``timber'' (0.06) \\ &$p_{pali}(\hat{m} | t_{cap3d}, A)$ & ``wood'' (0.46), ``oak'' (0.31) \\ &$p_{pali}(\hat{m} | t_{pali}, A)$ & ``wood'' (0.50), ``metal'' (0.26) \\ &$p_{blip}(\hat{m} | t_{cap3d})$ & ``a wooden staircase with metal railings'' (1.00), ``a wooden staircase with metal railings is called a balustrade'' (0.00) \\ &$p_{blip}(\hat{m} | t_{pali})$ & ``wood'' (0.73), ``wooden'' (0.27) \\ &$p_{blip}(\hat{m} | A)$ & ``wood'' (0.99), ``wooden railings'' (0.00) \\ &$p_{blip}(\hat{m} | t_{cap3d}, A)$ & ``wood'' (0.98), ``wooden staircase with metal railings'' (0.02) \\ &$p_{blip}(\hat{m} | t_{pali}, A)$ & ``wood'' (0.97), ``wooden'' (0.03) \\ \hline
\multirow{13}{*}{\includegraphics[width=0.25\linewidth]{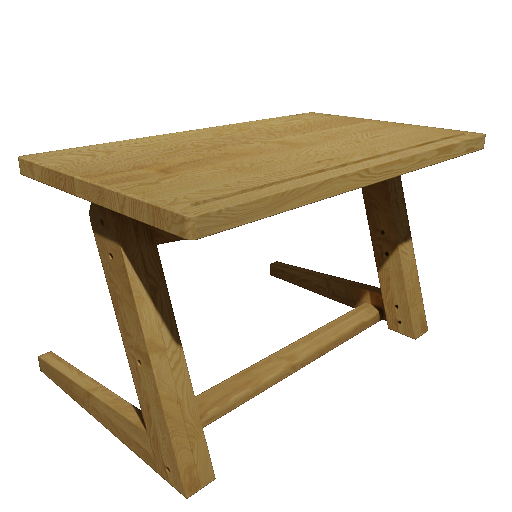}} & $m$ & ``oak'' \\ &$t_{cap3d}$ & ``small wooden table with two legs and a slanted top'' \\ &$t_{pali}$ & ``trestle table'' \\ &$p_{pali}(\hat{m} | t_{cap3d})$ & ``wood'' (0.65), ``oak'' (0.24) \\ &$p_{pali}(\hat{m} | t_{pali})$ & ``wood'' (0.88), ``timber'' (0.05) \\ &$p_{pali}(\hat{m} | A)$ & ``wood'' (0.63), ``oak'' (0.15) \\ &$p_{pali}(\hat{m} | t_{cap3d}, A)$ & ``wood'' (0.43), ``oak'' (0.42) \\ &$p_{pali}(\hat{m} | t_{pali}, A)$ & ``wood'' (0.70), ``oak'' (0.20) \\ &$p_{blip}(\hat{m} | t_{cap3d})$ & ``trestle table'' (0.80), ``a trestle table'' (0.20) \\ &$p_{blip}(\hat{m} | t_{pali})$ & ``wood'' (0.98), ``wooden trestle'' (0.02) \\ &$p_{blip}(\hat{m} | A)$ & ``wood'' (0.97), ``wooden'' (0.03) \\ &$p_{blip}(\hat{m} | t_{cap3d}, A)$ & ``wood'' (0.90), ``solid wood'' (0.09) \\ &$p_{blip}(\hat{m} | t_{pali}, A)$ & ``wood'' (0.99), ``solid wood'' (0.01) \\ \hline
\multirow{13}{*}{\includegraphics[width=0.25\linewidth]{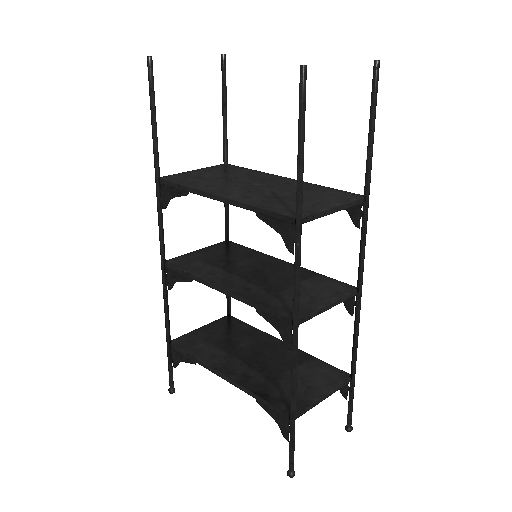}} & $m$ & ``metal'' \\ &$t_{cap3d}$ & ``three-tier metal shelving unit'' \\ &$t_{pali}$ & ``bookshelf'' \\ &$p_{pali}(\hat{m} | t_{cap3d})$ & ``steel'' (0.41), ``metal'' (0.29) \\ &$p_{pali}(\hat{m} | t_{pali})$ & ``wood'' (0.91), ``metal'' (0.03) \\ &$p_{pali}(\hat{m} | A)$ & ``metal'' (0.42), ``steel'' (0.36) \\ &$p_{pali}(\hat{m} | t_{cap3d}, A)$ & ``steel'' (0.49), ``metal'' (0.29) \\ &$p_{pali}(\hat{m} | t_{pali}, A)$ & ``metal'' (0.59), ``steel'' (0.20) \\ &$p_{blip}(\hat{m} | t_{cap3d})$ & ``steel'' (0.99), ``steel or stainless steel'' (0.01) \\ &$p_{blip}(\hat{m} | t_{pali})$ & ``wood'' (0.98), ``reclaimed wood'' (0.02) \\ &$p_{blip}(\hat{m} | A)$ & ``metal'' (0.72), ``steel'' (0.21) \\ &$p_{blip}(\hat{m} | t_{cap3d}, A)$ & ``black metal'' (0.43), ``steel'' (0.32) \\ &$p_{blip}(\hat{m} | t_{pali}, A)$ & ``metal'' (0.68), ``steel'' (0.20) \\ \hline
\multirow{13}{*}{\includegraphics[width=0.25\linewidth]{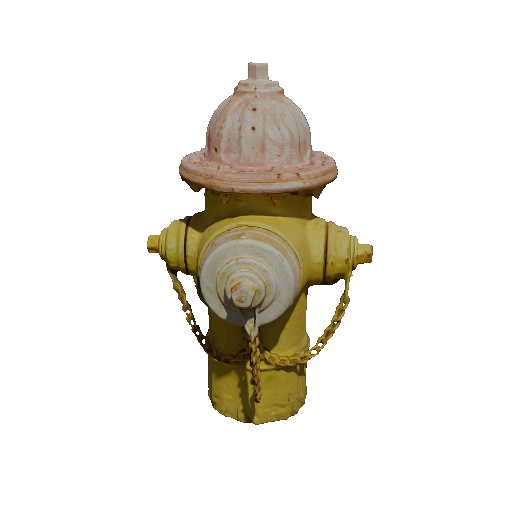}} & $m$ & ``metal'' \\ &$t_{cap3d}$ & ``yellow fire hydrant'' \\ &$t_{pali}$ & ``fire hydrant'' \\ &$p_{pali}(\hat{m} | t_{cap3d})$ & ``metal'' (0.37), ``steel'' (0.24) \\ &$p_{pali}(\hat{m} | t_{pali})$ & ``metal'' (0.32), ``steel'' (0.25) \\ &$p_{pali}(\hat{m} | A)$ & ``iron'' (0.31), ``metal'' (0.17) \\ &$p_{pali}(\hat{m} | t_{cap3d}, A)$ & ``metal'' (0.37), ``steel'' (0.19) \\ &$p_{pali}(\hat{m} | t_{pali}, A)$ & ``metal'' (0.32), ``iron'' (0.21) \\ &$p_{blip}(\hat{m} | t_{cap3d})$ & ``cast iron'' (0.91), ``cast-aluminum'' (0.09) \\ &$p_{blip}(\hat{m} | t_{pali})$ & ``a fire hydrant is a device used to extinguish a fire.'' (0.98), ``a fire hydrant is a device used to extinguish a fire by means of a pressurized stream of water'' (0.01) \\ &$p_{blip}(\hat{m} | A)$ & ``plastic'' (0.35), ``3ds max'' (0.24) \\ &$p_{blip}(\hat{m} | t_{cap3d}, A)$ & ``metal'' (0.79), ``plastic'' (0.20) \\ &$p_{blip}(\hat{m} | t_{pali}, A)$ & ``metal'' (0.70), ``plastic'' (0.17) \\ \hline
\multirow{13}{*}{\includegraphics[width=0.25\linewidth]{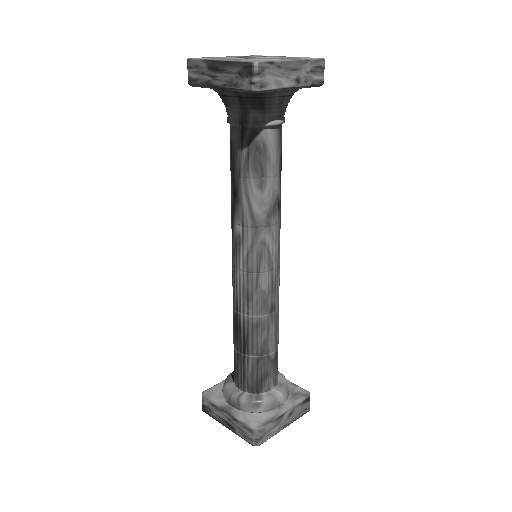}} & $m$ & ``marble'' \\ &$t_{cap3d}$ & ``white marble column'' \\ &$t_{pali}$ & ``pedestal'' \\ &$p_{pali}(\hat{m} | t_{cap3d})$ & ``marble'' (0.75), ``limestone'' (0.09) \\ &$p_{pali}(\hat{m} | t_{pali})$ & ``marble'' (0.44), ``stone'' (0.31) \\ &$p_{pali}(\hat{m} | A)$ & ``marble'' (0.69), ``stone'' (0.17) \\ &$p_{pali}(\hat{m} | t_{cap3d}, A)$ & ``marble'' (0.73), ``carrara'' (0.10) \\ &$p_{pali}(\hat{m} | t_{pali}, A)$ & ``marble'' (0.67), ``stone'' (0.21) \\ &$p_{blip}(\hat{m} | t_{cap3d})$ & ``marble'' (1.00) \\ &$p_{blip}(\hat{m} | t_{pali})$ & ``marble'' (1.00) \\ &$p_{blip}(\hat{m} | A)$ & ``marble'' (0.96), ``wood'' (0.04) \\ &$p_{blip}(\hat{m} | t_{cap3d}, A)$ & ``marble'' (0.73), ``white marble'' (0.27) \\ &$p_{blip}(\hat{m} | t_{pali}, A)$ & ``marble'' (0.95), ``wood'' (0.04) \\ \hline
\multirow{13}{*}{\includegraphics[width=0.25\linewidth]{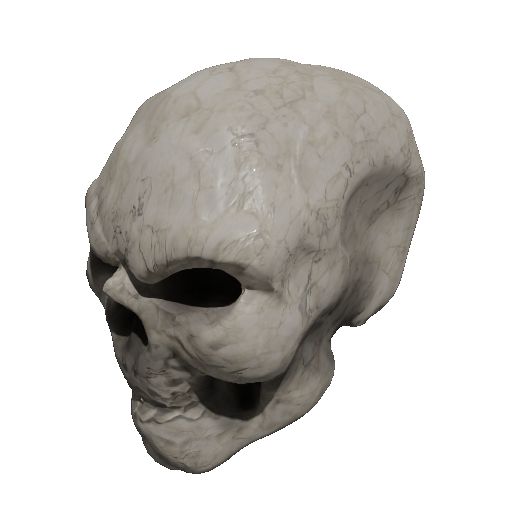}} & $m$ & ``marble'' \\ &$t_{cap3d}$ & ``white marble skull'' \\ &$t_{pali}$ & ``skull'' \\ &$p_{pali}(\hat{m} | t_{cap3d})$ & ``marble'' (0.79), ``porcelain'' (0.09) \\ &$p_{pali}(\hat{m} | t_{pali})$ & ``bone'' (0.75), ``bones'' (0.09) \\ &$p_{pali}(\hat{m} | A)$ & ``clay'' (0.35), ``marble'' (0.22) \\ &$p_{pali}(\hat{m} | t_{cap3d}, A)$ & ``marble'' (0.55), ``clay'' (0.20) \\ &$p_{pali}(\hat{m} | t_{pali}, A)$ & ``clay'' (0.33), ``marble'' (0.27) \\ &$p_{blip}(\hat{m} | t_{cap3d})$ & ``limestone'' (0.68), ``marble'' (0.32) \\ &$p_{blip}(\hat{m} | t_{pali})$ & ``calcium phosphate'' (0.83), ``calcareous limestone'' (0.08) \\ &$p_{blip}(\hat{m} | A)$ & ``marble'' (0.81), ``white marble'' (0.10) \\ &$p_{blip}(\hat{m} | t_{cap3d}, A)$ & ``white marble'' (0.85), ``marble'' (0.07) \\ &$p_{blip}(\hat{m} | t_{pali}, A)$ & ``marble'' (0.43), ``limestone'' (0.36) \\ \hline
\multirow{13}{*}{\includegraphics[width=0.25\linewidth]{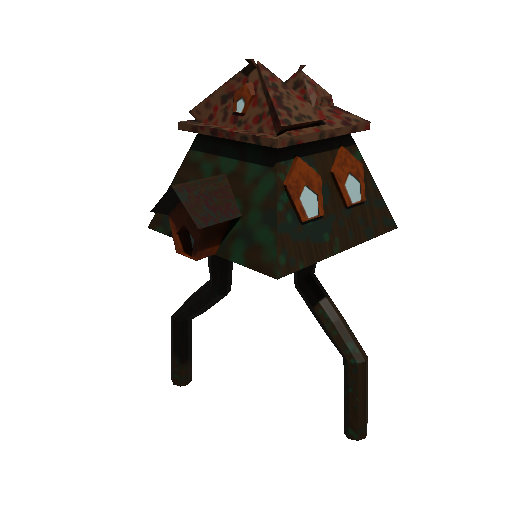}} & $m$ & ``wood'' \\ &$t_{cap3d}$ & ``small metal house with a roof and legs'' \\ &$t_{pali}$ & ``birdhouse'' \\ &$p_{pali}(\hat{m} | t_{cap3d})$ & ``aluminum'' (0.48), ``steel'' (0.34) \\ &$p_{pali}(\hat{m} | t_{pali})$ & ``wood'' (0.75), ``clay'' (0.10) \\ &$p_{pali}(\hat{m} | A)$ & ``wood'' (0.42), ``copper'' (0.23) \\ &$p_{pali}(\hat{m} | t_{cap3d}, A)$ & ``steel'' (0.21), ``iron'' (0.19) \\ &$p_{pali}(\hat{m} | t_{pali}, A)$ & ``wood'' (0.61), ``metal'' (0.17) \\ &$p_{blip}(\hat{m} | t_{cap3d})$ & ``a styrofoam styrofoam styrofoam sty'' (0.36), ``a styrofoam styrofoam styrofoam sandwich'' (0.33) \\ &$p_{blip}(\hat{m} | t_{pali})$ & ``wood'' (0.68), ``Cedar'' (0.31) \\ &$p_{blip}(\hat{m} | A)$ & ``metal'' (0.86), ``wood'' (0.07) \\ &$p_{blip}(\hat{m} | t_{cap3d}, A)$ & ``3d model'' (0.59), ``rusty metal'' (0.21) \\ &$p_{blip}(\hat{m} | t_{pali}, A)$ & ``metal'' (0.61), ``wood'' (0.33) \\ \hline
\pagebreak \multirow{13}{*}{\includegraphics[width=0.25\linewidth]{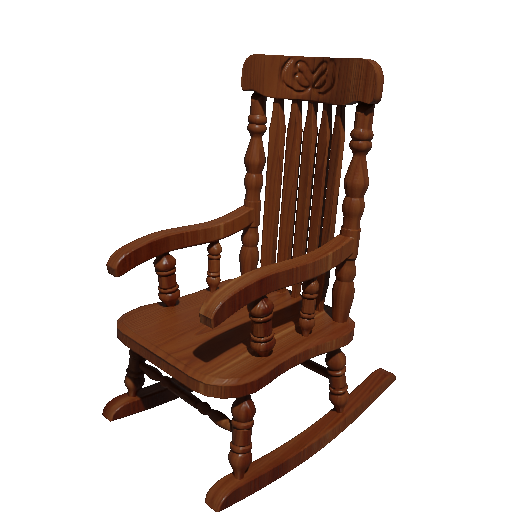}} & $m$ & ``wood'' \\ &$t_{cap3d}$ & ``wooden rocking chair'' \\ &$t_{pali}$ & ``rocking chair'' \\ &$p_{pali}(\hat{m} | t_{cap3d})$ & ``wood'' (0.58), ``oak'' (0.22) \\ &$p_{pali}(\hat{m} | t_{pali})$ & ``wood'' (0.81), ``wicker'' (0.08) \\ &$p_{pali}(\hat{m} | A)$ & ``wood'' (0.88), ``rattan'' (0.04) \\ &$p_{pali}(\hat{m} | t_{cap3d}, A)$ & ``oak'' (0.40), ``wood'' (0.22) \\ &$p_{pali}(\hat{m} | t_{pali}, A)$ & ``wood'' (0.93), ``mahogany'' (0.02) \\ &$p_{blip}(\hat{m} | t_{cap3d})$ & ``wood'' (0.96), ``rattan'' (0.04) \\ &$p_{blip}(\hat{m} | t_{pali})$ & ``wood'' (0.97), ``wooden rocking chair'' (0.03) \\ &$p_{blip}(\hat{m} | A)$ & ``wood'' (0.98), ``wooden'' (0.01) \\ &$p_{blip}(\hat{m} | t_{cap3d}, A)$ & ``wood'' (0.96), ``wooden rocking chair'' (0.04) \\ &$p_{blip}(\hat{m} | t_{pali}, A)$ & ``wood'' (1.00), ``wooden rocking chair'' (0.00) \\ \hline
\multirow{13}{*}{\includegraphics[width=0.25\linewidth]{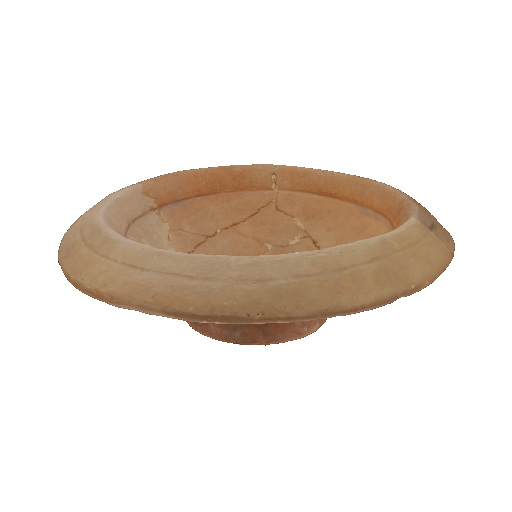}} & $m$ & ``ceramic'' \\ &$t_{cap3d}$ & ``terracotta bowl with a curved top, flat bottom'' \\ &$t_{pali}$ & ``tray'' \\ &$p_{pali}(\hat{m} | t_{cap3d})$ & ``ceramic'' (0.40), ``stoneware'' (0.25) \\ &$p_{pali}(\hat{m} | t_{pali})$ & ``wood'' (0.28), ``ceramic'' (0.24) \\ &$p_{pali}(\hat{m} | A)$ & ``clay'' (0.33), ``stoneware'' (0.28) \\ &$p_{pali}(\hat{m} | t_{cap3d}, A)$ & ``clay'' (0.47), ``ceramic'' (0.18) \\ &$p_{pali}(\hat{m} | t_{pali}, A)$ & ``clay'' (0.41), ``stoneware'' (0.19) \\ &$p_{blip}(\hat{m} | t_{cap3d})$ & ``earthenware'' (0.55), ``terracotta'' (0.31) \\ &$p_{blip}(\hat{m} | t_{pali})$ & ``stainless steel'' (1.00), ``stainless steel or stainless steel-alloys'' (0.00) \\ &$p_{blip}(\hat{m} | A)$ & ``clay'' (0.92), ``terracotta'' (0.05) \\ &$p_{blip}(\hat{m} | t_{cap3d}, A)$ & ``clay'' (0.64), ``terracotta'' (0.35) \\ &$p_{blip}(\hat{m} | t_{pali}, A)$ & ``clay'' (0.94), ``terracotta'' (0.05) \\ \hline
\multirow{13}{*}{\includegraphics[width=0.25\linewidth]{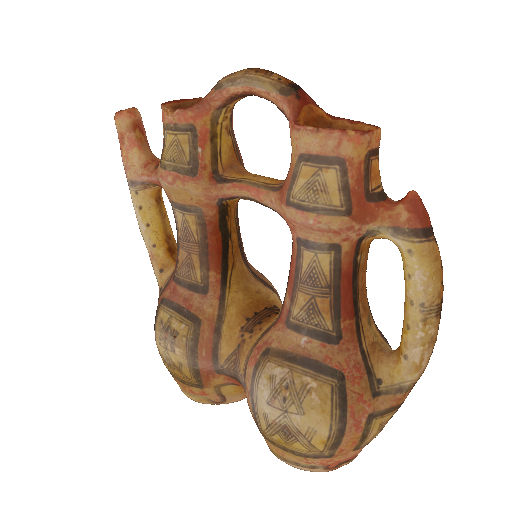}} & $m$ & ``ceramic'' \\ &$t_{cap3d}$ & ``vase with two handles and intricate designs'' \\ &$t_{pali}$ & ``jug'' \\ &$p_{pali}(\hat{m} | t_{cap3d})$ & ``ceramic'' (0.38), ``porcelain'' (0.27) \\ &$p_{pali}(\hat{m} | t_{pali})$ & ``glass'' (0.56), ``porcelain'' (0.17) \\ &$p_{pali}(\hat{m} | A)$ & ``stoneware'' (0.29), ``clay'' (0.23) \\ &$p_{pali}(\hat{m} | t_{cap3d}, A)$ & ``clay'' (0.36), ``pottery'' (0.27) \\ &$p_{pali}(\hat{m} | t_{pali}, A)$ & ``ceramic'' (0.29), ``clay'' (0.27) \\ &$p_{blip}(\hat{m} | t_{cap3d})$ & ``Chinese celadon'' (0.97), ``Chinese lacquerware'' (0.02) \\ &$p_{blip}(\hat{m} | t_{pali})$ & ``clay'' (0.87), ``tin'' (0.13) \\ &$p_{blip}(\hat{m} | A)$ & ``clay'' (0.73), ``ceramic'' (0.27) \\ &$p_{blip}(\hat{m} | t_{cap3d}, A)$ & ``clay'' (0.76), ``ceramic'' (0.24) \\ &$p_{blip}(\hat{m} | t_{pali}, A)$ & ``clay'' (0.87), ``ceramic'' (0.13) \\ \hline
\multirow{13}{*}{\includegraphics[width=0.25\linewidth]{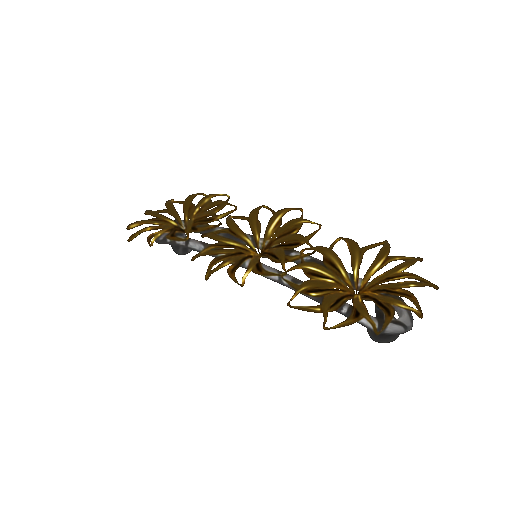}} & $m$ & ``gold'' \\ &$t_{cap3d}$ & ``gold flower ring featuring a yellow and white flower design'' \\ &$t_{pali}$ & ``hair slide'' \\ &$p_{pali}(\hat{m} | t_{cap3d})$ & ``gold'' (0.74), ``sterling silver'' (0.09) \\ &$p_{pali}(\hat{m} | t_{pali})$ & ``plastic'' (0.44), ``rubber'' (0.24) \\ &$p_{pali}(\hat{m} | A)$ & ``gold plate'' (0.33), ``brass'' (0.31) \\ &$p_{pali}(\hat{m} | t_{cap3d}, A)$ & ``gold'' (0.40), ``brass'' (0.24) \\ &$p_{pali}(\hat{m} | t_{pali}, A)$ & ``brass'' (0.23), ``metal'' (0.23) \\ &$p_{blip}(\hat{m} | t_{cap3d})$ & ``14K yellow gold'' (0.35), ``18k white gold'' (0.34) \\ &$p_{blip}(\hat{m} | t_{pali})$ & ``plastic'' (0.88), ``acetate'' (0.12) \\ &$p_{blip}(\hat{m} | A)$ & ``gold'' (0.65), ``metal'' (0.32) \\ &$p_{blip}(\hat{m} | t_{cap3d}, A)$ & ``gold'' (0.59), ``3d model'' (0.15) \\ &$p_{blip}(\hat{m} | t_{pali}, A)$ & ``gold'' (0.72), ``metal'' (0.22) \\ \hline
\multirow{13}{*}{\includegraphics[width=0.25\linewidth]{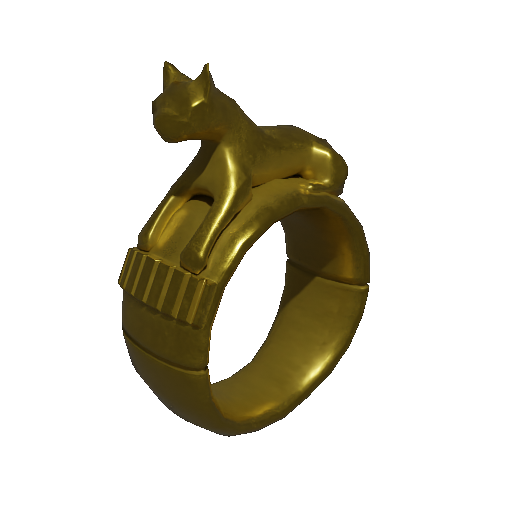}} & $m$ & ``gold'' \\ &$t_{cap3d}$ & ``gold Egyptian cat ring'' \\ &$t_{pali}$ & ``ring'' \\ &$p_{pali}(\hat{m} | t_{cap3d})$ & ``gold'' (0.68), ``gold plate'' (0.11) \\ &$p_{pali}(\hat{m} | t_{pali})$ & ``gold'' (0.74), ``brass'' (0.10) \\ &$p_{pali}(\hat{m} | A)$ & ``gold'' (0.63), ``gold plate'' (0.20) \\ &$p_{pali}(\hat{m} | t_{cap3d}, A)$ & ``gold'' (0.78), ``brass'' (0.10) \\ &$p_{pali}(\hat{m} | t_{pali}, A)$ & ``gold'' (0.82), ``brass'' (0.09) \\ &$p_{blip}(\hat{m} | t_{cap3d})$ & ``gold'' (1.00), ``gold-plated tibetan calfskin'' (0.00) \\ &$p_{blip}(\hat{m} | t_{pali})$ & ``precious metals, such as gold, silver, platinum, palladium, and rhodium'' (0.93), ``precious metals, such as gold, silver, platinum, palladium, rhodium, and tin'' (0.03) \\ &$p_{blip}(\hat{m} | A)$ & ``gold'' (1.00), ``gold 3d printed'' (0.00) \\ &$p_{blip}(\hat{m} | t_{cap3d}, A)$ & ``gold'' (0.90), ``3d printed'' (0.10) \\ &$p_{blip}(\hat{m} | t_{pali}, A)$ & ``gold'' (1.00) \\ \hline
\multirow{13}{*}{\includegraphics[width=0.25\linewidth]{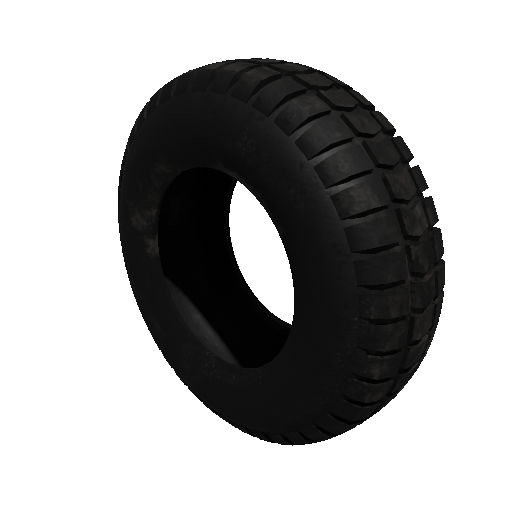}} & $m$ & ``rubber'' \\ &$t_{cap3d}$ & ``tire'' \\ &$t_{pali}$ & ``tire'' \\ &$p_{pali}(\hat{m} | t_{cap3d})$ & ``rubber'' (0.99), ``rubber and steel'' (0.00) \\ &$p_{pali}(\hat{m} | t_{pali})$ & ``rubber'' (0.99), ``rubber and steel'' (0.00) \\ &$p_{pali}(\hat{m} | A)$ & ``rubber'' (0.96), ``blacktop,blacktopping'' (0.02) \\ &$p_{pali}(\hat{m} | t_{cap3d}, A)$ & ``rubber'' (0.97), ``black rubber'' (0.01) \\ &$p_{pali}(\hat{m} | t_{pali}, A)$ & ``rubber'' (0.97), ``black rubber'' (0.01) \\ &$p_{blip}(\hat{m} | t_{cap3d})$ & ``rubber'' (0.90), ``pneumatic tires'' (0.09) \\ &$p_{blip}(\hat{m} | t_{pali})$ & ``rubber'' (0.73), ``Rubber'' (0.27) \\ &$p_{blip}(\hat{m} | A)$ & ``rubber'' (0.87), ``black rubber'' (0.10) \\ &$p_{blip}(\hat{m} | t_{cap3d}, A)$ & ``rubber'' (0.93), ``black rubber'' (0.06) \\ &$p_{blip}(\hat{m} | t_{pali}, A)$ & ``rubber'' (0.91), ``black rubber'' (0.09) \\ \hline
\multirow{13}{*}{\includegraphics[width=0.25\linewidth]{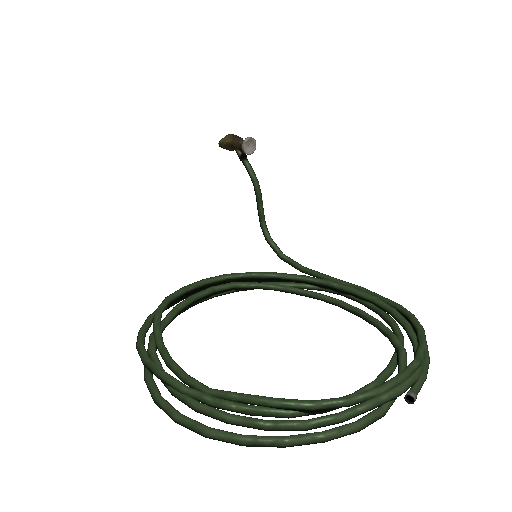}} & $m$ & ``rubber'' \\ &$t_{cap3d}$ & ``green coiled cable with a white plug and attached earbud'' \\ &$t_{pali}$ & ``hose'' \\ &$p_{pali}(\hat{m} | t_{cap3d})$ & ``nylon'' (0.44), ``plastic'' (0.36) \\ &$p_{pali}(\hat{m} | t_{pali})$ & ``rubber'' (0.86), ``plastic'' (0.05) \\ &$p_{pali}(\hat{m} | A)$ & ``hose'' (0.48), ``rubber'' (0.20) \\ &$p_{pali}(\hat{m} | t_{cap3d}, A)$ & ``rubber'' (0.38), ``plastic'' (0.19) \\ &$p_{pali}(\hat{m} | t_{pali}, A)$ & ``rubber'' (0.70), ``plastic'' (0.15) \\ &$p_{blip}(\hat{m} | t_{cap3d})$ & ``tin-alloy'' (0.79), ``tin-plated copper'' (0.20) \\ &$p_{blip}(\hat{m} | t_{pali})$ & ``rubber'' (0.95), ``PTFE'' (0.05) \\ &$p_{blip}(\hat{m} | A)$ & ``wire'' (0.33), ``metal'' (0.26) \\ &$p_{blip}(\hat{m} | t_{cap3d}, A)$ & ``teflon'' (0.92), ``stranded copper'' (0.05) \\ &$p_{blip}(\hat{m} | t_{pali}, A)$ & ``plastic'' (0.36), ``pvc'' (0.23) \\ \hline
\pagebreak \multirow{13}{*}{\includegraphics[width=0.25\linewidth]{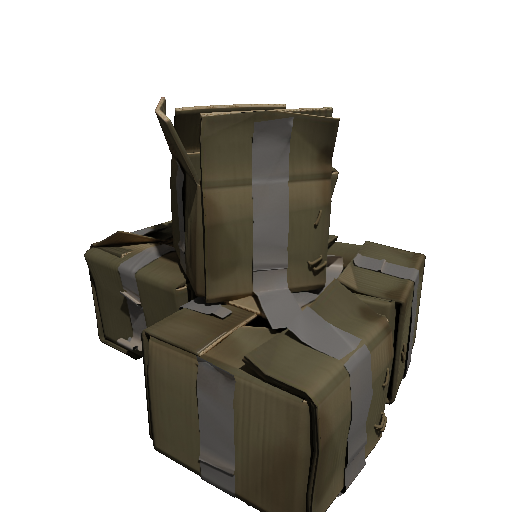}} & $m$ & ``cardboard'' \\ &$t_{cap3d}$ & ``stack of brown cardboard boxes with white tape on them'' \\ &$t_{pali}$ & ``packing box'' \\ &$p_{pali}(\hat{m} | t_{cap3d})$ & ``cardboard'' (0.64), ``paper'' (0.30) \\ &$p_{pali}(\hat{m} | t_{pali})$ & ``cardboard'' (0.74), ``paper'' (0.13) \\ &$p_{pali}(\hat{m} | A)$ & ``cardboard'' (0.52), ``cellulose tape,Scotch tape,Sellotape'' (0.17) \\ &$p_{pali}(\hat{m} | t_{cap3d}, A)$ & ``cardboard'' (0.67), ``paper'' (0.13) \\ &$p_{pali}(\hat{m} | t_{pali}, A)$ & ``cardboard'' (0.82), ``corrugated cardboard'' (0.06) \\ &$p_{blip}(\hat{m} | t_{cap3d})$ & ``shipping cartons'' (1.00), ``a receptacle for the shipment of goods'' (0.00) \\ &$p_{blip}(\hat{m} | t_{pali})$ & ``cardboard'' (0.65), ``paper'' (0.27) \\ &$p_{blip}(\hat{m} | A)$ & ``cardboard'' (1.00), ``styrofoam'' (0.00) \\ &$p_{blip}(\hat{m} | t_{cap3d}, A)$ & ``cardboard'' (0.96), ``3d model'' (0.01) \\ &$p_{blip}(\hat{m} | t_{pali}, A)$ & ``cardboard'' (0.99), ``paper'' (0.01) \\ \hline
\multirow{13}{*}{\includegraphics[width=0.25\linewidth]{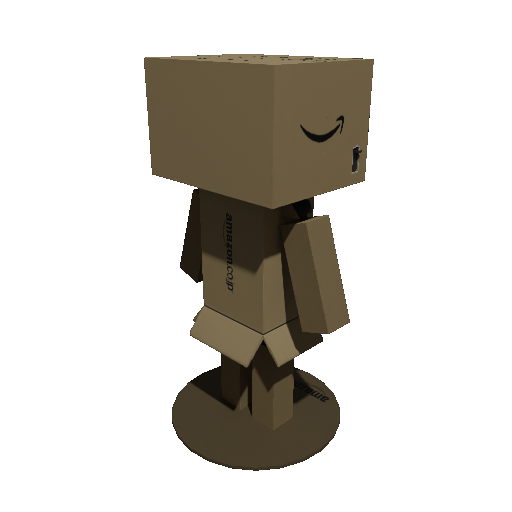}} & $m$ & ``cardboard'' \\ &$t_{cap3d}$ & ``cardboard Amazon robot toy with logo'' \\ &$t_{pali}$ & ``carton'' \\ &$p_{pali}(\hat{m} | t_{cap3d})$ & ``cardboard'' (0.57), ``paper'' (0.32) \\ &$p_{pali}(\hat{m} | t_{pali})$ & ``cardboard'' (0.47), ``paper'' (0.45) \\ &$p_{pali}(\hat{m} | A)$ & ``cardboard'' (0.80), ``carton'' (0.13) \\ &$p_{pali}(\hat{m} | t_{cap3d}, A)$ & ``cardboard'' (0.73), ``carton'' (0.10) \\ &$p_{pali}(\hat{m} | t_{pali}, A)$ & ``cardboard'' (0.85), ``corrugated cardboard'' (0.06) \\ &$p_{blip}(\hat{m} | t_{cap3d})$ & ``cardboard'' (0.99), ``acetate'' (0.01) \\ &$p_{blip}(\hat{m} | t_{pali})$ & ``paper'' (0.75), ``paperboard'' (0.25) \\ &$p_{blip}(\hat{m} | A)$ & ``cardboard'' (1.00) \\ &$p_{blip}(\hat{m} | t_{cap3d}, A)$ & ``cardboard'' (1.00), ``cardboard, cardboard boxes, cardboard boxes, cardboard boxes, cardboard boxes, cardboard boxes, cardboard boxes, cardboard boxes, cardboard boxes, cardboard boxes, cardboard'' (0.00) \\ &$p_{blip}(\hat{m} | t_{pali}, A)$ & ``cardboard'' (0.99), ``paper'' (0.01) \\ \hline
\multirow{13}{*}{\includegraphics[width=0.25\linewidth]{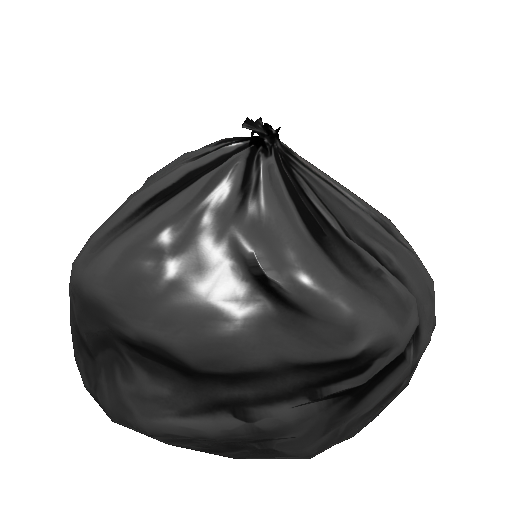}} & $m$ & ``plastic'' \\ &$t_{cap3d}$ & ``large silver trash bag'' \\ &$t_{pali}$ & ``garbage bag'' \\ &$p_{pali}(\hat{m} | t_{cap3d})$ & ``plastic'' (0.45), ``aluminum'' (0.35) \\ &$p_{pali}(\hat{m} | t_{pali})$ & ``plastic'' (0.80), ``polythene'' (0.07) \\ &$p_{pali}(\hat{m} | A)$ & ``garbage'' (0.45), ``plastic'' (0.42) \\ &$p_{pali}(\hat{m} | t_{cap3d}, A)$ & ``plastic'' (0.66), ``cellophane'' (0.13) \\ &$p_{pali}(\hat{m} | t_{pali}, A)$ & ``plastic'' (0.83), ``polythene'' (0.06) \\ &$p_{blip}(\hat{m} | t_{cap3d})$ & ``plastic'' (0.97), ``woven polypropylene'' (0.03) \\ &$p_{blip}(\hat{m} | t_{pali})$ & ``plastic'' (1.00), ``a polyethylene terephthalate (PET) film'' (0.00) \\ &$p_{blip}(\hat{m} | A)$ & ``black plastic'' (0.67), ``3ds max'' (0.15) \\ &$p_{blip}(\hat{m} | t_{cap3d}, A)$ & ``plastic'' (0.83), ``black plastic'' (0.11) \\ &$p_{blip}(\hat{m} | t_{pali}, A)$ & ``plastic'' (0.60), ``black plastic'' (0.38) \\ \hline
\pagebreak \multirow{13}{*}{\includegraphics[width=0.25\linewidth]{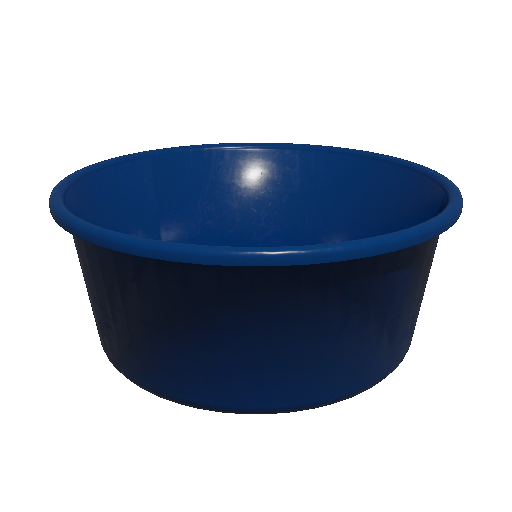}} & $m$ & ``plastic'' \\ &$t_{cap3d}$ & ``blue plastic bowl with a lid'' \\ &$t_{pali}$ & ``washtub'' \\ &$p_{pali}(\hat{m} | t_{cap3d})$ & ``polypropylene'' (0.53), ``plastic'' (0.47) \\ &$p_{pali}(\hat{m} | t_{pali})$ & ``porcelain'' (0.56), ``ceramic'' (0.35) \\ &$p_{pali}(\hat{m} | A)$ & ``plastic'' (0.65), ``polypropylene'' (0.18) \\ &$p_{pali}(\hat{m} | t_{cap3d}, A)$ & ``polypropylene'' (0.58), ``plastic'' (0.24) \\ &$p_{pali}(\hat{m} | t_{pali}, A)$ & ``plastic'' (0.78), ``polypropylene'' (0.10) \\ &$p_{blip}(\hat{m} | t_{cap3d})$ & ``borosilicate glass'' (0.97), ``PP (Polypropylene)'' (0.02) \\ &$p_{blip}(\hat{m} | t_{pali})$ & ``plastic'' (0.90), ``tin'' (0.10) \\ &$p_{blip}(\hat{m} | A)$ & ``plastic'' (1.00), ``polygons'' (0.00) \\ &$p_{blip}(\hat{m} | t_{cap3d}, A)$ & ``plastic'' (0.99), ``polypropylene'' (0.01) \\ &$p_{blip}(\hat{m} | t_{pali}, A)$ & ``plastic'' (1.00) \\ \hline

\end{longtable}

\clearpage
\twocolumn

%% file: icml_main.bbl
\begin{thebibliography}{58}
\providecommand{\natexlab}[1]{#1}
\providecommand{\url}[1]{\texttt{#1}}
\expandafter\ifx\csname urlstyle\endcsname\relax
  \providecommand{\doi}[1]{doi: #1}\else
  \providecommand{\doi}{doi: \begingroup \urlstyle{rm}\Url}\fi

\bibitem[Alayrac et~al.(2022)Alayrac, Donahue, Luc, Miech, Barr, Hasson, Lenc, Mensch, Millican, Reynolds, et~al.]{alayrac2022flamingo}
Alayrac, J.-B., Donahue, J., Luc, P., Miech, A., Barr, I., Hasson, Y., Lenc, K., Mensch, A., Millican, K., Reynolds, M., et~al.
\newblock Flamingo: a visual language model for few-shot learning.
\newblock \emph{NeurIPS}, 35:\penalty0 23716--23736, 2022.

\bibitem[Armeni et~al.(2019)Armeni, He, Gwak, Zamir, Fischer, Malik, and Savarese]{armeni20193d}
Armeni, I., He, Z.-Y., Gwak, J., Zamir, A.~R., Fischer, M., Malik, J., and Savarese, S.
\newblock 3d scene graph: A structure for unified semantics, 3d space, and camera.
\newblock In \emph{Proceedings of the IEEE/CVF international conference on computer vision}, pp.\  5664--5673, 2019.

\bibitem[Azuma et~al.(2022)Azuma, Miyanishi, Kurita, and Kawanabe]{azuma2022scanqa}
Azuma, D., Miyanishi, T., Kurita, S., and Kawanabe, M.
\newblock Scanqa: 3d question answering for spatial scene understanding. 2022 ieee.
\newblock In \emph{CVPR}, pp.\  19107--19117, 2022.

\bibitem[Batra et~al.(2020)Batra, Gokaslan, Kembhavi, Maksymets, Mottaghi, Savva, Toshev, and Wijmans]{batra2020objectnav}
Batra, D., Gokaslan, A., Kembhavi, A., Maksymets, O., Mottaghi, R., Savva, M., Toshev, A., and Wijmans, E.
\newblock Objectnav revisited: On evaluation of embodied agents navigating to objects.
\newblock \emph{arXiv preprint arXiv:2006.13171}, 2020.

\bibitem[Besl \& Jain(1985)Besl and Jain]{besl1985three}
Besl, P.~J. and Jain, R.~C.
\newblock Three-dimensional object recognition.
\newblock \emph{ACM Computing Surveys (CSUR)}, 17\penalty0 (1):\penalty0 75--145, 1985.

\bibitem[Cer et~al.(2018)Cer, Yang, Kong, Hua, Limtiaco, John, Constant, Guajardo-Cespedes, Yuan, Tar, et~al.]{cer2018universal}
Cer, D., Yang, Y., Kong, S.-y., Hua, N., Limtiaco, N., John, R.~S., Constant, N., Guajardo-Cespedes, M., Yuan, S., Tar, C., et~al.
\newblock Universal sentence encoder.
\newblock \emph{arXiv preprint arXiv:1803.11175}, 2018.

\bibitem[Chen et~al.(2020)Chen, Jin, Wang, and Wu]{Chen_2020_CVPR}
Chen, S., Jin, Q., Wang, P., and Wu, Q.
\newblock Say as you wish: Fine-grained control of image caption generation with abstract scene graphs.
\newblock In \emph{Proceedings of the IEEE/CVF Conference on Computer Vision and Pattern Recognition (CVPR)}, June 2020.

\bibitem[Chen et~al.(2023{\natexlab{a}})Chen, Zhu, Chen, Lei, Yu, and Chen]{chen2023end}
Chen, S., Zhu, H., Chen, X., Lei, Y., Yu, G., and Chen, T.
\newblock End-to-end 3d dense captioning with vote2cap-detr.
\newblock In \emph{Proceedings of the IEEE/CVF Conference on Computer Vision and Pattern Recognition}, pp.\  11124--11133, 2023{\natexlab{a}}.

\bibitem[Chen et~al.(2023{\natexlab{b}})Chen, Djolonga, Padlewski, Mustafa, Changpinyo, Wu, Ruiz, Goodman, Wang, Tay, et~al.]{chen2023palix}
Chen, X., Djolonga, J., Padlewski, P., Mustafa, B., Changpinyo, S., Wu, J., Ruiz, C.~R., Goodman, S., Wang, X., Tay, Y., et~al.
\newblock Pali-x: On scaling up a multilingual vision and language model.
\newblock \emph{arXiv preprint arXiv:2305.18565}, 2023{\natexlab{b}}.

\bibitem[Chen et~al.(2023{\natexlab{c}})Chen, Wang, Changpinyo, Piergiovanni, Padlewski, Salz, Goodman, Grycner, Mustafa, Beyer, Kolesnikov, Puigcerver, Ding, Rong, Akbari, Mishra, Xue, Thapliyal, Bradbury, Kuo, Seyedhosseini, Jia, Ayan, Ruiz, Steiner, Angelova, Zhai, Houlsby, and Soricut]{chen2023pali}
Chen, X., Wang, X., Changpinyo, S., Piergiovanni, A., Padlewski, P., Salz, D., Goodman, S., Grycner, A., Mustafa, B., Beyer, L., Kolesnikov, A., Puigcerver, J., Ding, N., Rong, K., Akbari, H., Mishra, G., Xue, L., Thapliyal, A.~V., Bradbury, J., Kuo, W., Seyedhosseini, M., Jia, C., Ayan, B.~K., Ruiz, C.~R., Steiner, A.~P., Angelova, A., Zhai, X., Houlsby, N., and Soricut, R.
\newblock Pa{LI}: A jointly-scaled multilingual language-image model.
\newblock In \emph{ICLR}, 2023{\natexlab{c}}.
\newblock URL \url{https://openreview.net/forum?id=mWVoBz4W0u}.

\bibitem[Chung et~al.(2022)Chung, Hou, Longpre, Zoph, Tay, Fedus, Li, Wang, Dehghani, Brahma, et~al.]{chung2022scaling}
Chung, H.~W., Hou, L., Longpre, S., Zoph, B., Tay, Y., Fedus, W., Li, Y., Wang, X., Dehghani, M., Brahma, S., et~al.
\newblock Scaling instruction-finetuned language models.
\newblock \emph{arXiv preprint arXiv:2210.11416}, 2022.

\bibitem[Collins et~al.(2022)Collins, Goel, Deng, Luthra, Xu, Gundogdu, Zhang, Yago~Vicente, Dideriksen, Arora, Guillaumin, and Malik]{collins2022abo}
Collins, J., Goel, S., Deng, K., Luthra, A., Xu, L., Gundogdu, E., Zhang, X., Yago~Vicente, T.~F., Dideriksen, T., Arora, H., Guillaumin, M., and Malik, J.
\newblock Abo: Dataset and benchmarks for real-world 3d object understanding.
\newblock \emph{CVPR}, 2022.

\bibitem[Community(2018)]{blender}
Community, B.~O.
\newblock \emph{Blender - a 3D modelling and rendering package}.
\newblock Blender Foundation, Stichting Blender Foundation, Amsterdam, 2018.
\newblock URL \url{http://www.blender.org}.

\bibitem[Dai et~al.(2023)Dai, Li, Li, Tiong, Zhao, Wang, Li, Fung, and Hoi]{instructblip}
Dai, W., Li, J., Li, D., Tiong, A. M.~H., Zhao, J., Wang, W., Li, B., Fung, P., and Hoi, S.
\newblock Instructblip: Towards general-purpose vision-language models with instruction tuning, 2023.

\bibitem[De~Luigi et~al.(2023)De~Luigi, Bolognini, Domeniconi, De~Gregorio, Poggi, and Di~Stefano]{deluigi2023scannerf}
De~Luigi, L., Bolognini, D., Domeniconi, F., De~Gregorio, D., Poggi, M., and Di~Stefano, L.
\newblock Scannerf: a scalable benchmark for neural radiance fields.
\newblock In \emph{Winter Conference on Applications of Computer Vision}, 2023.
\newblock WACV.

\bibitem[Dehghani et~al.(2023)Dehghani, Djolonga, Mustafa, Padlewski, Heek, Gilmer, Steiner, Caron, Geirhos, Alabdulmohsin, et~al.]{dehghani2023scaling}
Dehghani, M., Djolonga, J., Mustafa, B., Padlewski, P., Heek, J., Gilmer, J., Steiner, A.~P., Caron, M., Geirhos, R., Alabdulmohsin, I., et~al.
\newblock Scaling vision transformers to 22 billion parameters.
\newblock In \emph{International Conference on Machine Learning}, pp.\  7480--7512. PMLR, 2023.

\bibitem[Deitke et~al.(2023)Deitke, Schwenk, Salvador, Weihs, Michel, VanderBilt, Schmidt, Ehsani, Kembhavi, and Farhadi]{deitke2023objaverse}
Deitke, M., Schwenk, D., Salvador, J., Weihs, L., Michel, O., VanderBilt, E., Schmidt, L., Ehsani, K., Kembhavi, A., and Farhadi, A.
\newblock Objaverse: A universe of annotated 3d objects.
\newblock In \emph{CVPR}, pp.\  13142--13153, 2023.

\bibitem[Fang et~al.(2023)Fang, Wang, Xie, Sun, Wu, Wang, Huang, Wang, and Cao]{fang2023eva}
Fang, Y., Wang, W., Xie, B., Sun, Q., Wu, L., Wang, X., Huang, T., Wang, X., and Cao, Y.
\newblock Eva: Exploring the limits of masked visual representation learning at scale.
\newblock In \emph{CVPR}, pp.\  19358--19369, 2023.

\bibitem[Frostig et~al.(2018)Frostig, Johnson, and Leary]{frostig2018compiling}
Frostig, R., Johnson, M.~J., and Leary, C.
\newblock Compiling machine learning programs via high-level tracing.
\newblock \emph{Systems for Machine Learning}, 4\penalty0 (9), 2018.

\bibitem[Gao et~al.(2023)Gao, Sarkar, Xia, Xiao, Wu, Ichter, Majumdar, and Sadigh]{gao2023physically}
Gao, J., Sarkar, B., Xia, F., Xiao, T., Wu, J., Ichter, B., Majumdar, A., and Sadigh, D.
\newblock Physically grounded vision-language models for robotic manipulation.
\newblock \emph{arXiv preprint arXiv:2309.02561}, 2023.

\bibitem[Gordon et~al.(2018)Gordon, Kembhavi, Rastegari, Redmon, Fox, and Farhadi]{gordon2018iqa}
Gordon, D., Kembhavi, A., Rastegari, M., Redmon, J., Fox, D., and Farhadi, A.
\newblock Iqa: Visual question answering in interactive environments.
\newblock In \emph{Proceedings of the IEEE conference on computer vision and pattern recognition}, pp.\  4089--4098, 2018.

\bibitem[Gu et~al.(2023)Gu, Kuwajerwala, Morin, Jatavallabhula, Sen, Agarwal, Rivera, Paul, Ellis, Chellappa, Gan, de~Melo, Tenenbaum, Torralba, Shkurti, and Paull]{gu2023conceptgraphs}
Gu, Q., Kuwajerwala, A., Morin, S., Jatavallabhula, K.~M., Sen, B., Agarwal, A., Rivera, C., Paul, W., Ellis, K., Chellappa, R., Gan, C., de~Melo, C.~M., Tenenbaum, J.~B., Torralba, A., Shkurti, F., and Paull, L.
\newblock Conceptgraphs: Open-vocabulary 3d scene graphs for perception and planning, 2023.

\bibitem[Gupta et~al.(2019)Gupta, Dollar, and Girshick]{gupta2019lvis}
Gupta, A., Dollar, P., and Girshick, R.
\newblock Lvis: A dataset for large vocabulary instance segmentation.
\newblock In \emph{CVPR}, pp.\  5356--5364, 2019.

\bibitem[Gupta \& Kembhavi(2023)Gupta and Kembhavi]{Gupta_2023_CVPR}
Gupta, T. and Kembhavi, A.
\newblock Visual programming: Compositional visual reasoning without training.
\newblock In \emph{CVPR}, pp.\  14953--14962, June 2023.

\bibitem[Han et~al.(2020)Han, Chen, Liu, and Zwicker]{han2020shapecaptioner}
Han, Z., Chen, C., Liu, Y.-S., and Zwicker, M.
\newblock Shapecaptioner: Generative caption network for 3d shapes by learning a mapping from parts detected in multiple views to sentences.
\newblock In \emph{Proceedings of the 28th ACM International Conference on Multimedia}, pp.\  1018--1027, 2020.

\bibitem[He et~al.(2021)He, Yu, Liu, Yang, Sun, Wang, Fu, Zou, and Mian]{he2021deep}
He, Y., Yu, H., Liu, X., Yang, Z., Sun, W., Wang, Y., Fu, Q., Zou, Y., and Mian, A.
\newblock Deep learning based 3d segmentation: A survey.
\newblock \emph{arXiv preprint arXiv:2103.05423}, 2021.

\bibitem[Hermann et~al.(2020)Hermann, Chen, and Kornblith]{NEURIPS2020_db5f9f42}
Hermann, K., Chen, T., and Kornblith, S.
\newblock The origins and prevalence of texture bias in convolutional neural networks.
\newblock In Larochelle, H., Ranzato, M., Hadsell, R., Balcan, M., and Lin, H. (eds.), \emph{NeurIPS}, volume~33, pp.\  19000--19015. Curran Associates, Inc., 2020.
\newblock URL \url{https://proceedings.neurips.cc/paper_files/paper/2020/file/db5f9f42a7157abe65bb145000b5871a-Paper.pdf}.

\bibitem[Hong et~al.(2023)Hong, Zhen, Chen, Zheng, Du, Chen, and Gan]{hong20233d}
Hong, Y., Zhen, H., Chen, P., Zheng, S., Du, Y., Chen, Z., and Gan, C.
\newblock 3d-llm: Injecting the 3d world into large language models.
\newblock \emph{arXiv preprint arXiv:2307.12981}, 2023.

\bibitem[Jia et~al.(2021)Jia, Yang, Xia, Chen, Parekh, Pham, Le, Sung, Li, and Duerig]{jia2021scaling}
Jia, C., Yang, Y., Xia, Y., Chen, Y.-T., Parekh, Z., Pham, H., Le, Q., Sung, Y.-H., Li, Z., and Duerig, T.
\newblock Scaling up visual and vision-language representation learning with noisy text supervision.
\newblock In \emph{International conference on machine learning}, pp.\  4904--4916. PMLR, 2021.

\bibitem[Koo et~al.(2022)Koo, Huang, Achlioptas, Guibas, and Sung]{koo2022partglot}
Koo, J., Huang, I., Achlioptas, P., Guibas, L.~J., and Sung, M.
\newblock Partglot: Learning shape part segmentation from language reference games.
\newblock In \emph{Proceedings of the IEEE/CVF Conference on Computer Vision and Pattern Recognition}, pp.\  16505--16514, 2022.

\bibitem[Korman et~al.(2018)Korman, Mack, Jett, and Renear]{Korman2018-KORDTE}
Korman, D.~Z., Mack, E., Jett, J., and Renear, A.~H.
\newblock Defining textual entailment.
\newblock \emph{Journal of the Association for Information Science and Technology}, 69:\penalty0 763--772, 2018.

\bibitem[Kudo \& Richardson(2018)Kudo and Richardson]{kudo2018sentencepiece}
Kudo, T. and Richardson, J.
\newblock Sentencepiece: A simple and language independent subword tokenizer and detokenizer for neural text processing.
\newblock \emph{arXiv preprint arXiv:1808.06226}, 2018.

\bibitem[Kuznetsova et~al.(2020)Kuznetsova, Rom, Alldrin, Uijlings, Krasin, Pont-Tuset, Kamali, Popov, Malloci, Kolesnikov, et~al.]{kuznetsova2020open}
Kuznetsova, A., Rom, H., Alldrin, N., Uijlings, J., Krasin, I., Pont-Tuset, J., Kamali, S., Popov, S., Malloci, M., Kolesnikov, A., et~al.
\newblock The open images dataset v4: Unified image classification, object detection, and visual relationship detection at scale.
\newblock \emph{IJCV}, 128\penalty0 (7):\penalty0 1956--1981, 2020.

\bibitem[Li et~al.(2022{\natexlab{a}})Li, Li, Le, Wang, Savarese, and Hoi]{li2022lavis}
Li, D., Li, J., Le, H., Wang, G., Savarese, S., and Hoi, S.~C.
\newblock Lavis: A library for language-vision intelligence.
\newblock \emph{arXiv preprint arXiv:2209.09019}, 2022{\natexlab{a}}.

\bibitem[Li et~al.(2022{\natexlab{b}})Li, Li, Xiong, and Hoi]{li2022blip}
Li, J., Li, D., Xiong, C., and Hoi, S.
\newblock Blip: Bootstrapping language-image pre-training for unified vision-language understanding and generation.
\newblock In \emph{International Conference on Machine Learning}, pp.\  12888--12900. PMLR, 2022{\natexlab{b}}.

\bibitem[Li et~al.(2023)Li, Li, Savarese, and Hoi]{li2023blip}
Li, J., Li, D., Savarese, S., and Hoi, S.
\newblock Blip-2: Bootstrapping language-image pre-training with frozen image encoders and large language models.
\newblock \emph{arXiv preprint arXiv:2301.12597}, 2023.

\bibitem[Li et~al.(2019)Li, Yatskar, Yin, Hsieh, and Chang]{li2019visualbert}
Li, L.~H., Yatskar, M., Yin, D., Hsieh, C.-J., and Chang, K.-W.
\newblock Visualbert: A simple and performant baseline for vision and language.
\newblock \emph{arXiv preprint arXiv:1908.03557}, 2019.

\bibitem[Luo et~al.(2023)Luo, Rockwell, Lee, and Johnson]{luo2023scalable}
Luo, T., Rockwell, C., Lee, H., and Johnson, J.
\newblock Scalable 3d captioning with pretrained models.
\newblock In Oh, A., Naumann, T., Globerson, A., Saenko, K., Hardt, M., and Levine, S. (eds.), \emph{Advances in Neural Information Processing Systems}, volume~36, pp.\  75307--75337. Curran Associates, Inc., 2023.

\bibitem[Miller(1995)]{miller1995wordnet}
Miller, G.~A.
\newblock Wordnet: a lexical database for english.
\newblock \emph{Communications of the ACM}, 38\penalty0 (11):\penalty0 39--41, 1995.

\bibitem[Naseer et~al.(2021)Naseer, Ranasinghe, Khan, Hayat, Shahbaz~Khan, and Yang]{NEURIPS2021_c404a5ad}
Naseer, M.~M., Ranasinghe, K., Khan, S.~H., Hayat, M., Shahbaz~Khan, F., and Yang, M.-H.
\newblock Intriguing properties of vision transformers.
\newblock In Ranzato, M., Beygelzimer, A., Dauphin, Y., Liang, P., and Vaughan, J.~W. (eds.), \emph{NeurIPS}, volume~34, pp.\  23296--23308. Curran Associates, Inc., 2021.
\newblock URL \url{https://proceedings.neurips.cc/paper_files/paper/2021/file/c404a5adbf90e09631678b13b05d9d7a-Paper.pdf}.

\bibitem[O'Connell et~al.(2023)O'Connell, Bonnen, Friedman, Tewari, Tenenbaum, Sitzmann, and Kanwisher]{o2023approaching}
O'Connell, T.~P., Bonnen, T., Friedman, Y., Tewari, A., Tenenbaum, J.~B., Sitzmann, V., and Kanwisher, N.
\newblock Approaching human 3d shape perception with neurally mappable models.
\newblock \emph{arXiv preprint arXiv:2308.11300}, 2023.

\bibitem[OpenAI(2023)]{openai2023gpt4}
OpenAI.
\newblock Gpt-4 technical report, 2023.

\bibitem[Piergiovanni et~al.(2022)Piergiovanni, Kuo, and Angelova]{piergiovanni2022pre}
Piergiovanni, A., Kuo, W., and Angelova, A.
\newblock Pre-training image-language transformers for open-vocabulary tasks.
\newblock \emph{arXiv preprint arXiv:2209.04372}, 2022.

\bibitem[Radford et~al.(2021)Radford, Kim, Hallacy, Ramesh, Goh, Agarwal, Sastry, Askell, Mishkin, Clark, et~al.]{radford2021learning}
Radford, A., Kim, J.~W., Hallacy, C., Ramesh, A., Goh, G., Agarwal, S., Sastry, G., Askell, A., Mishkin, P., Clark, J., et~al.
\newblock Learning transferable visual models from natural language supervision.
\newblock In \emph{International conference on machine learning}, pp.\  8748--8763. PMLR, 2021.

\bibitem[Raffel et~al.(2020)Raffel, Shazeer, Roberts, Lee, Narang, Matena, Zhou, Li, and Liu]{raffel2020exploring}
Raffel, C., Shazeer, N., Roberts, A., Lee, K., Narang, S., Matena, M., Zhou, Y., Li, W., and Liu, P.~J.
\newblock Exploring the limits of transfer learning with a unified text-to-text transformer.
\newblock \emph{The Journal of Machine Learning Research}, 21\penalty0 (1):\penalty0 5485--5551, 2020.

\bibitem[Roberts et~al.(2022)Roberts, Chung, Levskaya, Mishra, Bradbury, Andor, Narang, Lester, Gaffney, Mohiuddin, Hawthorne, Lewkowycz, Salcianu, van Zee, Austin, Goodman, Soares, Hu, Tsvyashchenko, Chowdhery, Bastings, Bulian, Garcia, Ni, Chen, Kenealy, Clark, Lee, Garrette, Lee-Thorp, Raffel, Shazeer, Ritter, Bosma, Passos, Maitin-Shepard, Fiedel, Omernick, Saeta, Sepassi, Spiridonov, Newlan, and Gesmundo]{roberts2022t5x}
Roberts, A., Chung, H.~W., Levskaya, A., Mishra, G., Bradbury, J., Andor, D., Narang, S., Lester, B., Gaffney, C., Mohiuddin, A., Hawthorne, C., Lewkowycz, A., Salcianu, A., van Zee, M., Austin, J., Goodman, S., Soares, L.~B., Hu, H., Tsvyashchenko, S., Chowdhery, A., Bastings, J., Bulian, J., Garcia, X., Ni, J., Chen, A., Kenealy, K., Clark, J.~H., Lee, S., Garrette, D., Lee-Thorp, J., Raffel, C., Shazeer, N., Ritter, M., Bosma, M., Passos, A., Maitin-Shepard, J., Fiedel, N., Omernick, M., Saeta, B., Sepassi, R., Spiridonov, A., Newlan, J., and Gesmundo, A.
\newblock Scaling up models and data with $\texttt{t5x}$ and $\texttt{seqio}$.
\newblock \emph{arXiv preprint arXiv:2203.17189}, 2022.
\newblock URL \url{https://arxiv.org/abs/2203.17189}.

\bibitem[Sun et~al.(2017)Sun, Shrivastava, Singh, and Gupta]{sun2017revisiting}
Sun, C., Shrivastava, A., Singh, S., and Gupta, A.
\newblock Revisiting unreasonable effectiveness of data in deep learning era.
\newblock In \emph{ICCV}, pp.\  843--852, 2017.

\bibitem[Sur{\'\i}s et~al.(2023)Sur{\'\i}s, Menon, and Vondrick]{suris2023vipergpt}
Sur{\'\i}s, D., Menon, S., and Vondrick, C.
\newblock Vipergpt: Visual inference via python execution for reasoning.
\newblock \emph{arXiv preprint arXiv:2303.08128}, 2023.

\bibitem[Tay et~al.(2022)Tay, Dehghani, Tran, Garcia, Bahri, Schuster, Zheng, Houlsby, and Metzler]{tay2022unifying}
Tay, Y., Dehghani, M., Tran, V.~Q., Garcia, X., Bahri, D., Schuster, T., Zheng, H.~S., Houlsby, N., and Metzler, D.
\newblock Unifying language learning paradigms.
\newblock \emph{arXiv preprint arXiv:2205.05131}, 2022.

\bibitem[Vaswani et~al.(2017)Vaswani, Shazeer, Parmar, Uszkoreit, Jones, Gomez, Kaiser, and Polosukhin]{vaswani2017attention}
Vaswani, A., Shazeer, N., Parmar, N., Uszkoreit, J., Jones, L., Gomez, A.~N., Kaiser, {\L}., and Polosukhin, I.
\newblock Attention is all you need.
\newblock \emph{NeurIPS}, 30, 2017.

\bibitem[Wald et~al.(2020)Wald, Dhamo, Navab, and Tombari]{3DSSG2020}
Wald, J., Dhamo, H., Navab, N., and Tombari, F.
\newblock Learning 3d semantic scene graphs from 3d indoor reconstructions.
\newblock In \emph{CVPR}, 2020.

\bibitem[Wu et~al.(2023)Wu, Zhang, Fu, Wang, Ren, Pan, Wu, Yang, Wang, Qian, et~al.]{wu2023omniobject3d}
Wu, T., Zhang, J., Fu, X., Wang, Y., Ren, J., Pan, L., Wu, W., Yang, L., Wang, J., Qian, C., et~al.
\newblock Omniobject3d: Large-vocabulary 3d object dataset for realistic perception, reconstruction and generation.
\newblock In \emph{CVPR}, pp.\  803--814, 2023.

\bibitem[Wu et~al.(2016)Wu, Schuster, Chen, Le, Norouzi, Macherey, Krikun, Cao, Gao, Macherey, et~al.]{wu2016google}
Wu, Y., Schuster, M., Chen, Z., Le, Q.~V., Norouzi, M., Macherey, W., Krikun, M., Cao, Y., Gao, Q., Macherey, K., et~al.
\newblock Google's neural machine translation system: Bridging the gap between human and machine translation.
\newblock \emph{arXiv preprint arXiv:1609.08144}, 2016.

\bibitem[Yang et~al.(2019)Yang, Tang, Zhang, and Cai]{yang2019auto}
Yang, X., Tang, K., Zhang, H., and Cai, J.
\newblock Auto-encoding scene graphs for image captioning.
\newblock In \emph{Proceedings of the IEEE/CVF conference on computer vision and pattern recognition}, pp.\  10685--10694, 2019.

\bibitem[Yao et~al.(2020)Yao, Lin, Xia, Zhao, and Zhou]{yao2020video}
Yao, R., Lin, G., Xia, S., Zhao, J., and Zhou, Y.
\newblock Video object segmentation and tracking: A survey.
\newblock \emph{ACM Transactions on Intelligent Systems and Technology (TIST)}, 11\penalty0 (4):\penalty0 1--47, 2020.

\bibitem[Zareian et~al.(2021)Zareian, Rosa, Hu, and Chang]{zareian2021open}
Zareian, A., Rosa, K.~D., Hu, D.~H., and Chang, S.-F.
\newblock Open-vocabulary object detection using captions.
\newblock In \emph{Proceedings of the IEEE/CVF Conference on Computer Vision and Pattern Recognition}, pp.\  14393--14402, 2021.

\bibitem[Zhai et~al.(2022)Zhai, Kolesnikov, Houlsby, and Beyer]{zhai2022scaling}
Zhai, X., Kolesnikov, A., Houlsby, N., and Beyer, L.
\newblock Scaling vision transformers.
\newblock In \emph{CVPR}, pp.\  12104--12113, 2022.

\bibitem[Zhu et~al.(2023)Zhu, Chen, Haydarov, Shen, Zhang, and Elhoseiny]{zhu2023chatgpt}
Zhu, D., Chen, J., Haydarov, K., Shen, X., Zhang, W., and Elhoseiny, M.
\newblock Chatgpt asks, blip-2 answers: Automatic questioning towards enriched visual descriptions.
\newblock \emph{arXiv preprint arXiv:2303.06594}, 2023.

\end{thebibliography}
